\documentclass[10pt,twocolumn,letterpaper]{article}

\usepackage{wacv}
\usepackage{times}
\usepackage{epsfig}
\usepackage{graphicx}
\usepackage{amsmath}
\usepackage{amssymb}
\usepackage{multirow}
\usepackage{subcaption}
\usepackage[table]{xcolor}
\usepackage{pdfpages}

\usepackage[pagebackref=true,breaklinks=true,letterpaper=true,colorlinks,bookmarks=false]{hyperref}

\newcommand{\mpage}[2]
{
\begin{minipage}{#1\linewidth}\centering
#2
\end{minipage}
}

\wacvfinalcopy 

\author{
    Iaroslav Melekhov$^1$\thanks{The majority of the work was done during internship at ETH Z{\"u}rich.} \\
\and
    Aleksei Tiulpin$^2$ \\
\and
    Torsten Sattler$^3$ \\
\and
    Marc Pollefeys$^{3,5}$ \\
\and
    Esa Rahtu$^4$\\
\and
    Juho Kannala$^1$\\
    \small$^1$Aalto University \hspace{1pt} $^2$University of Oulu \hspace{1pt} $^3$ETH Z{\"u}rich \hspace{1pt} $^4$Tampere University of Technology \hspace{1pt} $^5$Microsoft\\
    \small\texttt{\{melekhi1, jkannala1\}@aalto.fi} \hspace{2.5pt} \texttt{aleksei.tiulpin@oulu.fi}\\
    \small\texttt{esa.rahtu@tut.fi} \hspace{2.5pt} \texttt{\{sattlert, marc.pollefeys\}@inf.ethz.ch}\\
}

\ifwacvfinal\fi
\begin{document}

\title{DGC-Net: Dense Geometric Correspondence Network}

\maketitle

\begin{abstract}
This paper addresses the challenge of dense pixel correspondence estimation between two images. This problem is closely related to optical flow estimation task where ConvNets (CNNs) have recently achieved significant progress. While optical flow methods produce very accurate results for the small pixel translation and limited appearance variation scenarios, they hardly deal with the strong geometric transformations that we consider in this work. In this paper, we propose a coarse-to-fine CNN-based framework that can leverage the advantages of optical flow approaches and extend them to the case of large transformations providing dense and subpixel accurate estimates. It is trained on synthetic transformations and demonstrates very good performance to unseen, realistic, data. Further, we apply our method to the problem of relative camera pose estimation and demonstrate that the model outperforms existing dense approaches. 
\end{abstract}

\section{Introduction}

Finding correspondences between images is a key task in many computer vision applications, including image alignment~\cite{Rocco17,Rocco18}, visual localization~\cite{SemanticLocalization,inloc,Localization2}, image retrieval~\cite{NeuralCodes,Gordo}, structure-from-motion~\cite{colmap}, semantic correspondence~\cite{scnet,zak1}, optical flow~\cite{FlowNet2,Janai2018ECCV,Spynet,PWC-Net} and relative camera pose estimation~\cite{RelativePoseCNN,demon}. In general, there are two ways to establish a pixel-wise correspondence field between images. The first group of methods is based on applying feature descriptors to an image pair and utilizing 
nearest neighbor criterion to match keypoints globally. However, these approaches do not produce dense correspondences explicitly and apply interpolation or local affine transformations~\cite{bFunEccv14} to turn a sparse set into a pixel-wise correspondences. Another possible direction of finding dense correspondences is to compare image patches in feature space. Neural networks have been widely used to learn discriminative and robust descriptors~\cite{Balntas,DeepStereo}. 
Those descriptors are then compared pair-wise by thresholding Euclidean distance between them~\cite{UCN,Hierarchy2D3DMatching,patchM} or by predicting a binary label~\cite{ZagoruykoCVPR2015,Zbontar}. In contrast, the proposed approach processes the image as a whole, and thus, it can handle a broader set of geometric changes in images and directly predict dense correspondences without any post-processing steps. Recent optical flow methods~\cite{FlowNet2,PWC-Net} have demonstrated great success at estimating dense sub-pixel correspondences. However, the main limitation of these methods is a spatially constrained correlation layer predicting the matches in a small vicinity around the center pixel of each image patch. Thus, captured transformations are very restricted. To some extent this restriction can be alleviated with pyramid structure~\cite{PWC-Net} but not completely.

In this paper we propose a convolutional neural network (CNN) architecture, called DGC-Net, for learning dense pixel correspondences between a pair of images with strong geometric transformations. Following more recent optical flow methods~\cite{FlowNet2,Spynet,PWC-Net} and the concept introduced by Lucas-Kanade~\cite{lucas-kanade}, we exploit a coarse-to-fine image warping idea by creating a hierarchical network structure. Rather than considering only affine and thin-plate spline (TPS) transformations~\cite{Rocco17}, 
we train our system on synthetic data in an end-to-end manner handling diverse geometric transformations present in real world. We demonstrate that the proposed approach substantially outperforms CNN-based optical flow and image matching methods on the challenging HPatches~\cite{HPatches} and DTU~\cite{DTU} datasets. 

The main contributions of this paper are: 1) We propose an end-to-end CNN-based method, DGC-Net, to establish dense pixel correspondences between images with strong geometric transformations; 2) We demonstrate that even if DGC-Net is trained only on synthetic transformations, it can generalize well to real data; 3) We apply the proposed approach to the problem of relative camera pose estimation and demonstrate that our method outperforms strong baseline approaches by a large margin. In addition, we modify the original structure of DGC-Net and seamlessly integrate a matchability decoder into DGC-Net that can significantly improve the computational efficiency of the relative camera pose estimation pipeline by removing tentative correspondences with low confidence scores.

\vspace{-1mm}
\section{Related Work}
The traditional image matching pipeline begins with the detection of interest points and computation of descriptors. However, many of the most widely used descriptors~\cite{SURF,SIFT} are based on hand-crafted features and have limited ability to cope with negative factors, such as strong illumination changes and large variation in viewpoint. In contrast, more recent~\cite{MODS,ASIFT} methods based on view-synthesizing have demonstrated state-of-the-art results in image matching by handling large viewing angle difference and appearance changes. However, they do not produce dense per-pixel correspondences and do not perform any learning. 

Applying machine learning techniques has proven very effective in optical flow estimation problem~\cite{FlowNet2,Spynet,PWC-Net} which is closely related to finding pixel correspondence task. Recently proposed methods,~\ie PWC-Net~\cite{PWC-Net} and FlowNet2~\cite{FlowNet2}, utilize a correlation layer to predict image similarities in some neighborhood around the center pixel in a coarse-to-fine manner. While such a spatially constrained correlation layer leads to state-of-the-art results in optical flow, it performs poorly for very strong geometric transformations that we consider in this work. Rocco~\etal~\cite{Rocco17} proposed a CNN-based approach for determining correspondences between two images and applying it to instance-level and category-level tasks. In contrast to optical flow methods~\cite{FlowNet2,PWC-Net}, it comprises a matching layer calculating the correlation between target and reference feature maps without any spatial constraint. The method casts finding pixel correspondences task as a regression problem and consisting of two independent Siamese CNNs trained separately and directly predicting affine and TPS geometric transformations parametrizing 6-element and 18-element vectors. On the contrary, we propose a more general approach handling more diverse transformations and operating in an end-to-end fashion. 

Similarly to~\cite{UCN}, Fathy~\etal~\cite{Hierarchy2D3DMatching} proposed a CNN-based dense descriptor for 2D and 3D matching. However, their goal is very different to ours requiring strong supervision in the form of per-pixel ground-truth labels to compare extracted feature vectors and establish correspondences. 


\begin{figure*}[t]
  \centering
    \includegraphics[width=.9\linewidth]{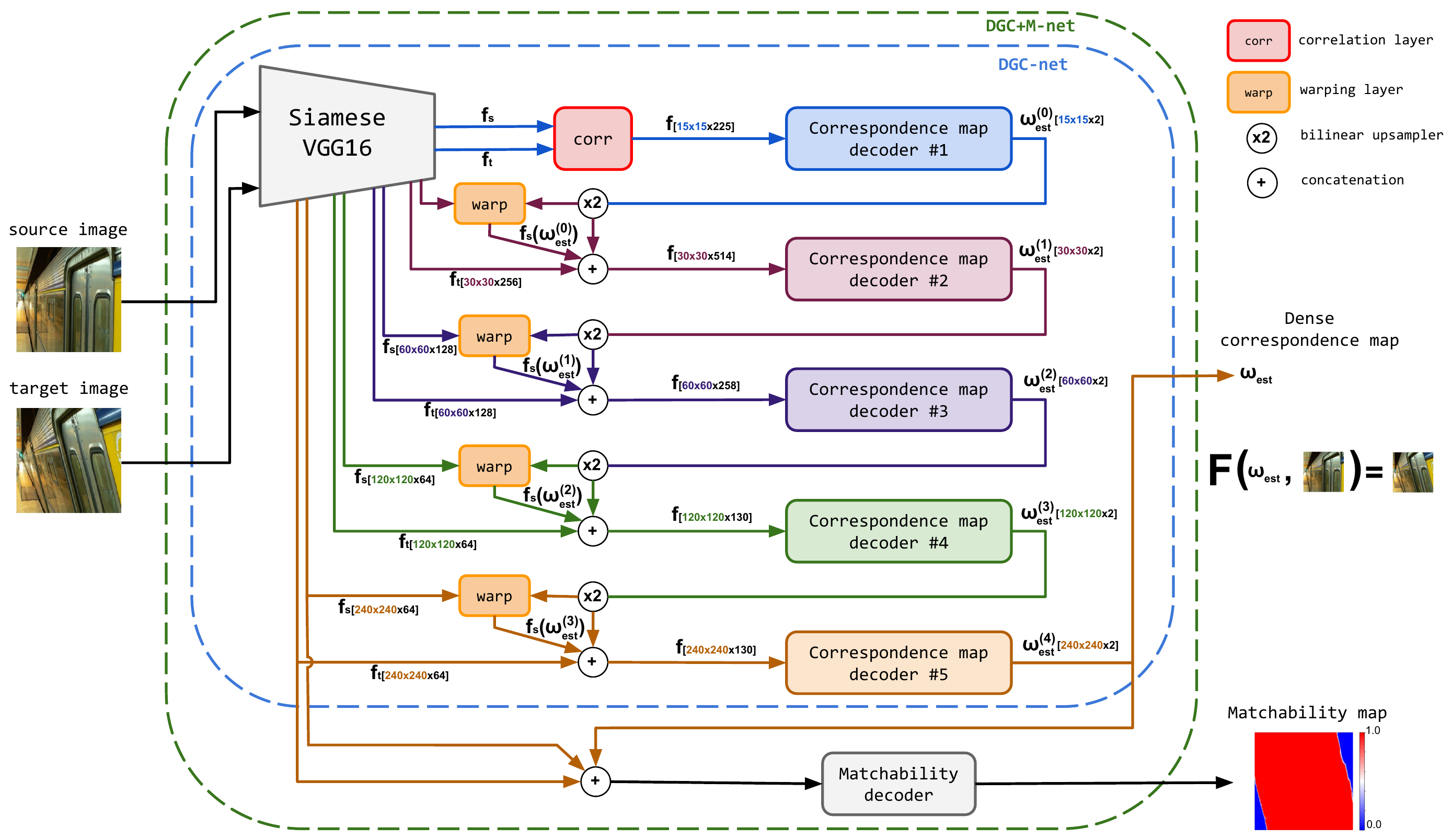}
  \caption{Overview of our proposed iterative architecture \texttt{DGC-Net} consisting of four major components: 1) the \textbf{feature pyramid creator}. 2) the \textbf{correlation layer} estimates the pairwise similarity score of the source and target feature descriptors. 3) the fully convolutional \textbf{correspondence map decoders} predict the dense correspondence map between input image pair at each level of the feature pyramid. 4) the \textbf{warping layer} warps features of the source image using the upsampled transforming grid from a correspondence map decoder. The \textbf{matchability decoder} is a tiny CNN that predicts a confidence map with higher scores for those pixels in the source image that have correspondences in the target. See Sec.~\ref{ssec:net_arch} for more details.}
 \label{fig:f_pipeline}
 \vspace{-2mm}
\end{figure*}

\vspace{-2mm}
\section{Method}\label{sec:method}
\vspace{-2mm}

Our goal is to determine correspondences between two input images $\mathbf{I_s}, \mathbf{I_t} \in \mathbb{R}^{W\times H\times 3}$. The most straightforward way to solve this task is to predict the parameters of the relative transformation matrix parametrized for different geometric transformations, such as an homography~\cite{DeepHomography}, an affine  or a TPS~\cite{Rocco17} transformation. However, realistic scenes usually contain more complex geometric transformations which can be hardly described by such parametrization. Inspired by recent work in image compositing~\cite{st-gan} and optical flow estimation, we propose to predict a dense pixel correspondence map $\omega \in \mathbb{R}^{W\times H\times 2}$ in an coarse-to-fine manner.

\subsection{Network Architecture}\label{ssec:net_arch}
In this section, we first present the structure of the proposed network and the general principles behind it, then formulate the view correspondence objective function to predict geometric transformations between image pairs.

Schematic representation of the proposed approach is shown in Fig.~\ref{fig:f_pipeline}. A pair of input images is fed into a module consisting of two pre-trained CNN branches which construct a feature pyramid. The correlation layer takes feature maps of the source and target images from the coarse (top) level of the pyramid and estimates the pairwise similarity between them. Then, the correspondence map decoder takes the output of the correlation layer and directly predicts pixel correspondences for this particular level of the pyramid. The estimates are then refined in an iterative manner. 

\noindent\textbf{Feature pyramid creator.}
In order to create a representation of an input image pair in feature space, we construct a Siamese neural network with two branches with shared weights. The branches use the VGG-16 architecture~\cite{vgg} trained on ImageNet~\cite{imagenet} and truncated at the last pooling layer followed by $L2$-normalization~\cite{Rocco17}. We extract features $\mathbf{f}_{s},\,\mathbf{f}_{t}$ at different parts of each branch to create a 5-layer feature pyramid with the following spatial resolutions (from top to bottom): $\left[15\times15,\,30\times30,\,60\times60,\,120\times120,\, 240\times240\right]$ and encoded with different colors in Fig.~\ref{fig:f_pipeline}. The weights of CNN-branches are then fixed throughout the rest of the network training procedure.

\noindent\textbf{Correlation layer.} In order to estimate a similarity score between two images, we follow an idea proposed in~\cite{Rocco17} and calculate the correlation volume between the normalized feature maps of the source and target images. In contrast to optical flow approaches~\cite{FlowNet2,PWC-Net}, where the correlation volume is computed for the raw features in a restricted area around the center pixel, we compute global correlation and apply $L2$-normalization before and after the correlation layer to strongly down-weight ambiguous matches (\cf Fig.~\ref{fig:f_pipeline}). Specifically, the correlation layer computes the scalar product between each feature vector of the source $\mathbf{f}_{s}$ and all vectors of the target $\mathbf{f}_{t}$ feature maps $\left(\mathbf{f}_{s},\mathbf{f}_{t} \in \mathbb{R}^{W\times H\times C}\right)$ and can be defined in the following way:

\begin{equation}\label{eq:eq_corr}
    \mathbf{c}_{st}\left(i,j\right) = \langle \mathbf{f}_{s}\left(i,j\right), \mathbf{f}_{t}\left(i,j\right)\rangle \enspace ,
\end{equation}
\noindent where $\langle.\rangle$ denotes the scalar product and $\mathbf{c}_{st}$ is a L2-normalized correlation volume $\mathbf{c}_{st} \in \mathbb{R}^{W \times H \times \left(W \times H \right)}$. Since the third dimension of the correlation volume is a product of its $W$ and $H$, it is not feasible to calculate such volumes in the bottom layers of the pyramid where the spatial resolution of the feature maps is large. Thus, at the bottom feature pyramid layers, we concatenate descriptors channel-wise.

\noindent\textbf{Correspondence map decoder.} 
The output of the correlation layer is then fed into a correspondence map decoder consisting of 5 convolutional blocks (Conv-BN-ReLU) to estimate a 2D dense correspondence field $\omega_{\text{est}}^{\left(l\right)}$ at a particular level $l$ of the feature pyramid. The estimates are parameterized such that each predicted pixel location in the map belongs to the interval $\left[-1,1\right]$ representing width and height normalized image coordinates. 
That is, we upsample the predicted correspondence field at the $\left(l-1\right)^{th}$ level to warp the feature maps of the source image at $l^{th}$ level toward the target features. Finally, the upsampled field, warped source $f_s(\omega_{\text{est}}^{\left(l\right)})$ and target $f_t^{\left(l\right)}$ features are concatenated along the channel dimension and provided accordingly as input to the correspondence map decoder at $l^{th}$ level. 

Each convolution layer in the decoder is padded to keep the spatial resolution of the feature maps intact. Moreover, in order to be able to capture more spatial context at the bottom layers of the pyramid, starting from $l=3$ different dilation factors have been added to the convolution blocks to increase the receptive field. The feature pyramid creator, correlation layer and a hierarchical chain of the correspondence map decoders together form a CNN architecture that we will refer to as \texttt{DGC-Net} in the following.

Given an image pair and the ground truth pixel correspondence map $\omega_\text{gt}$, we can define a hierarchical objective loss function as follows: 
\begin{equation}\label{eq:eq_loss_crsp}
    \mathcal{L}_c = \sum_{l=0}^{L-1} \alpha^{\left(l\right)} \frac{1}{N_\text{val}^{\left(l\right)}} \sum_{x}^{N_\text{val}^{\left(l\right)}} M_\text{gt}^{\left(l\right)} \left\| \omega_\text{est}^{\left(l\right)}\left(x\right) - \omega_\text{gt}^{\left(l\right)}\left(x\right)  \right\|_1
\end{equation}
\noindent where $\left\| . \right\|_1$ is the $L1$ distance between an estimated $\omega_\text{est}^{\left(l\right)}$ and the ground truth $\omega_\text{gt}^{\left(l\right)}$ dense correspondence map; $M_\text{gt}^{\left(l\right)}$ is the ground truth binary mask (matchability mask) indicating whether each pixel in the source image has a correspondence in the target; $x$ indexes over valid pixel locations $N_\text{val}^{\left(l\right)}$ according to the ground truth mask at each level $l$ of the $L$-level feature pyramid. In order to adjust the weight of different pyramid layers, we introduce a vector of scalar weight coefficients $\alpha^{\left(l\right)}$. 

\noindent\textbf{Matchability decoder.}
According to recent advances in optical flow~\cite{Janai2018ECCV,PWC-Net}, it is still very challenging to estimate correct correspondences for ill-posed cases, such as occluded regions of an image pair. Thus, in addition to the pixel correspondence map produced by DGC-Net, we would like to directly predict a measure of confidence for each correspondence. Specifically, we  modify the DGC-Net structure by adding a matchability branch. It contains four convolutional layers outputting a probability map (parametrized as a sigmoid) indicating a confidence score for each pixel location in the predicted correspondence map. We will refer to this architecture as called~\texttt{DGC+M-Net}. Since, we consider this problem as a pixel classification task, we optimize a binary cross entropy (BCE) with logits loss that is defined as:
\begin{equation}\label{eq:eq_loss_map}
    \mathcal{L}_{m} = -\frac{1}{N}\sum_{i=0}^{N-1}\left(y_i\log\sigma\left(\hat{y}_i\right) + \left(1 - y_i\right)\log\left(1-\sigma\left(\hat{y}_i\right)\right)\right)
\end{equation}
\noindent where $y_i$ and $\hat{y}_i$ are ground truth and  estimated matchability masks, respectively; $\sigma$ is the element-wise sigmoid function. The total loss for the DGC+M-Net model is the sum of the correspondence loss $\mathcal{L}_{c}$ and the matchability loss $\mathcal{L}_{m}$ with a weighted coefficient $\beta$ ($\beta=1$):
\begin{equation}
    \mathcal{L} = \mathcal{L}_{c} + \beta \mathcal{L}_{m} \enspace .
\end{equation}

We provide the detailed information about the hyperparameters used in training as well as the exact network definitions of all network components in supplementary.

\begin{figure*}[t!]
 	\centering
 	\begin{subfigure}[t]{.2\textwidth}
 		\centering
 		\includegraphics[width=\textwidth]{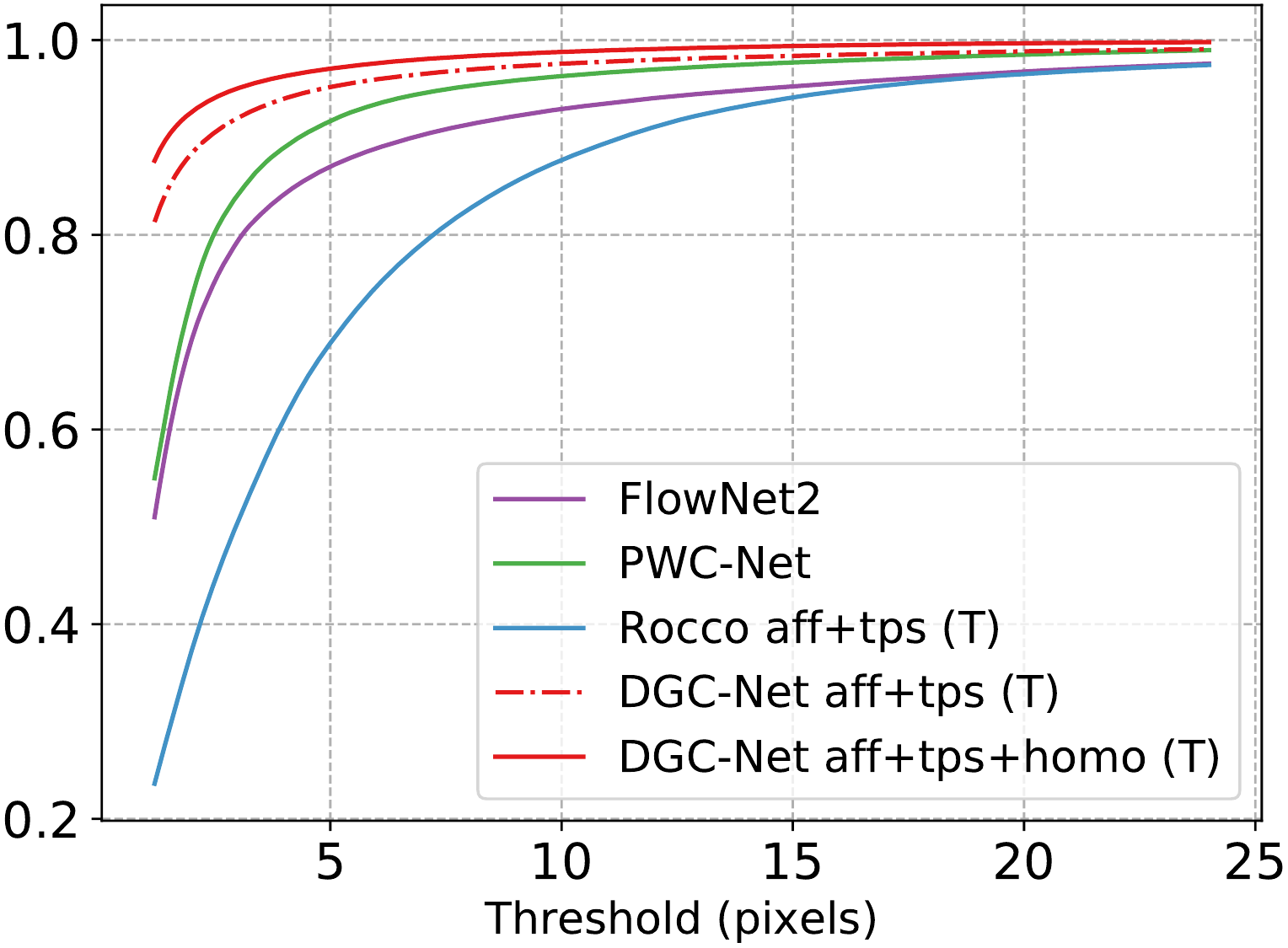}
 		\caption{Viewpoint I}
 	\end{subfigure}%
 	~
 	\begin{subfigure}[t]{.2\textwidth}
 		\centering
 		\includegraphics[width=\textwidth]{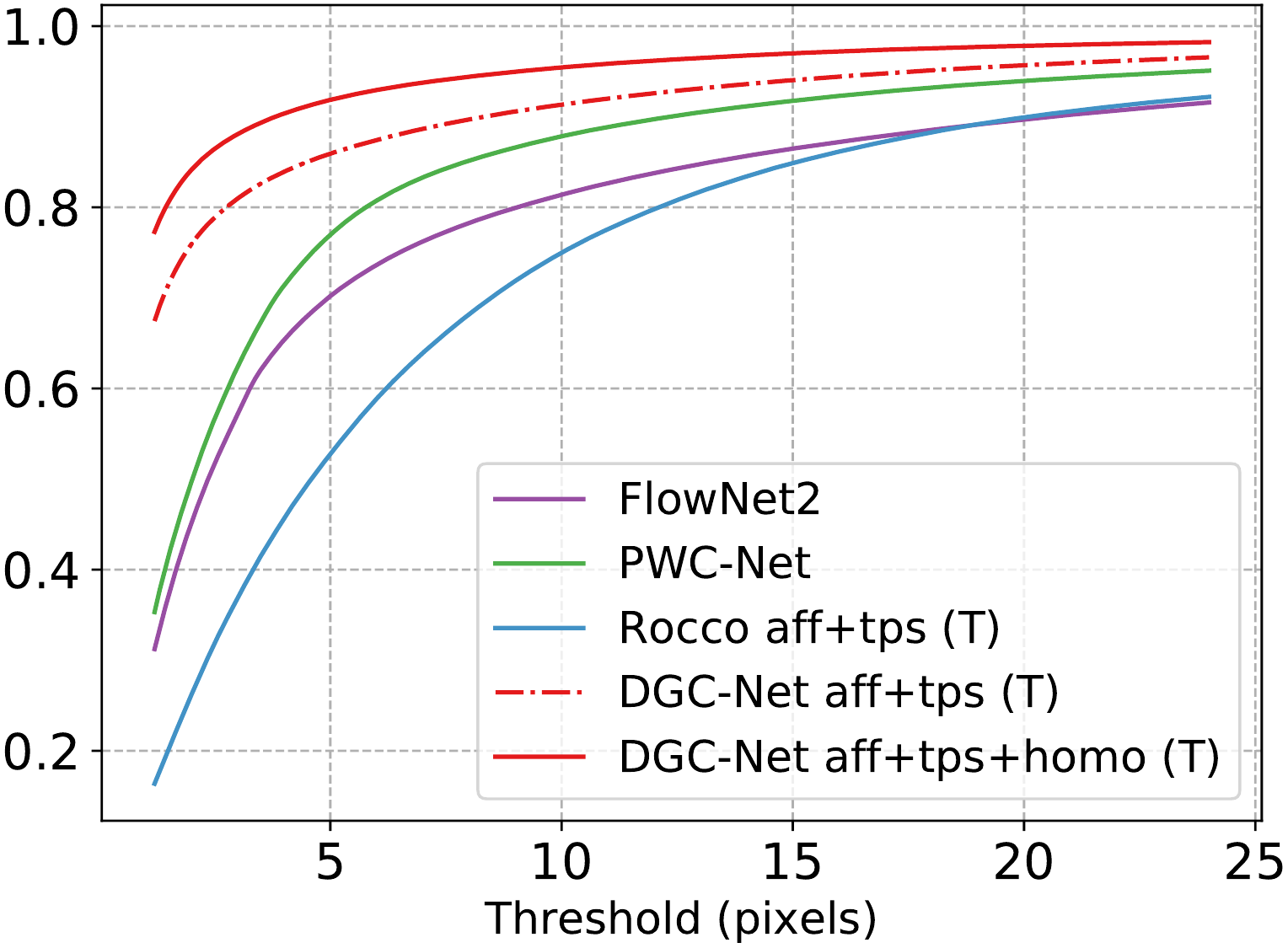}
 		\caption{Viewpoint II}
 	\end{subfigure}%
 	~
 	\begin{subfigure}[t]{.2\textwidth}
 		\centering
 		\includegraphics[width=\textwidth]{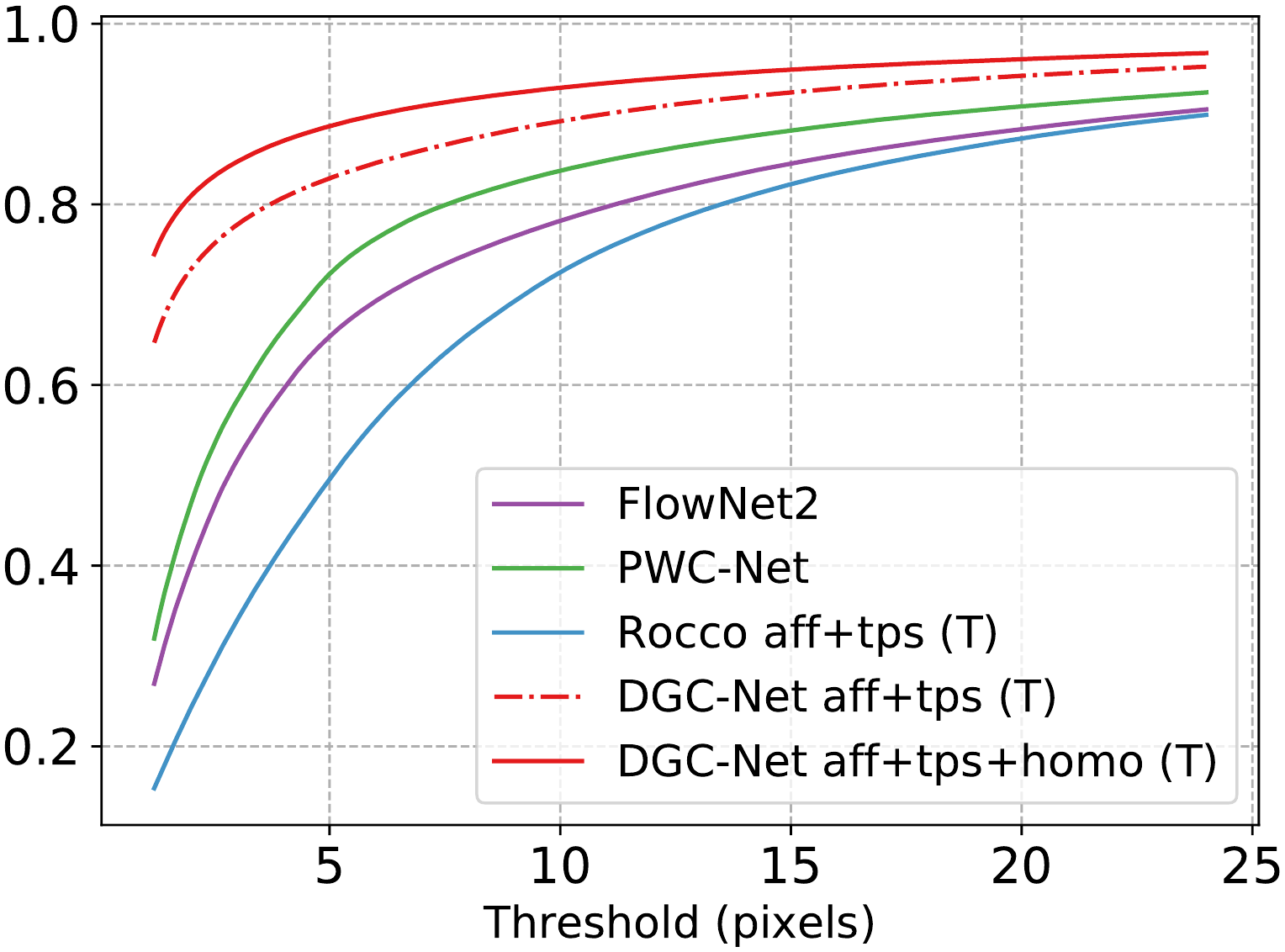}
 		\caption{Viewpoint III}
 	\end{subfigure}%
 	~
 	\begin{subfigure}[t]{.2\textwidth}
 		\centering
 		\includegraphics[width=\textwidth]{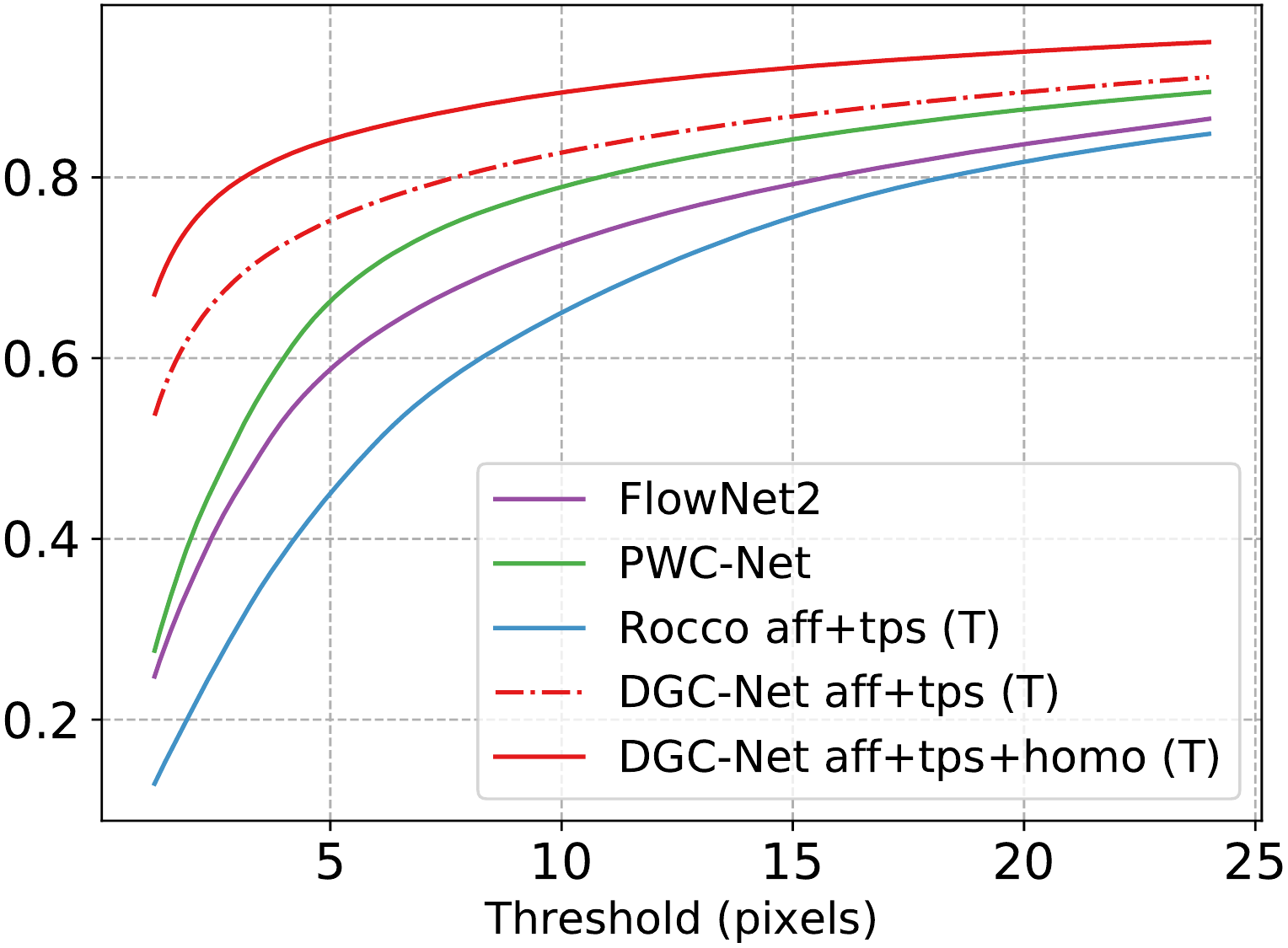}
 		\caption{Viewpoint IV}
 	\end{subfigure}%
 	~
 	\begin{subfigure}[t]{.2\textwidth}
 		\centering
 		\includegraphics[width=\textwidth]{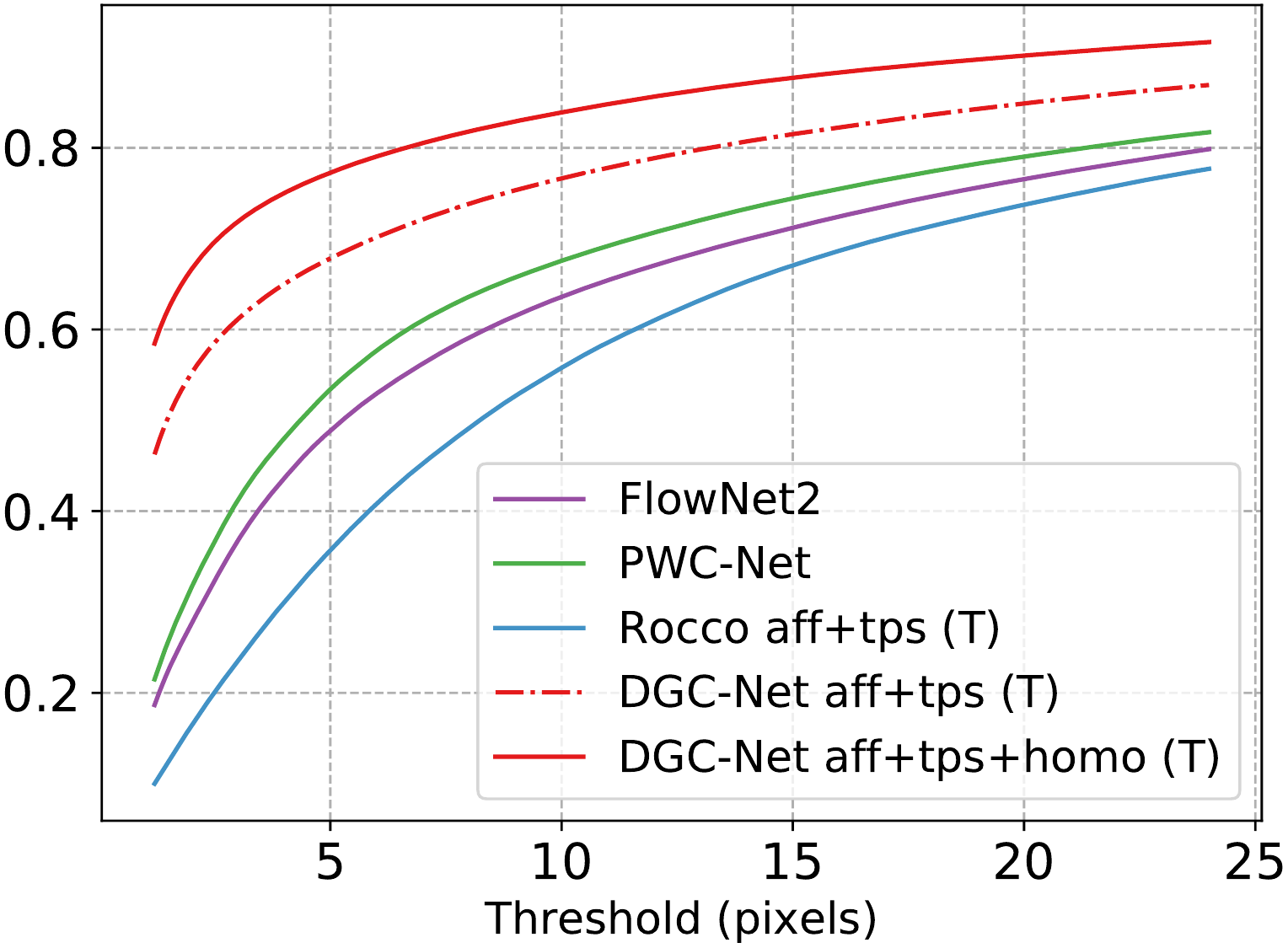}
 		\caption{Viewpoint V}%
 	\end{subfigure}
\caption{PCK metric calculated for different Viepoint IDs of the HPatches dataset. The proposed DGC-Net model outperforms all the baseline methods with a large margin.}\label{fig:pck_hpatches}
\vspace{-2mm}
\end{figure*}

\vspace{-2mm}
\section{Experiments}
\vspace{-1mm}
We discuss the experimental settings and evaluate the proposed method on two closely related tasks,~\ie finding correspondences between images and relative camera pose estimation. 

\subsection{Baselines}
\vspace{-1mm}
In this work we compare our approach with several strong baselines.

\noindent\textbf{Image alignment}. Rocco~\etal~\cite{Rocco17} propose a CNN-based method to estimate geometric transformations between two images achieving state-of-the art results in a semantic correspondence task. The transformations are parameterized as a 18-element vector and directly regressed by the network. We apply the estimates to a regular grid of the size of the input images to produce a dense pixel correspondence map.

\noindent\textbf{Optical flow} estimation requires finding correspondences between two input images. Therefore, we consider three CNN-based optical flow approaches,~\ie SPyNet~\cite{Spynet}, FlowNet2~\cite{FlowNet2} and the recently proposed PWC-Net~\cite{PWC-Net} as baseline methods. In detail, PWC-Net is based on a coarse-to-fine paradigm and predicts optical flow at different scales of feature maps produced by a Siamese CNN. The coarse estimates are then used to refine the flow. For optical flow methods we use pre-trained models from the original authors.

\noindent\textbf{DeepMatching}~\cite{DeepMatching} is matching algorithm aiming at finding semi-dense image correspondences. Specifically, it relies on a multi-scale image pyramid architecture with no any trainable parts and can cope with very challenging scenes, such as repetitive textures and non-rigid image transformations.

\vspace{-2mm}
\subsection{Datasets}\label{ssec:datasets}
\vspace{-1mm}
We compare the proposed approach with different baseline methods on two \textit{evaluation datasets}.

\noindent\textbf{HPatches}~\cite{HPatches} consists of several sequences of real images with varying photometric and geometric changes. Each image sequence contains a reference (target) image and 5 source images taken under a different viewpoint. For all images the estimated ground truth homography $\mathbf{H}$ is provided, thus, dense correspondence maps can be obtained for each test image pair. There are 59 image sequences with challenging geometric transformations in total. 

\noindent\textbf{DTU.} The pixel correspondences produced by our method can be also used for relative camera pose estimation problem. Thus, in order to measure the performance of the proposed approach for this task, we utilize the DTU image dataset~\cite{DTU} consisting of 124 scenes with very accurate absolute camera poses collected by a precisely positioned robot. We create a list of camera pairs which have overlapping fields of view and then randomly choose about $3k$ image pairs covering all the scenes.

\noindent\textbf{Training datasets.} We use training and validation splits proposed by~\cite{Rocco17} to compare both approaches fairly. Specifically, Rocco~\etal ~\cite{Rocco17} generate synthetic affine (aff) and thin-plate spline (TPS) transformations and apply them to images from Pascal VOC 2011 $(P)$ and Tokyo Time Machine $(T)$ datasets. Each synthetic dataset has 20k training and validation image pairs, respectively. However, those transformations are not very diverse. 
To be able to estimate the correspondences for HPatches scenes accurately, we therefore generate 20k labeled training examples~\cite{DeepHomography} by applying random homography transformations to the $(T)$ dataset. 
All training datasets mentioned above represent only synthetic geometric transformations between images. However, it is hard to artificially generate such diverse transformations that are present in real 3D world. Therefore, in addition to synthetic data, we utilize the \textbf{Citywall} dataset used for 3D reconstruction and provided by~\cite{MVE}. Based on camera poses and depth maps estimated with the Multiview Reconstruction Environment~\cite{MVE}, we create a list of 10k image pairs and ground truth correspondence maps. We use this data to \textit{fine-tune} the proposed model. We emphasize that the objective of this experiment is to demonstrate that fine-tuning on realistic data leads to further improvement of the results.

\vspace{-1mm}
\subsection{Metrics}
As predicting a dense corresponding grid is closely related to optical flow estimation, we follow the standard evaluation metric used in this task, \ie the average endpoint error (AEPE). AEPE is  defined as the average Euclidean distance between the estimated and ground truth correspondence map. In addition to AEPE, we also use Percentage of Correct Keypoints (PCK) as the evaluation metric. PCK shows the percentage of the correctly matched estimated points $\hat{x}_i$ that are within a certain threshold (in pixels) from the ground truth corresponding points ${x}_i$.

In order to estimate the accuracy of matchability mask predictions, we report normalized Jaccard index (Intersection Over Union, IoU),~\ie $0 \leq J \leq 1$ for the ground truth and estimated masks. This metric is interpreted as a similarity measure between two finite sample sets and widely used in semantic segmentation~\cite{ods_segm}.

\vspace{-1mm}
\subsection{Results}\label{ssec:results}
\noindent\textbf{Synthetic Datasets.} First, we experimentally compare the proposed DGC-Net and DGC-Net+M models with~\cite{Rocco17} by calculating AEPE. All the models have been trained on the data provided by~\cite{Rocco17}. More specifically, *-aff methods utilize only synthetic affine transformations during training but *-aff+tps methods additionally trained on TPS transformations. AEPE is measured only for valid pixel locations of $(P)$ and $(T)$ test data by applying the ground-truth mask. For DGC-Net+M-* models we also report normalized Jaccard index. Tab.~\ref{tbl:perf_rocco} shows that DGC-Net significantly outperforms all baseline methods on both evaluation datasets. Despite the fact that DGC-Net+M model is marginally worse than DGC-Net in the case that the transformation between images can be described by an affine transformation, it is more universal approach as it additionally predicts a matchability map which is quite accurate according to the Jacard similarity score. It is worth noting that the proposed models generalize well to unseen data, since AEPE metric varies slightly for $(P)$ and $(T)$ evaluation datasets respectively. It shows that the model has learned the geometric transformations and not overfitting to the visual representation of images.

\begin{table}[t!]
\begin{center}
\resizebox{.49\textwidth}{!}{%
    \begin{tabular}{l l| p{15mm} p{15mm} | p{15mm} p{10mm}}
        \multirow{2}{*}{Method} & \multicolumn{1}{r|}{Train:} & \multicolumn{2}{c|}{Pascal-voc11 (P)} & \multicolumn{2}{c}{Tokyo Time Machine (T)} \\
        \cline{2-6}
        & \multicolumn{1}{r|}{Test:} & (P)-\textit{aff} & (T)-\textit{aff} & (T)-\textit{aff} & (P)-\textit{aff}\\
        \hline
        \hline
        \multicolumn{2}{l|}{Rocco\cite{Rocco17} aff} & 3.82 & 3.93 & 4.10 & 4.45 \\
        \multicolumn{2}{l|}{DGC-Net+M aff} & 0.92/0.847 & 0.97/0.847 & 1.03/0.848 & 1.14/0.848 \\
        \multicolumn{2}{l|}{DGC-Net aff} & 0.95 & 0.99 & 0.90 & 1.03 \\
        \hline
        \multicolumn{2}{l|}{Rocco\cite{Rocco17} aff+tps} & 3.28 & 3.30 & 4.83 & 4.97 \\
        \multicolumn{2}{l|}{DGC-Net+M aff+tps} & 0.82/0.849 & 0.96/0.849 & 0.83/0.853 & 0.92/0.853 \\
        \multicolumn{2}{l|}{DGC-Net aff+tps} & \textbf{0.57} & \textbf{0.69} & \textbf{0.54} & \textbf{0.61} \\
        \hline
\end{tabular}
}
\end{center}
\caption{AEPE metric on the data from~\cite{Rocco17}. For DGC-Net+M models, the Jaccard index is also  reported.}\label{tbl:perf_rocco}
\vspace{-2mm}
\end{table}

\noindent\textbf{Realistic Datasets.} To demonstrate the performance on more realistic data, we evaluate all baseline methods and our approach on the HPatches dataset. That is, we calculate AEPE over all image sequences belonging to the same viewpoint ID and report the numbers in Tab.~\ref{tbl:perf_hpatches}. Compared to *-aff models, fine-tuning on TPS transformations lead to a significant improvement in the performance reducing the overall EPE by 20\% for Viewpoint II and by 9\% for Viewpoint V, respectively. The performance is improved further by finetuning the model on synthetic homography data. To prevent large errors caused by interpolation, we directly calculate AEPE metric for the semi-dense DeepMatching~\cite{DeepMatching} estimates (\ie hence~\cite{DeepMatching} has unfair advantage in terms of AEPE).
The Jaccard index for DGC+M-Net-* models is provided in Tab.~\ref{tbl:jaccard_hpatches}.

In addition, we report a number of correctly matched pixels between two images by calculating PCK metric with different thresholds. Especially the comparison with~\cite{Rocco17} is interesting as the coarse level of our pipeline is based on its matching strategy. As shown in Fig.~\ref{fig:pck_hpatches}, the proposed method correctly matches around 85\% pixels for the case where geometric transformations are quite small (Viewpoint I). It significantly outperforms~\cite{Rocco17} trained on the same data without any external synthetic datasets and can be further improved by utilizing more diverse transformations during training. Compared to FlowNet2 and PWC-Net, DGC-Net, our method can handle scenarios exhibiting drastic changes between views (Viewpoint IV and V), achieving 59\% of PCK with a 1-pixel threshold for the most challenging case.

\begin{table}[t!]
\begin{center}
\resizebox{.49\textwidth}{!}{%
    \begin{tabular}{l|l l l l l}
        \multirow{2}{*}{Method} & \multicolumn{5}{|c}{Viewpoint ID} \\
        & \multicolumn{1}{|c}{I} & \multicolumn{1}{c}{II} & \multicolumn{1}{c}{III} & \multicolumn{1}{c}{IV} & \multicolumn{1}{c}{V}\\
    \hline
    \hline
    SPyNet~\cite{Spynet} & 36.94 & 50.92 & 54.29 & 62.60 & 72.57 \\
    DeepMatching*~\cite{DeepMatching} & \textit{5.84} & \textbf{\textit{4.63}} & \textit{12.43} & \textit{12.17} & \textit{22.55} \\
    FlowNet2~\cite{FlowNet2} & 5.99 & 15.55 & 17.09 & 22.13 & 30.68 \\
    PWC-Net~\cite{PWC-Net} & 4.43 & 11.44 & 15.47 & 20.17 & 28.30 \\
    \hline 
    Rocco~\cite{Rocco17} aff (P) & 14.85 & 29.09 & 31.04 & 39.35 & 45.92 \\
    DGC+M-Net aff (P) & 5.96 & 12.85 & 16.23 & 20.64 & 27.63 \\
    DGC-Net aff (P) & 6.80 & 13.82 & 17.15 & 22.62 & 28.39 \\
    \hline 
    Rocco~\cite{Rocco17} aff (T) & 15.02 & 28.23 & 29.27 & 36.57 & 43.68 \\
    DGC+M-Net aff (T) & 6.22 & 14.46 & 17.21 & 22.92 & 29.65 \\
    DGC-Net aff (T) & 5.12 & 13.01 & 15.08 & 20.14 & 26.47 \\
    \hline 
    Rocco~\cite{Rocco17} aff+tps (P) & 9.50 & 22.47 & 24.73 & 34.20 & 41.46 \\ 
    DGC+M-Net aff+tps (P)  & 4.35 & 11.17 & 14.09 & 18.66 & 25.04 \\
    DGC-Net aff+tps (P) & 4.20 & 10.78 & 14.34 & 18.48 & 25.00 \\
    \hline
    Rocco~\cite{Rocco17} aff+tps (T) & 9.59 & 18.55 & 21.15 & 27.83 & 35.19 \\
    DGC+M-Net aff+tps (T) & 4.40 & 8.92 & 11.94 & 16.33 & 22.01 \\
    DGC-Net aff+tps (T) & 3.10 & 8.18 & 10.97 & 16.29 & 22.29 \\
    \hline
    DGC+M-Net aff+tps+homo (T) & 2.97 & 6.85 & 9.95 & 12.87 & 19.13 \\
    DGC-Net aff+tps+homo (T) & \textbf{1.55} & 5.53 & \textbf{8.98} & \textbf{11.66} & \textbf{16.70} \\
    \hline
\end{tabular}
}
\end{center}
\vspace{-2mm}
\caption{AEPE metric for different viewpoint IDs of the HPatches dataset (lower is better). The training datasets for our and~\cite{Rocco17}'s approaches are given in parentheses. *Note that since~\cite{DeepMatching} produces only semi-dense matches it has unfair advantage in terms of AEPE.}\label{tbl:perf_hpatches}
\end{table}

\begin{table}[t!]
\begin{center}
\resizebox{.40\textwidth}{!}{%
    \begin{tabular}{l|l l l l l}
        \multirow{2}{*}{Transformation} & \multicolumn{5}{|c}{Viewpoint ID} \\
        & \multicolumn{1}{|c}{I} & \multicolumn{1}{c}{II} & \multicolumn{1}{c}{III} & \multicolumn{1}{c}{IV} & \multicolumn{1}{c}{V}\\
    \hline
    \hline
    aff          & 0.682 & 0.617 & 0.562 & 0.523 & 0.445  \\
    aff+tps      & 0.700 & 0.650 & 0.603 & 0.573 & 0.496  \\
    aff+tps+homo & \textbf{0.730} & \textbf{0.687} & \textbf{0.629} & \textbf{0.590} & \textbf{0.525}  \\
    \hline
\end{tabular}
}
\end{center}
\vspace{-2mm}
\caption{Normalized Jaccard index (higher is better) produced by the~\texttt{DGC+M-Net} model on HPatches evaluation dataset with different types of synthetic transformations of ($T$) training dataset.}\label{tbl:jaccard_hpatches}
\end{table}

Qualitative results on HPatches and DTU are illustrated in Fig.~\ref{fig:comparison_hpatches_qual} and Fig.~\ref{fig:comparison_dtu_qual}, respectively.

\begin{figure*}[t!]
 	\centering
 	\begin{subfigure}[t]{.3\textwidth}
 		\centering
 		\includegraphics[width=\textwidth]{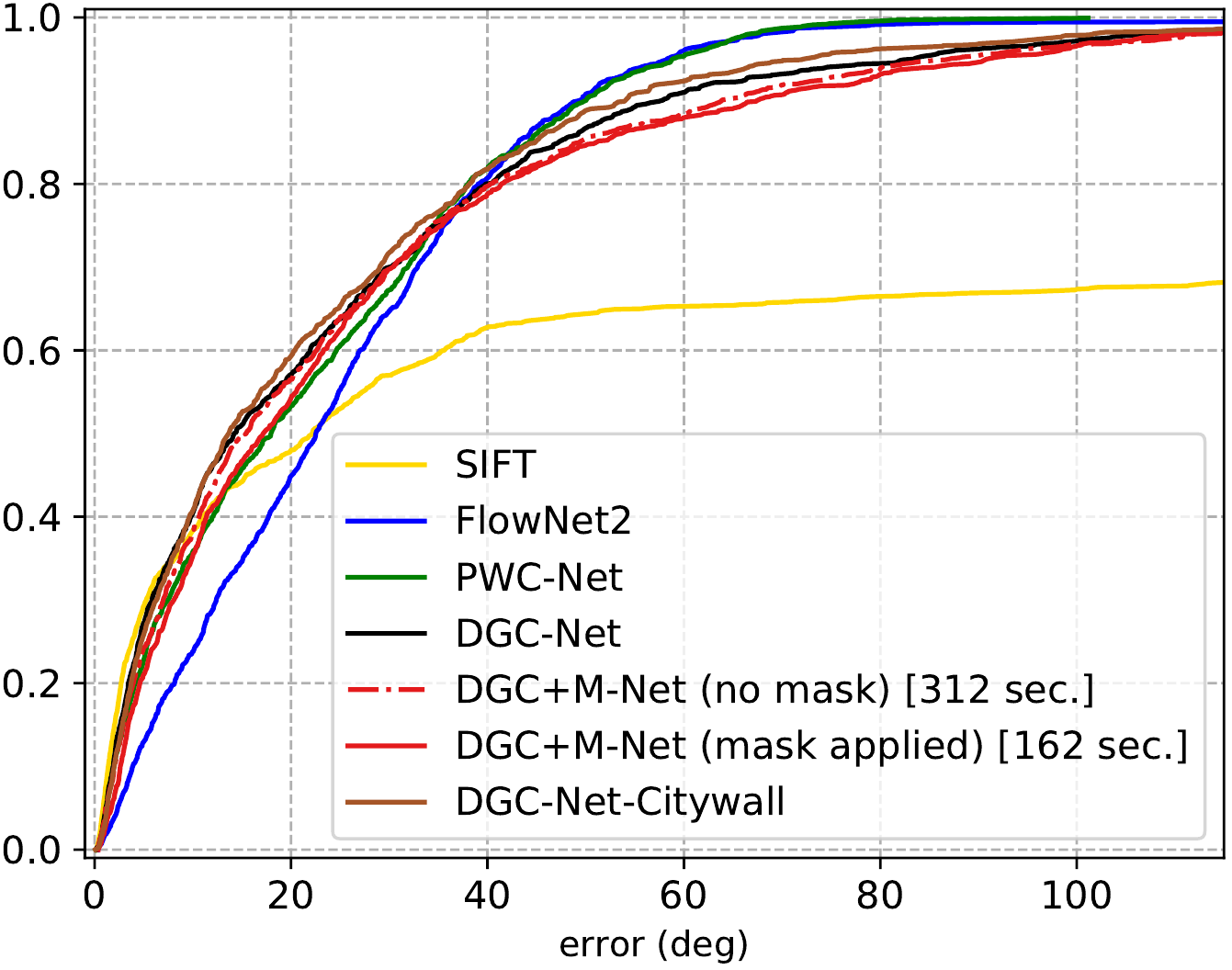}
 		\caption{Relative orientation error}\label{fig:relative_pose_dtu_orient}
 	\end{subfigure}%
 	~
 	\begin{subfigure}[t]{.3\textwidth}
 		\centering
 		\includegraphics[width=\textwidth]{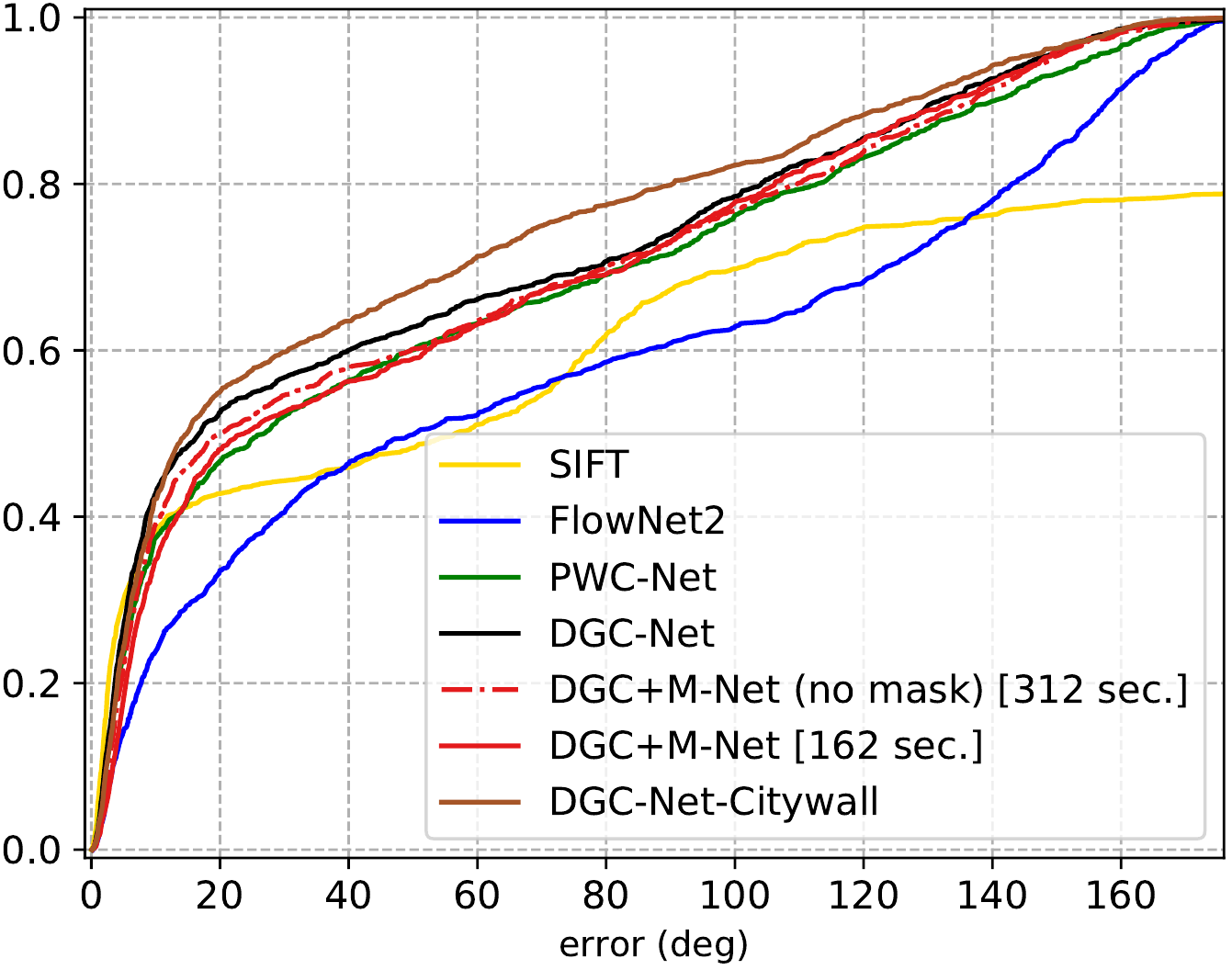}
 		\caption{Relative translation error}\label{fig:relative_pose_dtu_transl}
 	\end{subfigure}%
 	~
 	\begin{subfigure}[t]{.3\textwidth}
 		\centering
 		\includegraphics[width=\textwidth]{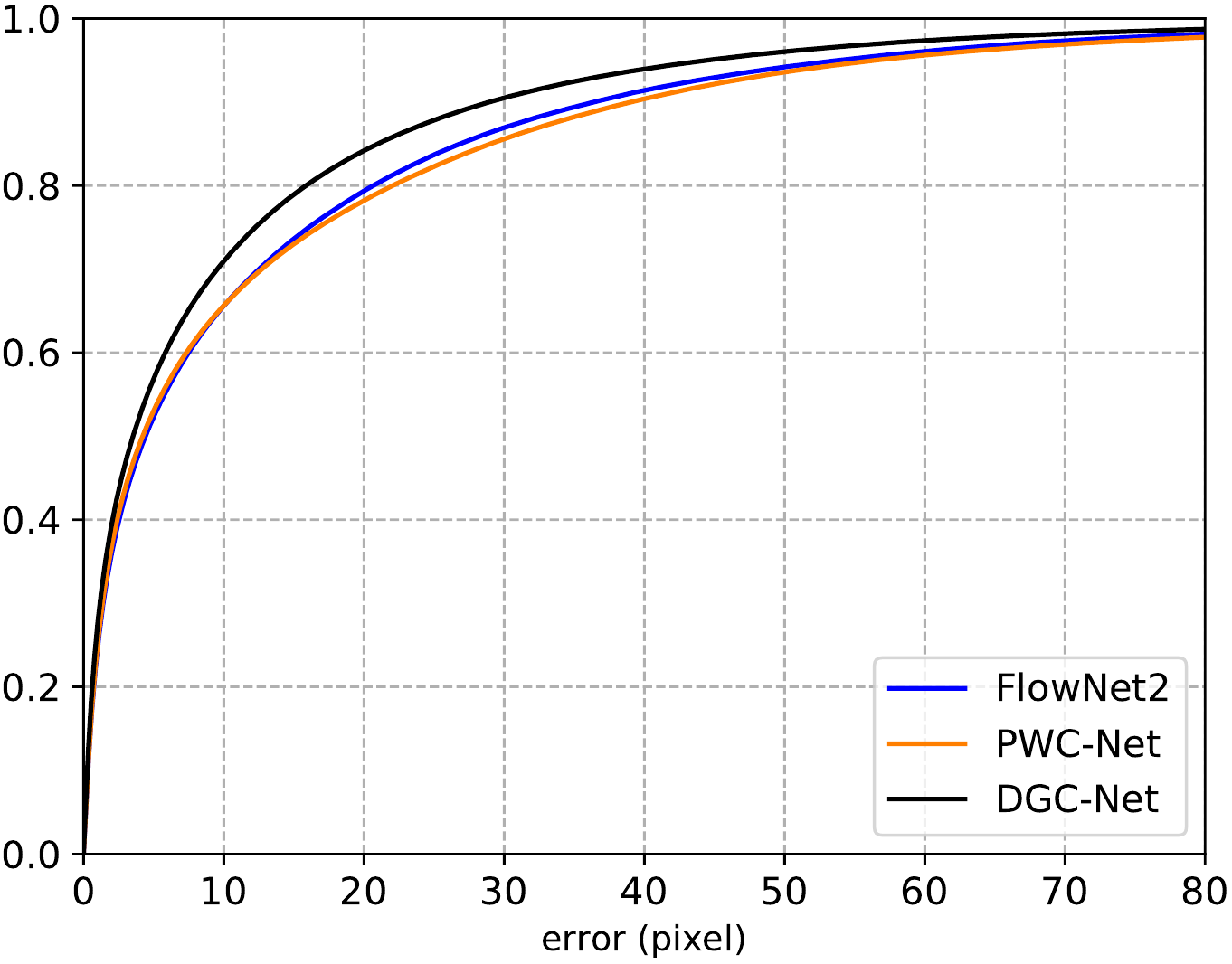}
 		\caption{Symmetric epipolar line distance error}\label{fig:f_epipolar_error}
 	\end{subfigure}
\caption{Comparison of the proposed approach with different baseline methods on the DTU dataset.}\label{fig:relative_pose_dtu}
\end{figure*}

\noindent\textbf{Relative camera pose.} In this section, we demonstrate the application of the proposed method for predicting relative camera pose. Given a list of correspondences and the intrinsic camera parameters matrix $\mathbf{K}$, we estimate the essential matrix~$\mathbf{E}$ by applying RANSAC. To decrease the randomness of RANSAC, for each image pair we run a 1000-iteration loop for 5 times and choose the estimated essential matrix corresponding to the maximum inliers count. Once this process is predicted, relative pose can be recovered based on~$\mathbf{E}$ and~$\mathbf{K}$ respectively. Similarly to~\cite{RelativePoseCNN}, we use the relative orientation error and the relative translation error as metrics for evaluating  the performance. Both metrics compute the angle between the estimated orientation/translation and the ground truth. Fig.~\ref{fig:relative_pose_dtu_orient} and~\ref{fig:relative_pose_dtu_transl} show a set of normalized cumulative histograms of relative orientation and translation errors for each baseline models evaluated on all scenes of the DTU dataset (Sec.~\ref{ssec:datasets}). As before, DGC-Net and DGC+M-Net have been trained on only synthetic transformations (aff+tps+homo). For a fair comparison, we resize images to $256 \times 256$ size for all baseline methods and change internal camera parameters accordingly. Interestingly, both PWC-Net~\cite{PWC-Net} and FlowNet2~\cite{FlowNet2} estimate relative orientation quite well achieving $20^{\circ}$ and $24^{\circ}$ median error calculated at level $0.5$, respectively. The proposed approach outperforms all CNN-based baselines by 18\% and 40\% at estimating relative orientation and translation median error compared to PWC-Net. 
We also evaluate DGC+M-Net model which additionally predicts a matchability mask. This mask can be considered as a filter to remove tentative correspondences with small confidence score from the relative pose estimation pipeline. According to Fig.~\ref{fig:relative_pose_dtu}, DGC+M-Net falls slightly behind of DGC-Net in estimating relative pose but it achieves significant advantages in terms of computational efficiency decreasing the elapsed time from 312 sec. to 162 sec. for estimating relative camera pose for all test image pairs. 

To experiment with more realistic transformations, we fine-tune DGC-Net model on the Citywall dataset (Sec.~\ref{ssec:datasets}), illustrated in the supplementary material. 
We refer to this model as~\texttt{DGC-Net-Citywall}. As can be clearly seen, ground-truth transformation maps are incomplete leading to multiple missing regions in the warped reference images (see the supplementary). However, using external data with more diverse transformations helps to improve the performance of the method remarkably, decreasing the median relative translation error by 17\% according to Fig.~\ref{fig:relative_pose_dtu_transl}. 


In addition, we calculate the epipolar error for the matches produced by our method, PWC-Net and FlowNet2. The error is defined in terms of the squared distances ($d^2$) between the points and corresponding epipolar lines as follows:
\begin{equation}\label{eq:eq_epipolar_err}
    \mathcal{D}_{e} = \sqrt{\frac{d^2\left(\mathbf{x}'_i,\mathbf{F}\mathbf{x}_i\right) + d^2\left(\mathbf{x}_i,\mathbf{F}^{T}\mathbf{x'}_i\right)}{2}}, \forall i \in N,
\end{equation}

\noindent where $\mathbf{x'}_i=(x'_{i}, y'_i, 1)^{T}$ and $\mathbf{x}_i=(x_{i}, y_{i}, 1)^{T}$ denote a pair of matching points in two images;  $\mathbf{F}$ is the ground-truth fundamental matrix between two views; $N$ is the number of image pixels (image resolution). 
The normalized cumulative histogram of the error is presented in Fig.~\ref{fig:f_epipolar_error}. Quantitatively, the proposed method provides quite accurate pixel correspondences between two views achieving a median error less than 4 pixels across the whole test dataset.

\vspace{-1mm}
\subsection{Ablation Study}\label{ssec:ablation_study}
In this section, we analyze some design decisions of the proposed approach. More specifically, our goal is to investigate the benefits of using global correlation layer compared to the one utilized in recent optical flow methods~\cite{FlowNet2,PWC-Net}. In addition, we experiment with another type of parametrization of ground truth data by representing a correspondence map as flow. Furthermore, we demonstrate the importance of $L2$ normalization of the correlation map. The results are presented in Tab.~\ref{tbl:ablation_study}.

\begin{table}[h!]
    \begin{subtable}[h]{0.49\textwidth}
        \centering
        \resizebox{\textwidth}{!}{%
            \begin{tabular}{l l| p{15mm} p{15mm} | p{15mm} p{10mm}}
                \multirow{2}{*}{Method} & \multicolumn{1}{r|}{Train:} & \multicolumn{2}{c|}{Pascal-voc11 (P)} & \multicolumn{2}{c}{Tokyo Time Machine (T)} \\
            \cline{2-6}
                & \multicolumn{1}{r|}{Test:} & (P)-\textit{aff} & (T)-\textit{aff} & (T)-\textit{aff} & (P)-\textit{aff}\\
            \hline
            \hline
            \multicolumn{2}{l|}{Rocco\cite{Rocco17}} & 3.82 & 3.93 & 4.10 & 4.45 \\
            \multicolumn{2}{l|}{DGC-Net} & 0.95 & 0.99 & 0.90 & 1.03 \\
            \multicolumn{2}{l|}{DGC-Net-flow} & 0.96 & 0.99 & 0.84 & 0.94 \\
            \multicolumn{2}{l|}{PWCm-Net} & 6.73  & 6.87 & 6.31  & 6.82 \\
            \hline
            \multicolumn{2}{l|}{DGC-Net no L2norm} & 1.12 & 1.15 & 1.01 & 1.36 \\
            \multicolumn{2}{l|}{PWCm-Net no L2norm} & 6.92 & 7.35 & 6.55 & 7.11\\

            \end{tabular}
        }
        \vspace{-1mm}
    \caption{Performance of synthetic data. All the models trained on synthetic \textit{affine} transformations provided by~\cite{Rocco17}.}\label{tbl:as_rocco_data}    
    \end{subtable}
        
    \begin{subtable}[h]{.49\textwidth}
        \centering 
        \resizebox{\textwidth}{!}{%
            \begin{tabular}{l|l l l l l}
                \multirow{2}{*}{Method} & \multicolumn{5}{|c}{Viewpoint ID} \\
                & \multicolumn{1}{|c}{I} & \multicolumn{1}{c}{II} & \multicolumn{1}{c}{III} & \multicolumn{1}{c}{IV} & \multicolumn{1}{c}{V}\\
            \hline
            \hline 
            DGC-Net aff no L2norm  & 6.18 & 15.81 & 19.26 & 25.64 & 33.82 \\
            DGC-Net-flow aff & 5.09 & 13.14 & 15.76 & 22.21 & 29.64 \\ 
            DGC-Net aff  & 5.12 & 13.01 & 15.08 & 20.14 & 26.47 \\
            \hline 
            DGC-Net-flow aff+tps & 4.03 & 10.95 & 13.57 & 20.52 & 27.30 \\
            DGC-Net aff+tps  & 3.10 & 8.18 & 10.97 & 16.29 & 22.29 \\
            \hline 
            DGC-Net-flow aff+tps+homo & 2.93 & 7.06 & 9.76 & 12.52 & 18.63 \\
            DGC-Net aff+tps+homo & \textbf{1.55} & \textbf{5.53} & \textbf{8.98} & \textbf{11.66} & \textbf{16.70} \\
            \hline
            \end{tabular}
        }
        \vspace{-1mm}
        \caption{AEPE on HPatches.}\label{tbl:as_hpatches_data}
    \end{subtable}
    
    \begin{subtable}[h]{.49\textwidth}
    \centering
    \resizebox{\textwidth}{!}{%
        \begin{tabular}{l|c c c }
            \multirow{2}{*}{} & \multicolumn{3}{|c}{Model} \\
            & \multicolumn{1}{|c}{DGC-Net} & \multicolumn{1}{c}{PWC-Net~\cite{PWC-Net}} & \multicolumn{1}{c}{PWCm-Net}\\
        \hline
        \hline
        CNN feature pyramid encoder  & fixed & updating & fixed  \\
        Training data & synthetic data &
        Optical Flow datasets & synthetic data \\
        Number of pyramid layers  & 5 & 7 & 5  \\
        Correlation layer     & global  & spatially constrained & spatially constrained \\
        Interpolation         & bilinear  & transposed conv (upconv) & bilinear \\
        Activation              & ReLU & LeakyReLU & ReLU \\
        Number of learnable params. & $2.7M$ & $9.4M$ & $2.7M$ \\
        \hline
        \end{tabular}
        }
        \vspace{-2mm}
        \caption{Comparison of DGC-Net and off-the-shelf PWC-Net architecture.}\label{tbl:as_pwc_difference}
        \end{subtable}
        \vspace{-2mm}
        \caption{\textbf{Ablation study.} We analyze the influence of different design choices of the proposed method. See Sec.~\ref{ssec:ablation_study} for more details. 
        } \label{tbl:ablation_study}
\end{table}

\noindent\textbf{Global correlation layer:} 
In contrast to the proposed approach, the PWC-Net architecture comprises a local correlation layer computing similarities between two feature maps in some restricted area around the center pixel at each level of the feature pyramid. However, it is very hard to compare DGC-Net and off-the-shelf PWC-Net approach fairly due to the significant difference in network structures (see Tab.~\ref{tbl:as_pwc_difference}). Therefore, we construct a new coarse-to-fine $N$-level CNN model by keeping all the blocks of DGC-Net except the correlation layer. More specifically, each feature pyramid level is complemented by a local correlation layer as it is used in PWC-Net structure. We dubbed this model to~\texttt{PWCm-Net}. As shown in Tab.~\ref{tbl:as_rocco_data}, the global correlation layer achieves a significant improvement over the case with a set of spatially constrained correlation layers. 
Particularly, the error is reduced from $6.73$ to $0.95$ pixels on the $(P)$ dataset. All results have been obtained for only affine transformations in training data.

\noindent\textbf{L2 normalization:} As explained in Sec.~\ref{ssec:net_arch}, we $L2$ normalize the output of the correlation layer to down-weigh the putative matches. In Tab.~\ref{tbl:as_rocco_data} we compare original DGC-Net model and its modified version without correlation layer normalization step (DGC-Net no L2norm). According to the results, the normalization improves the error by about 15\% for all test cases demonstrating the importance of this step.

\noindent\textbf{Different parametrization:} Given two images, the proposed approach predicts a dense pixel correspondence map representing the~\textit{absolute location} of each image pixel. In contrast, all optical flow methods estimate pixel ~\textit{displacements} between images. To dispel this doubt in parameterization, we train DGC-Net model on the same synthetic data as before but with ground-truth labels recalculated in an optical flow manner. We title this model~\texttt{DGC-Net-flow} and provide the results in Tab.~\ref{tbl:as_rocco_data} and Tab.~\ref{tbl:as_hpatches_data}. Interestingly, while DGC-Net-flow model marginally performs better on synthetic data, DGC-Net producing more accurate results in large geometric transformations case (Tab.~\ref{tbl:as_hpatches_data}) demonstrating the benefit of the original parametrization.
\vspace{-2mm}
\section{Conclusion}
Our paper addressed the challenging problem of finding dense pixel correspondences. We have proposed a coarse-to-fine network architecture that efficiently handles diverse transformations between two views. We have shown that our contributions were crucial to outperforming strong baselines on the challenging realistic datasets. Additionally, we have also applied the proposed method to the relative camera pose estimation problem, demonstrating very promising results. We hope this paper inspires more research into applying deep learning to accurate and reliable dense pixel correspondence estimation.

\begin{figure*}[t!]
	\centering
    \includegraphics[width=0.135\textwidth,height=0.12\textwidth]{}
    \includegraphics[width=0.135\textwidth,height=0.12\textwidth]{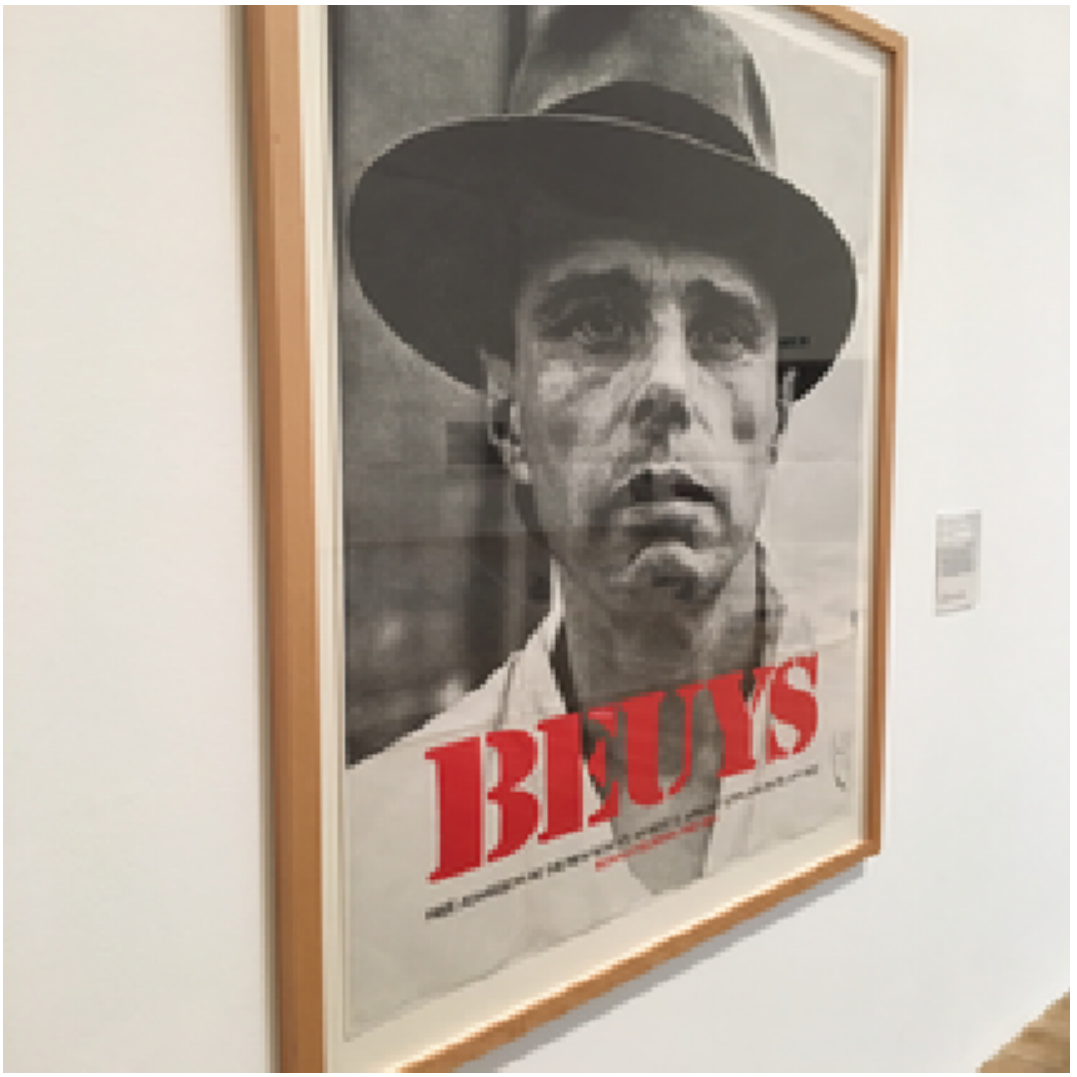}
    \includegraphics[width=0.135\textwidth,height=0.12\textwidth]{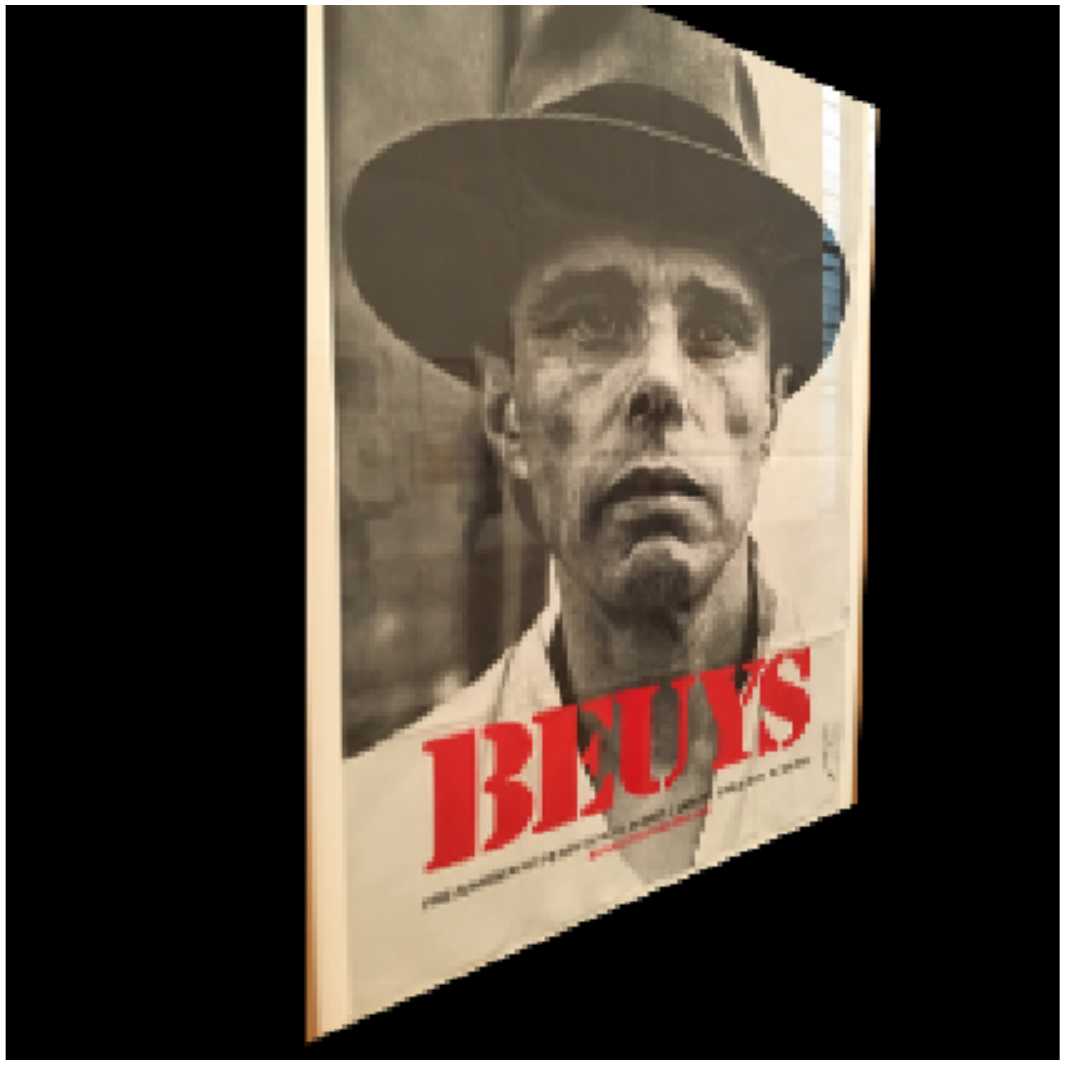}
    \includegraphics[width=0.135\textwidth,height=0.12\textwidth]{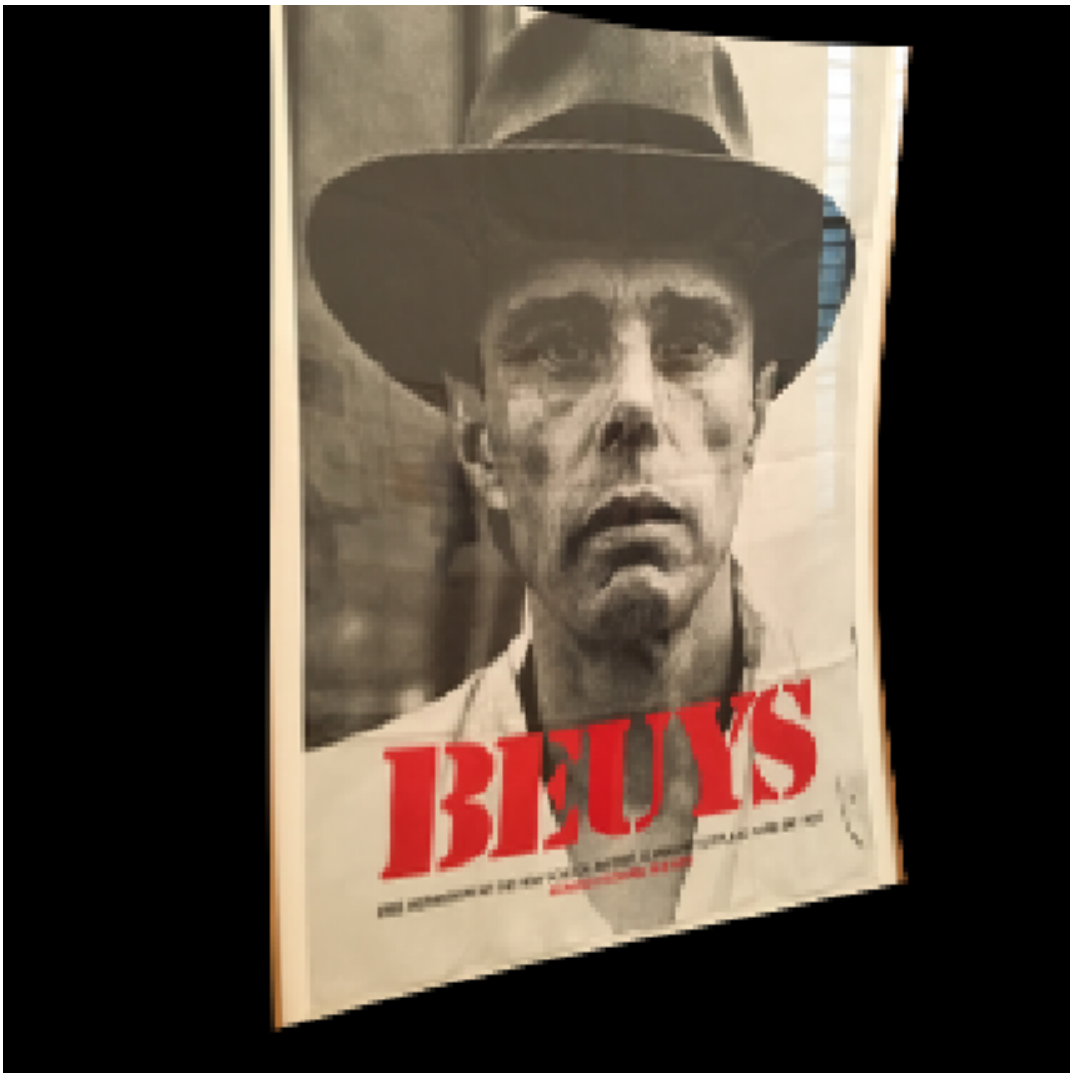}
    \includegraphics[width=0.135\textwidth,height=0.12\textwidth]{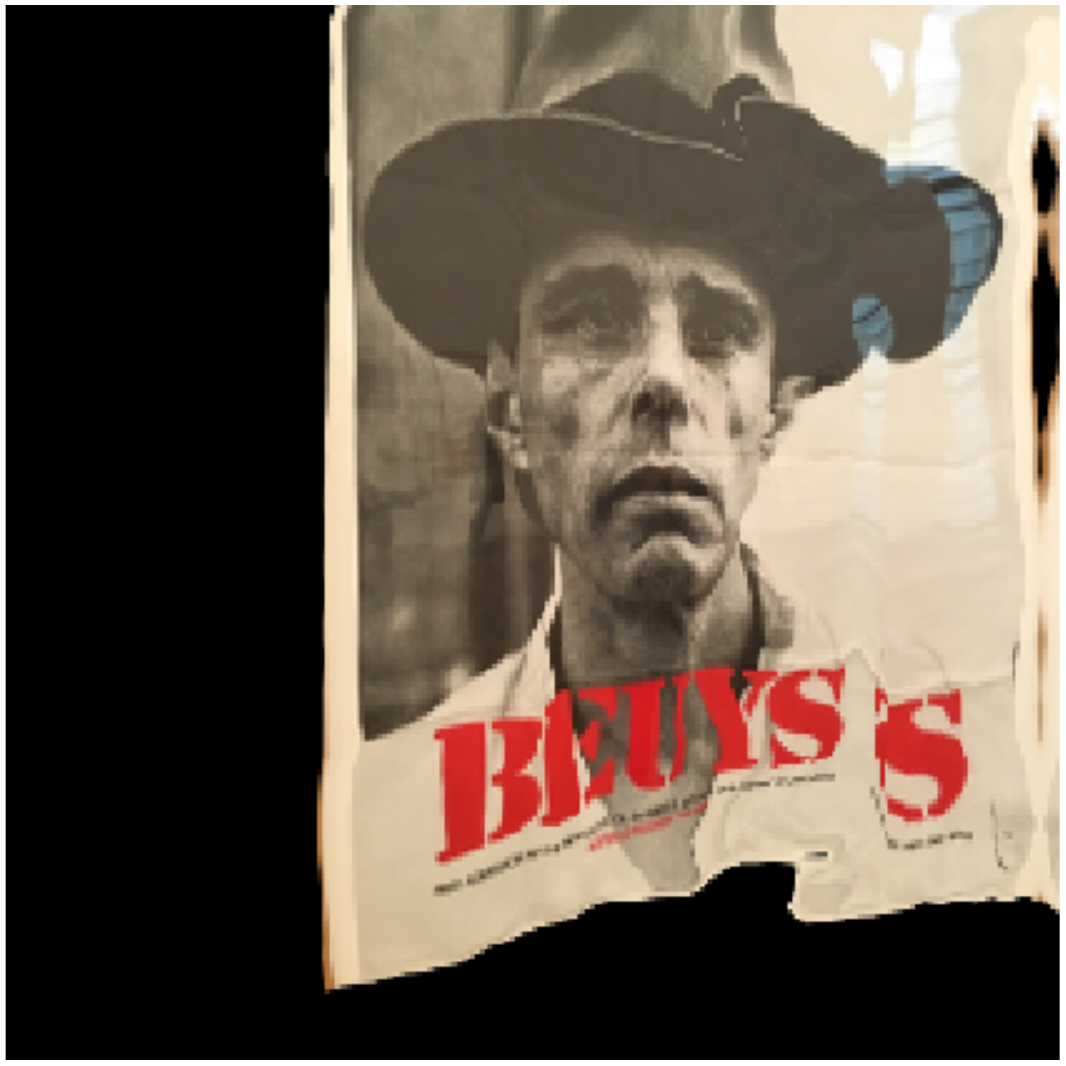}
    \includegraphics[width=0.135\textwidth,height=0.12\textwidth]{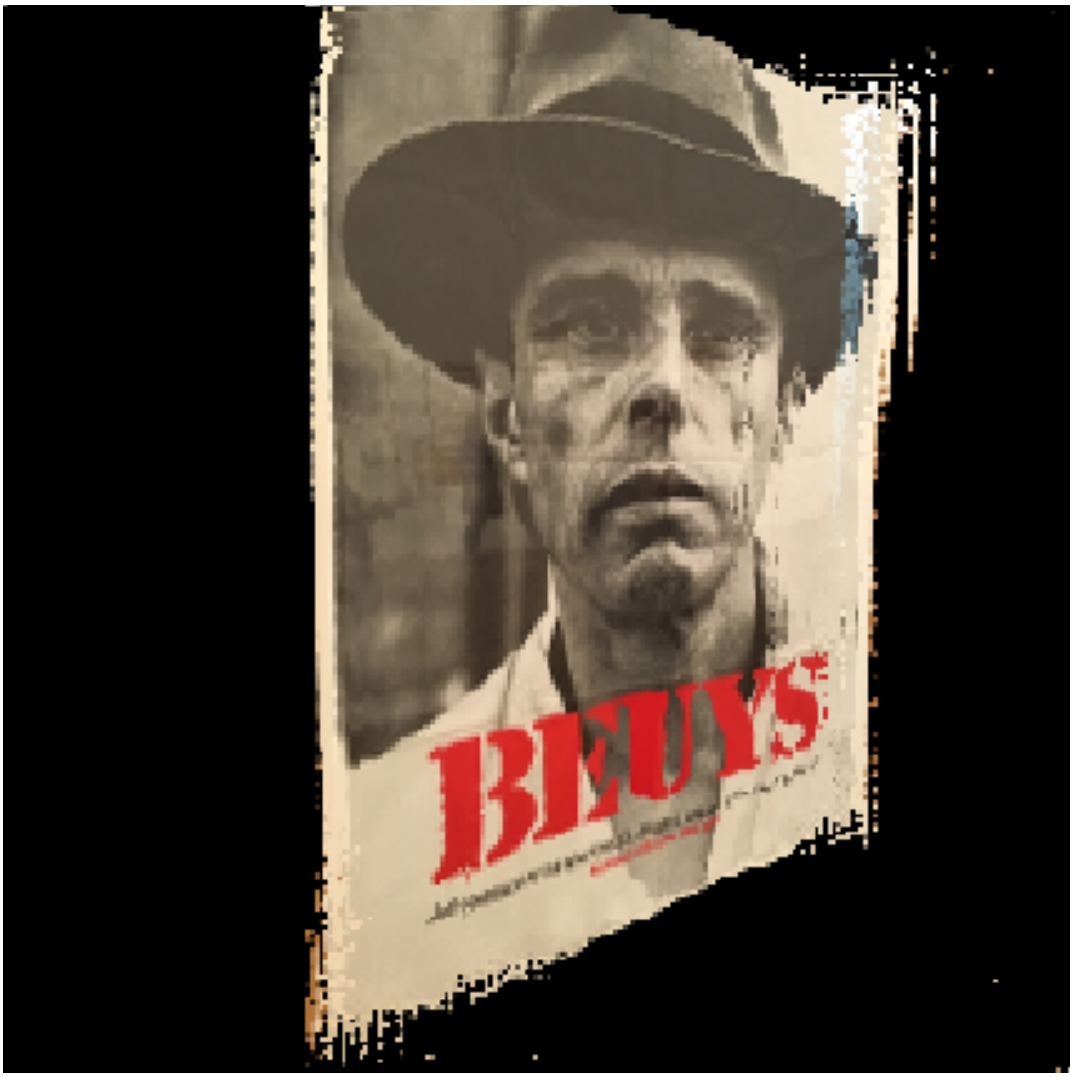}
    \includegraphics[width=0.135\textwidth,height=0.12\textwidth]{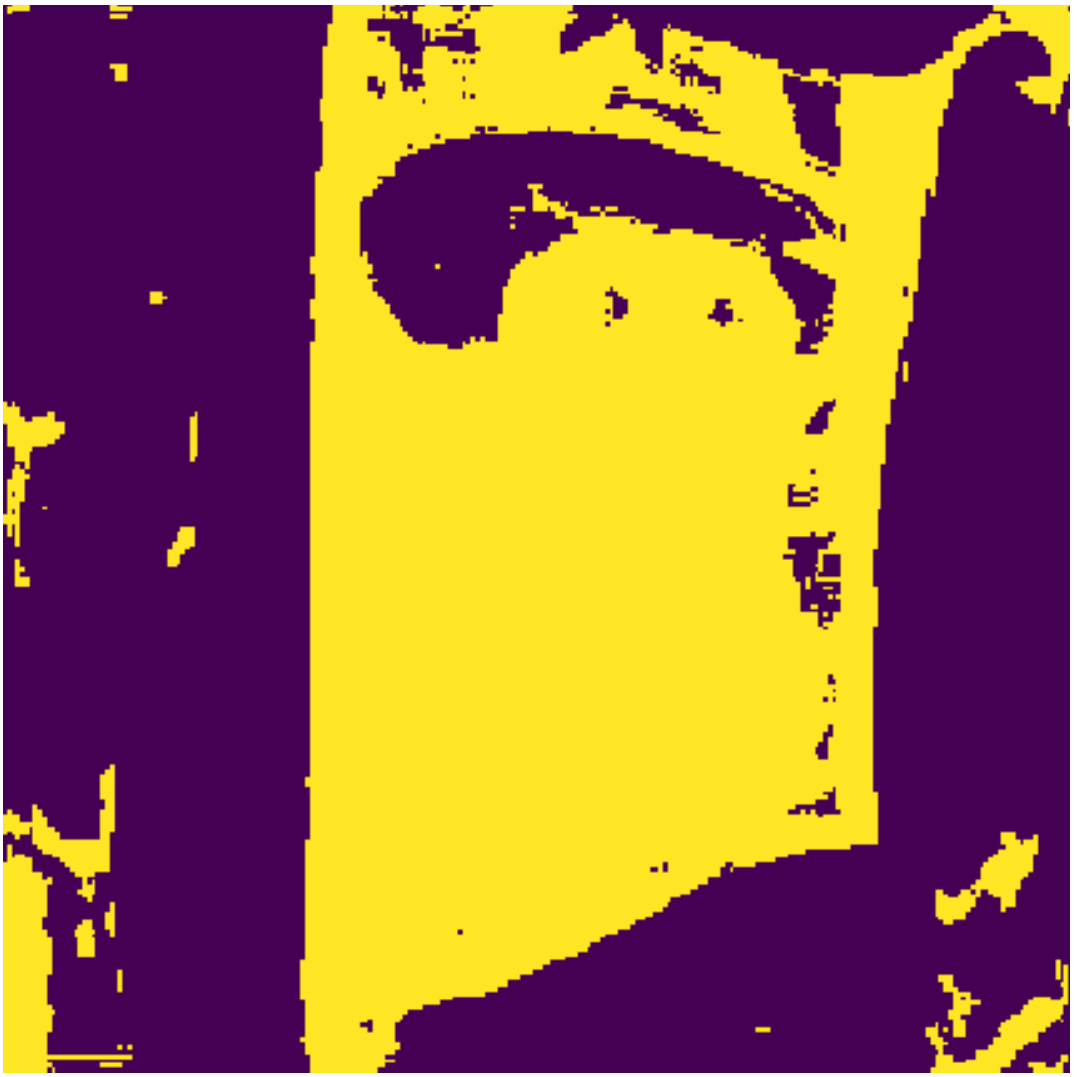}
    \vspace{0.5mm}
    \includegraphics[width=0.135\textwidth,height=0.12\textwidth]{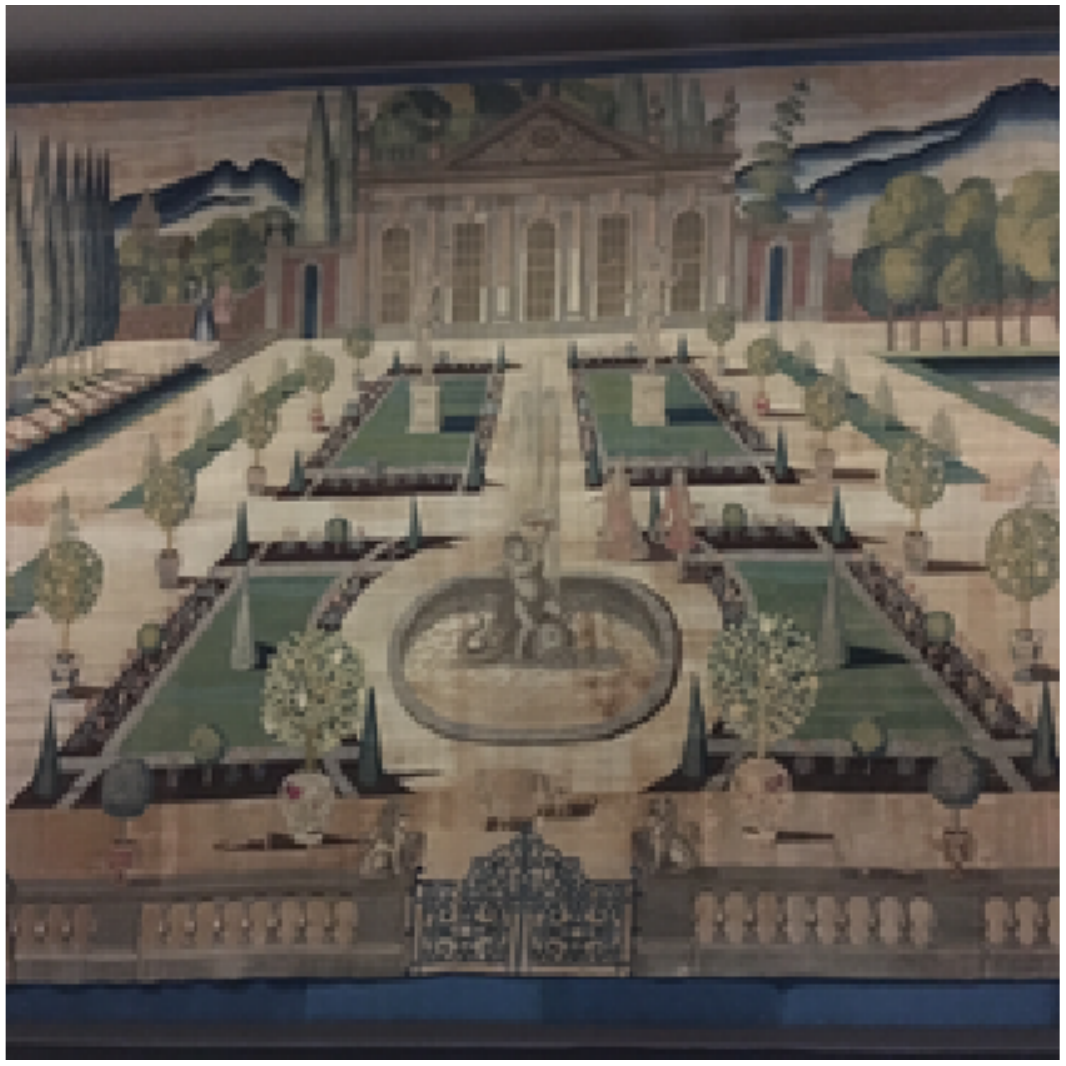}
    \includegraphics[width=0.135\textwidth,height=0.12\textwidth]{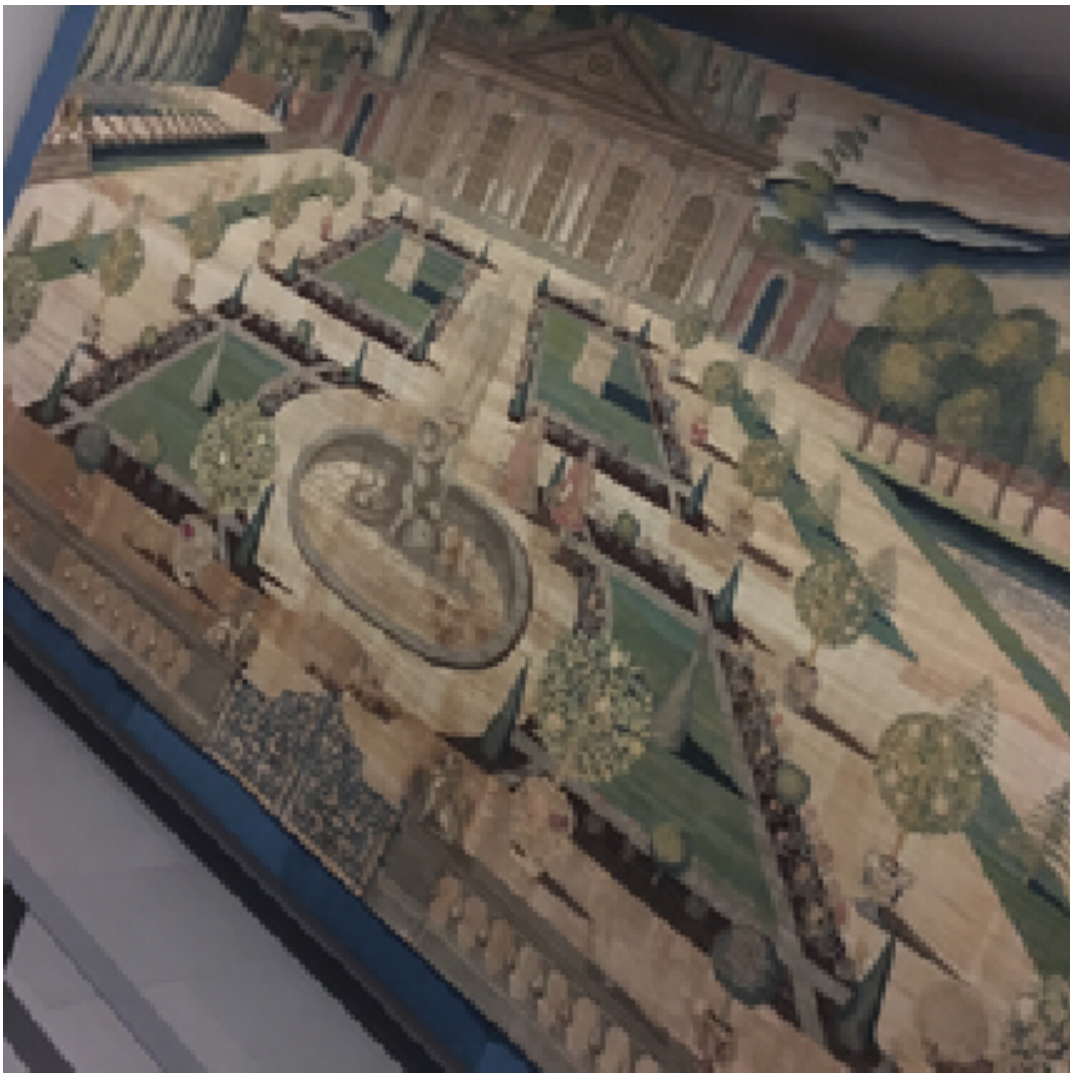}
    \includegraphics[width=0.135\textwidth,height=0.12\textwidth]{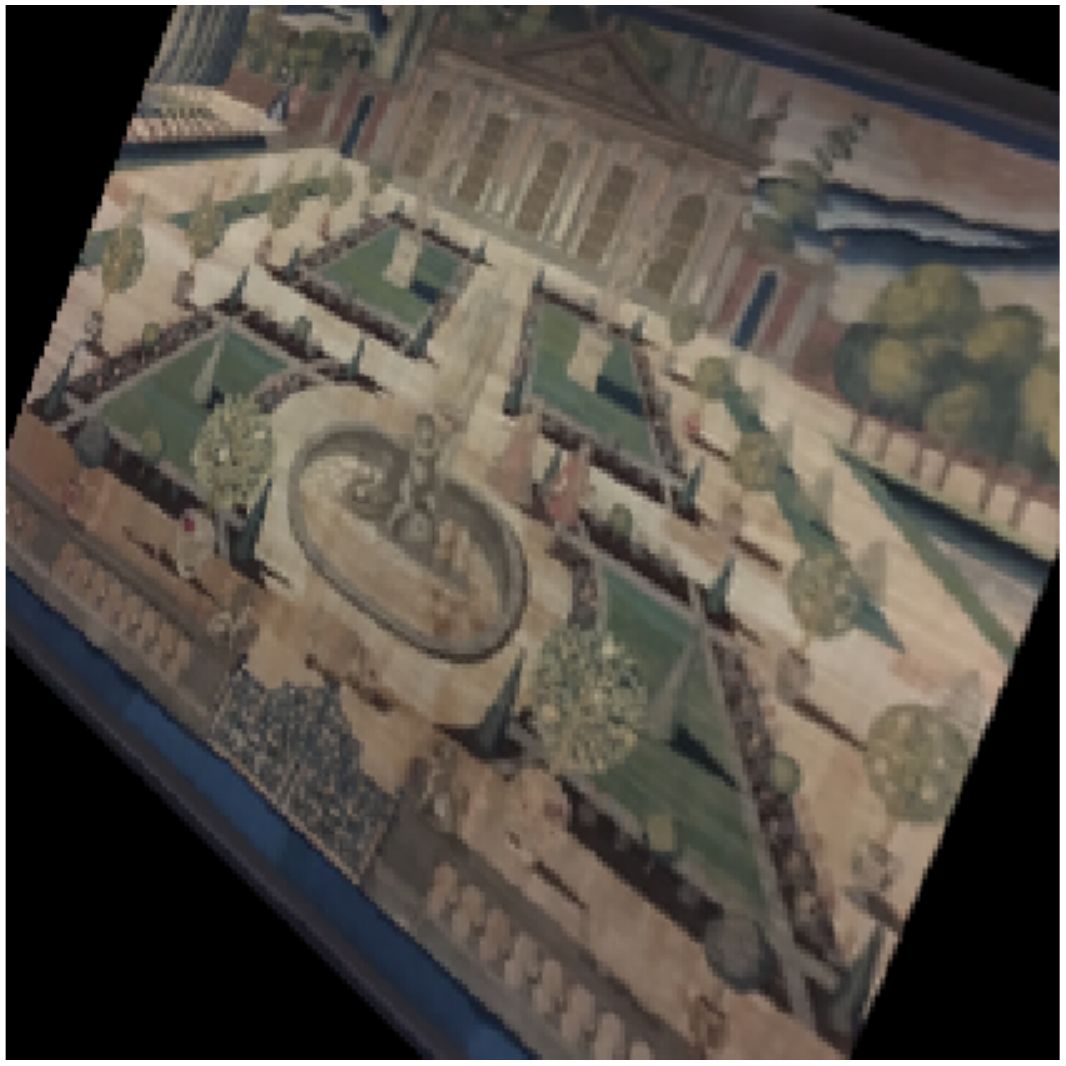}
    \includegraphics[width=0.135\textwidth,height=0.12\textwidth]{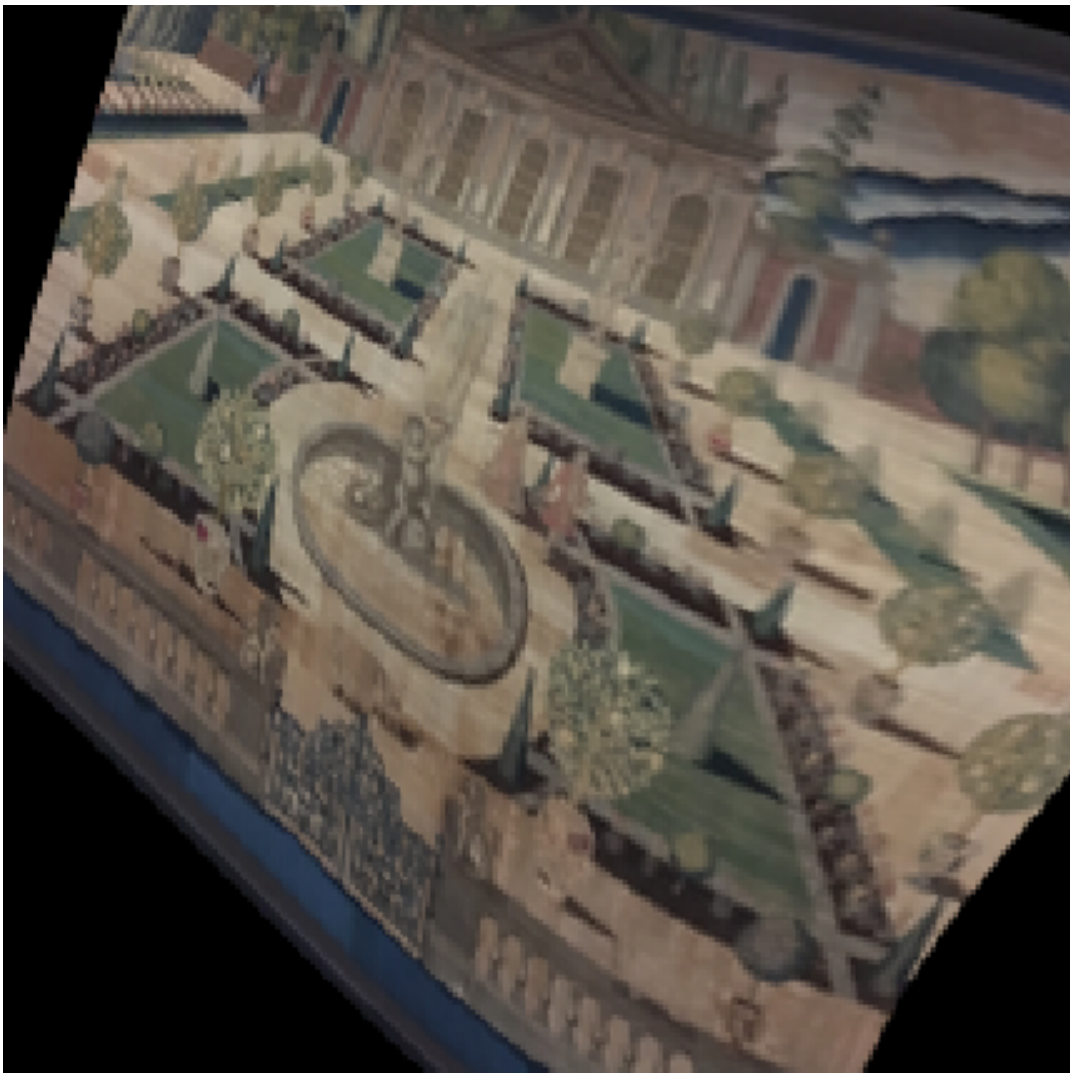}
    \includegraphics[width=0.135\textwidth,height=0.12\textwidth]{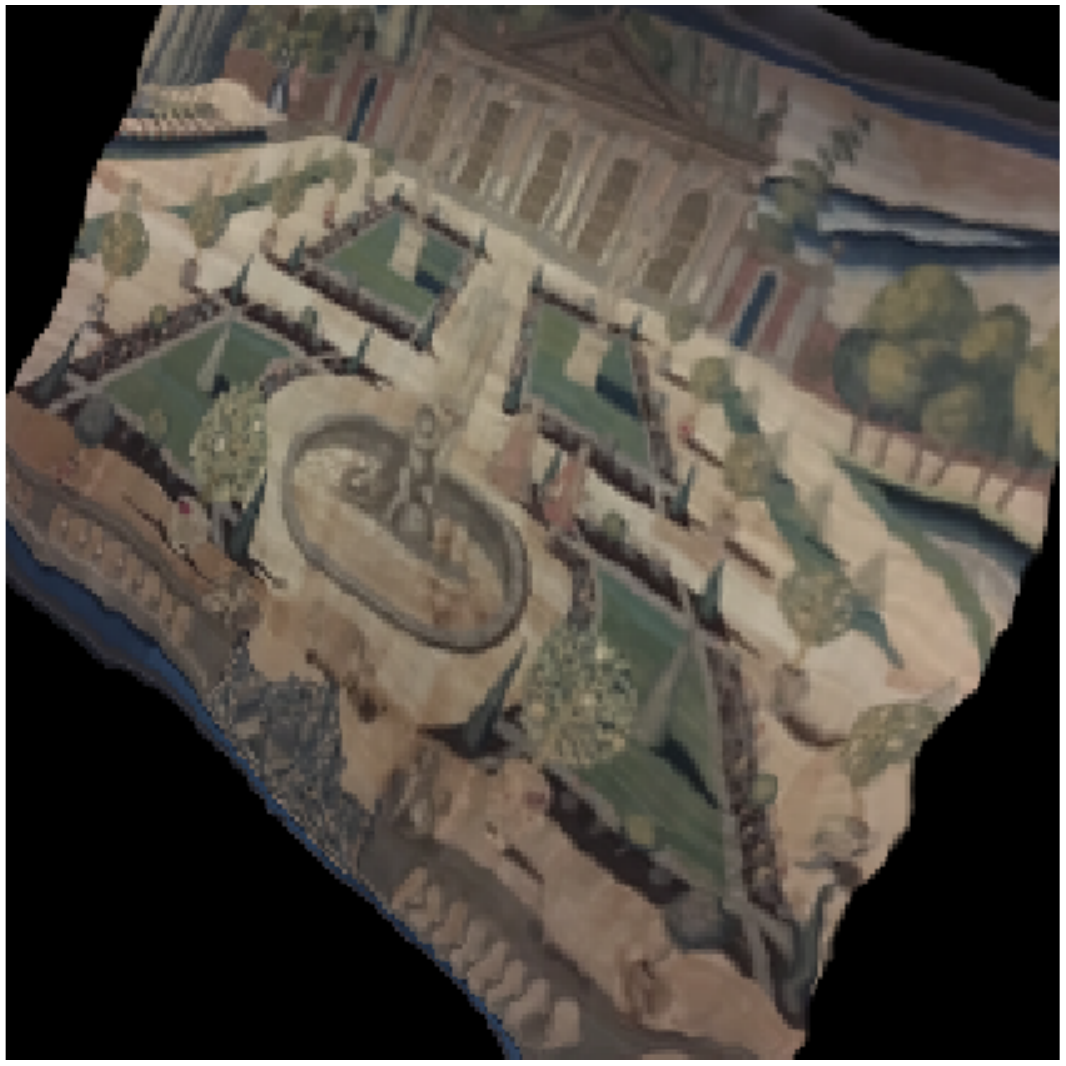}
    \includegraphics[width=0.135\textwidth,height=0.12\textwidth]{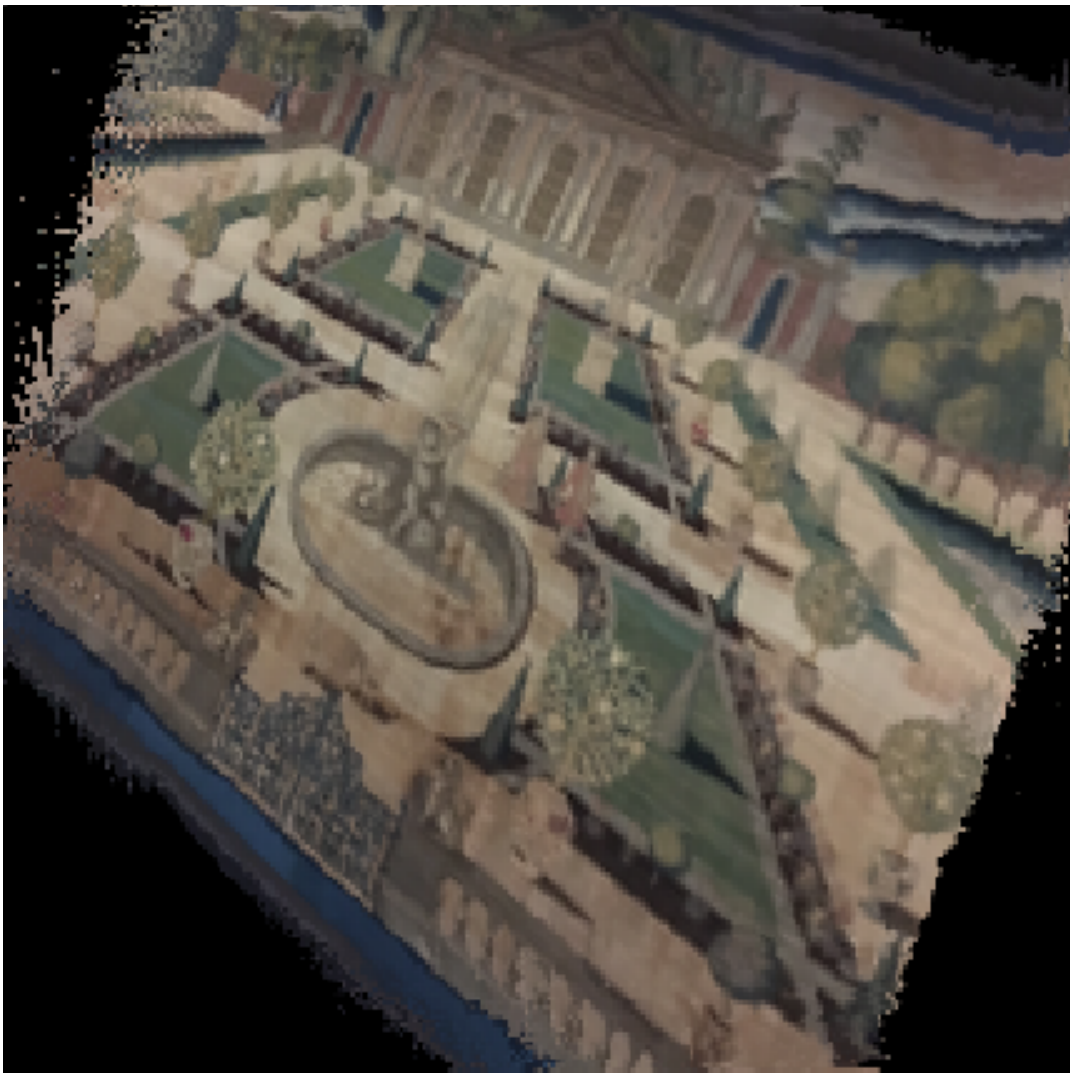}
    \includegraphics[width=0.135\textwidth,height=0.12\textwidth]{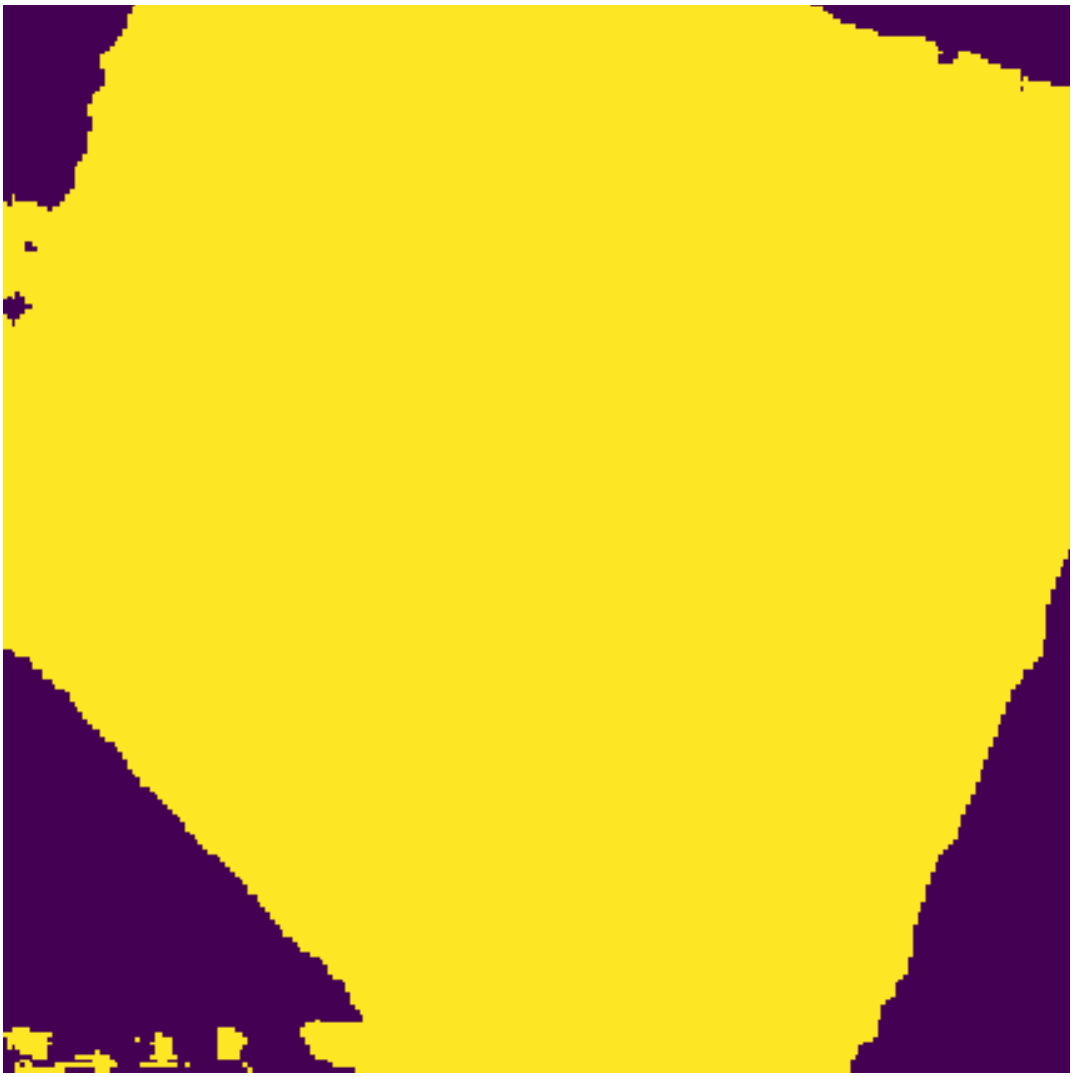}
    \vspace{0.5mm}
    \includegraphics[width=0.135\textwidth,height=0.12\textwidth]{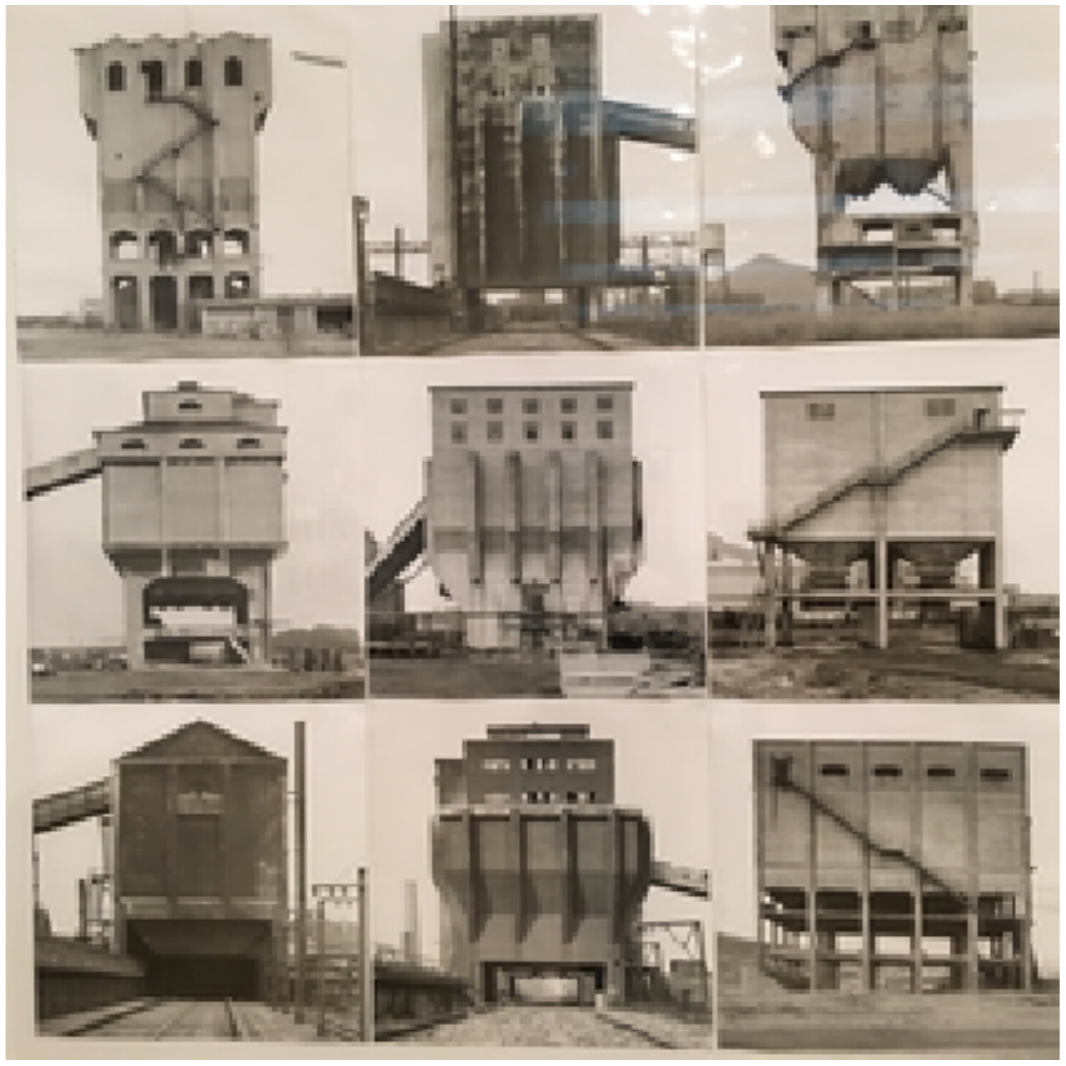}
    \includegraphics[width=0.135\textwidth,height=0.12\textwidth]{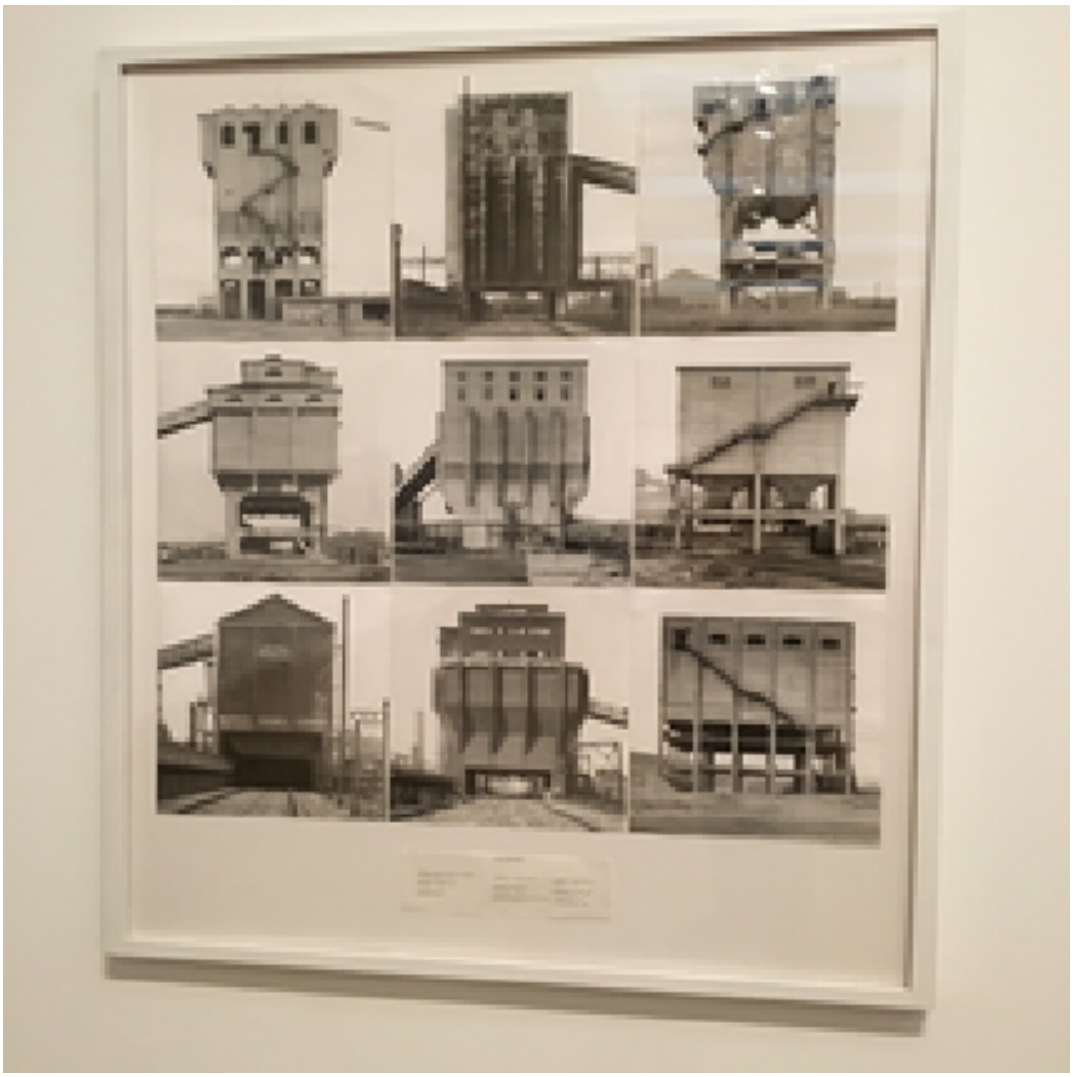}
    \includegraphics[width=0.135\textwidth,height=0.12\textwidth]{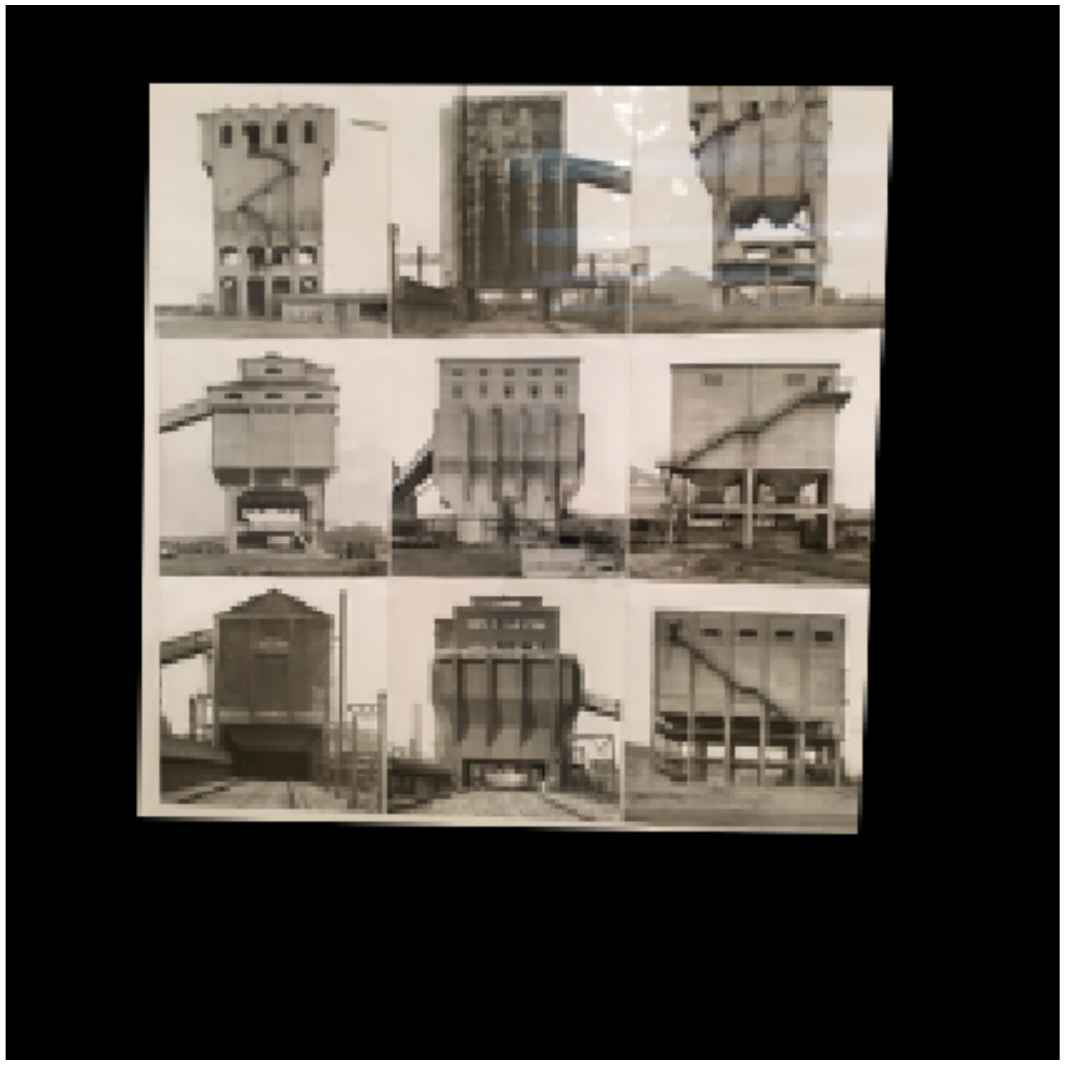}
    \includegraphics[width=0.135\textwidth,height=0.12\textwidth]{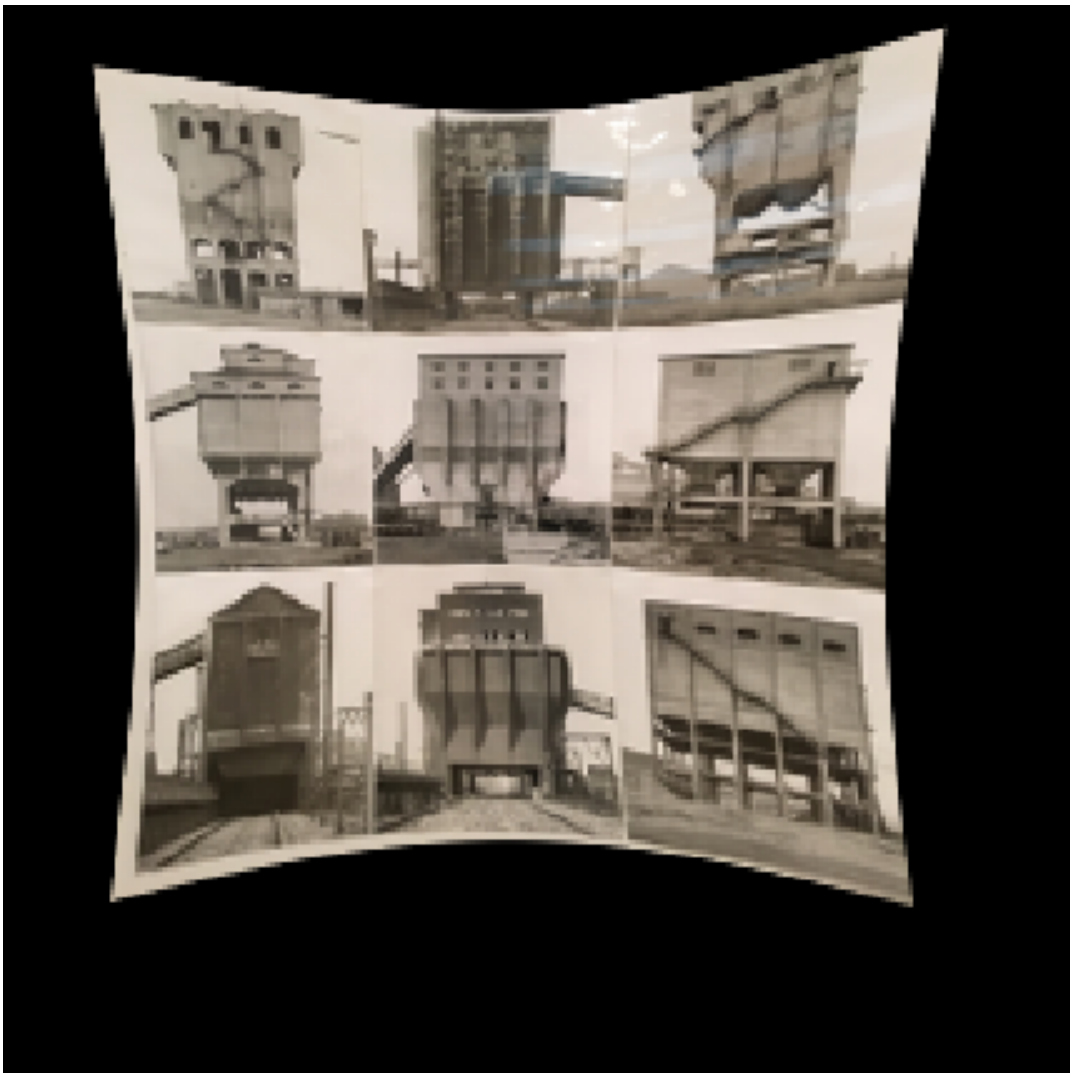}
    \includegraphics[width=0.135\textwidth,height=0.12\textwidth]{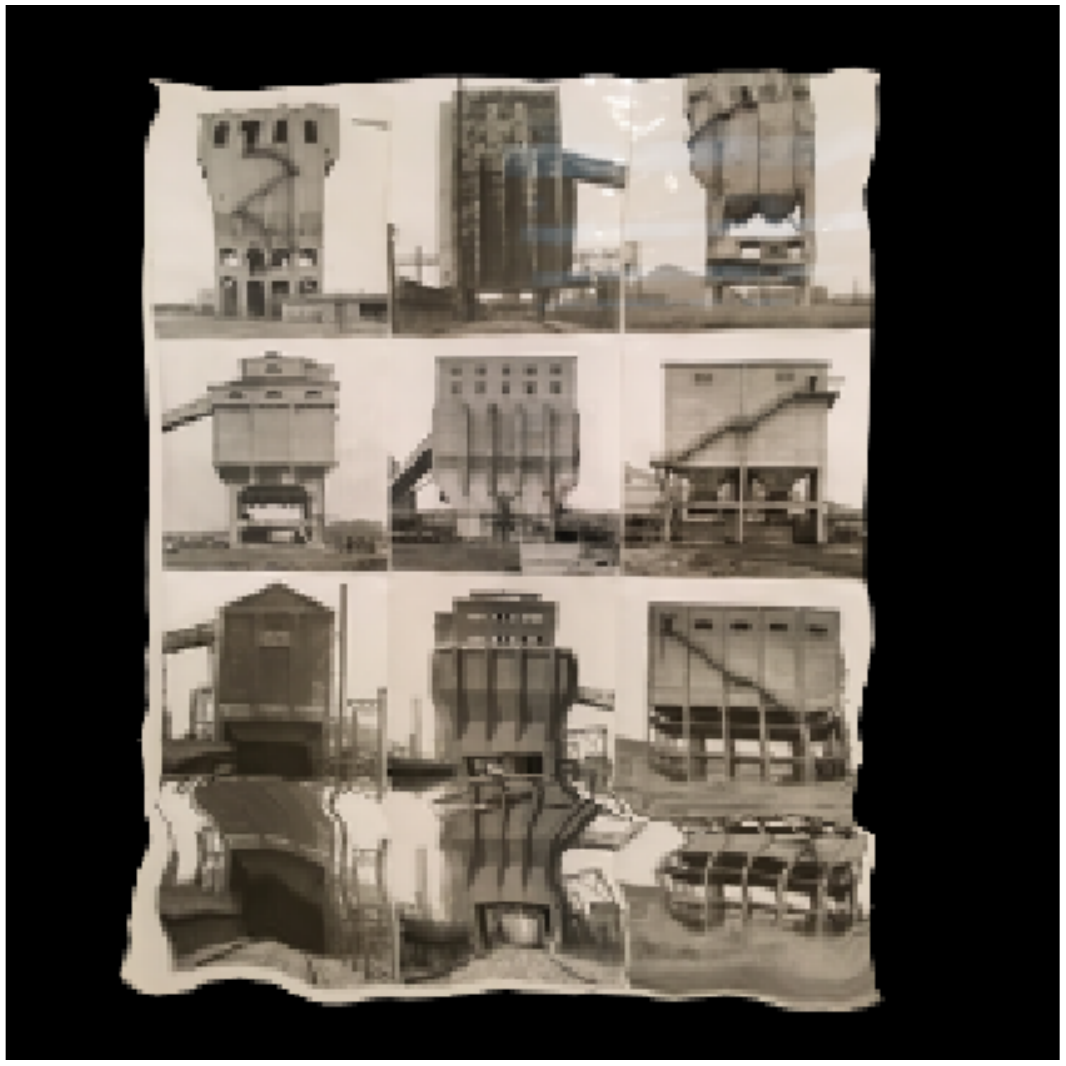}
    \includegraphics[width=0.135\textwidth,height=0.12\textwidth]{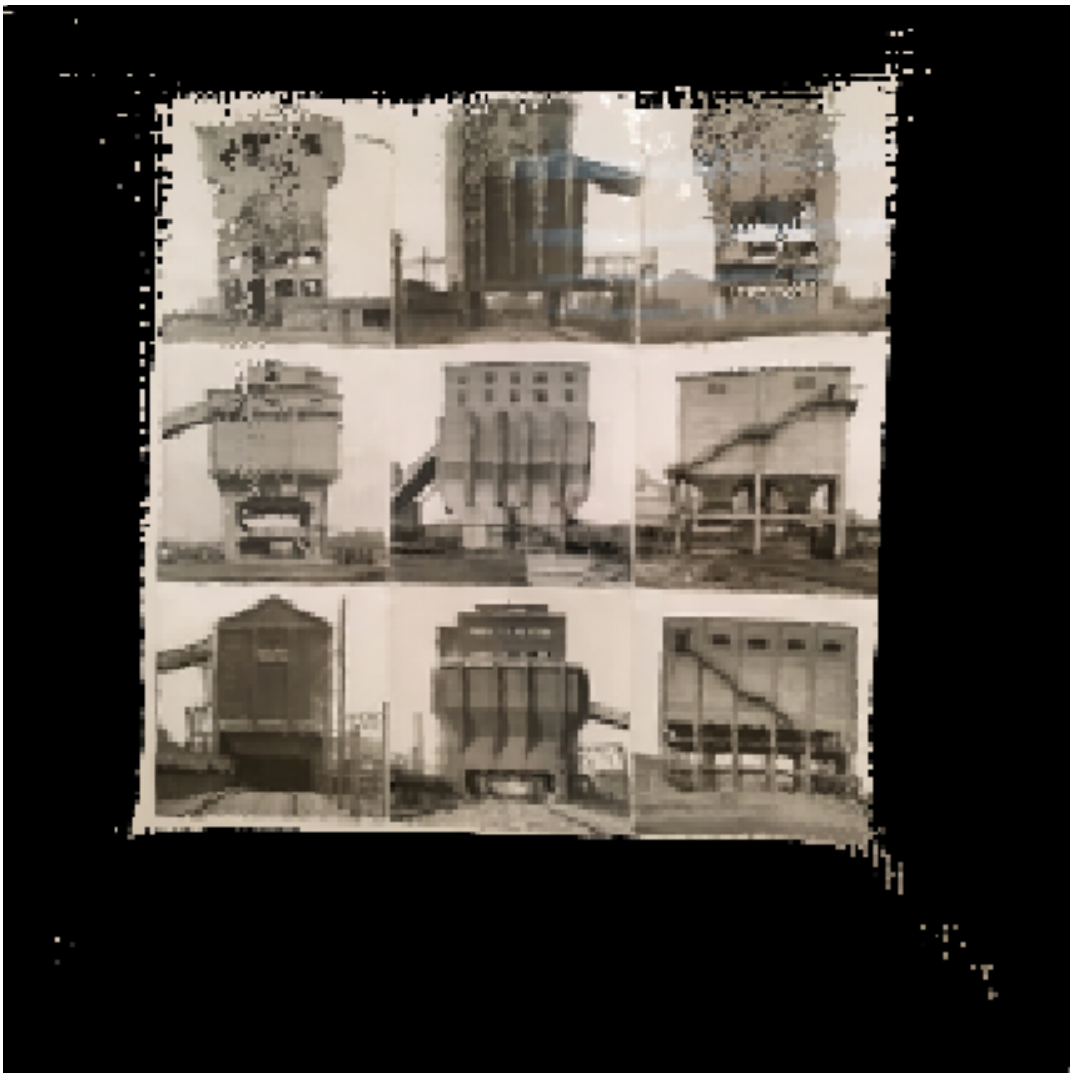}
    \includegraphics[width=0.135\textwidth,height=0.12\textwidth]{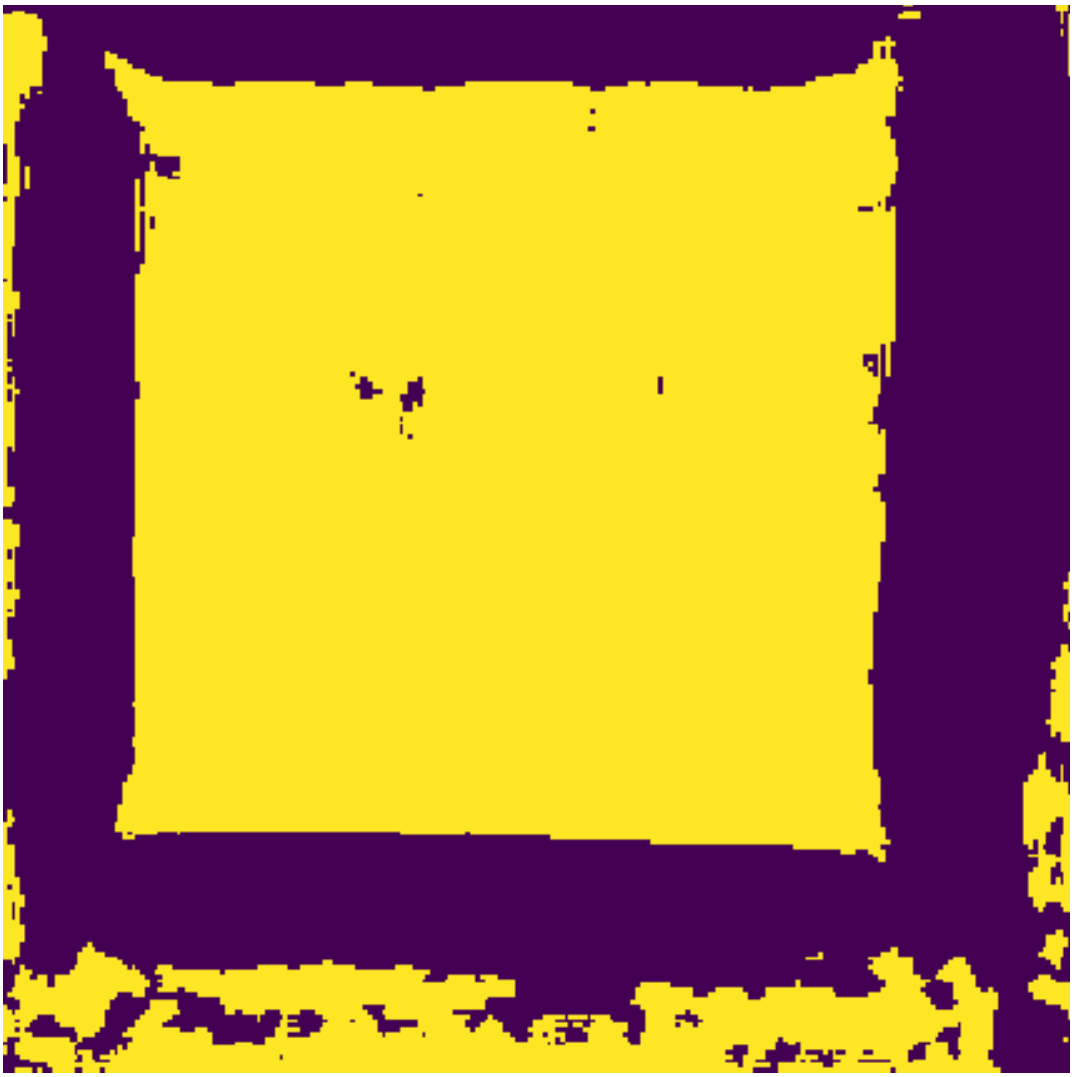}
    \vspace{0.5mm}
    \includegraphics[width=0.135\textwidth,height=0.12\textwidth]{}
    \includegraphics[width=0.135\textwidth,height=0.12\textwidth]{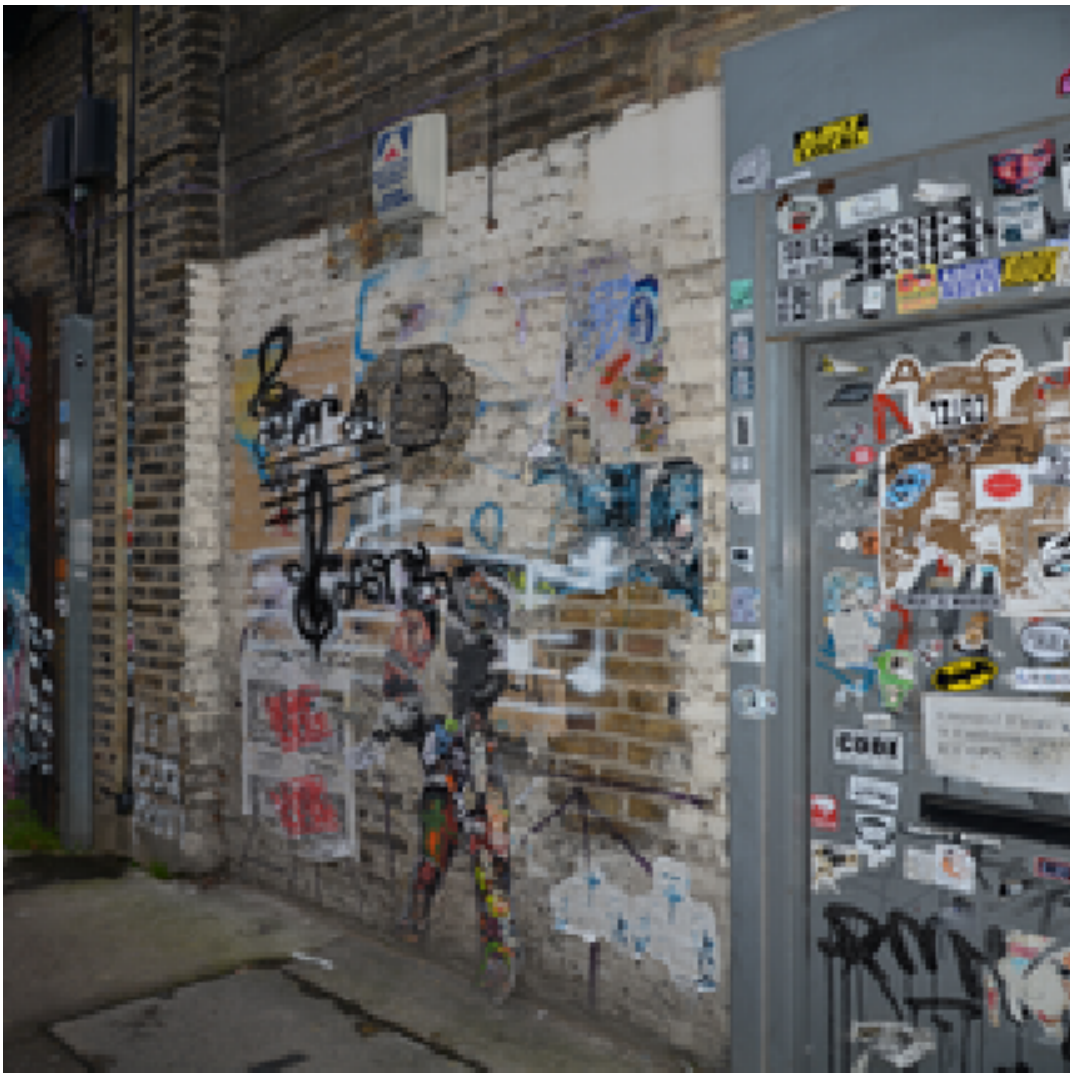}
    \includegraphics[width=0.135\textwidth,height=0.12\textwidth]{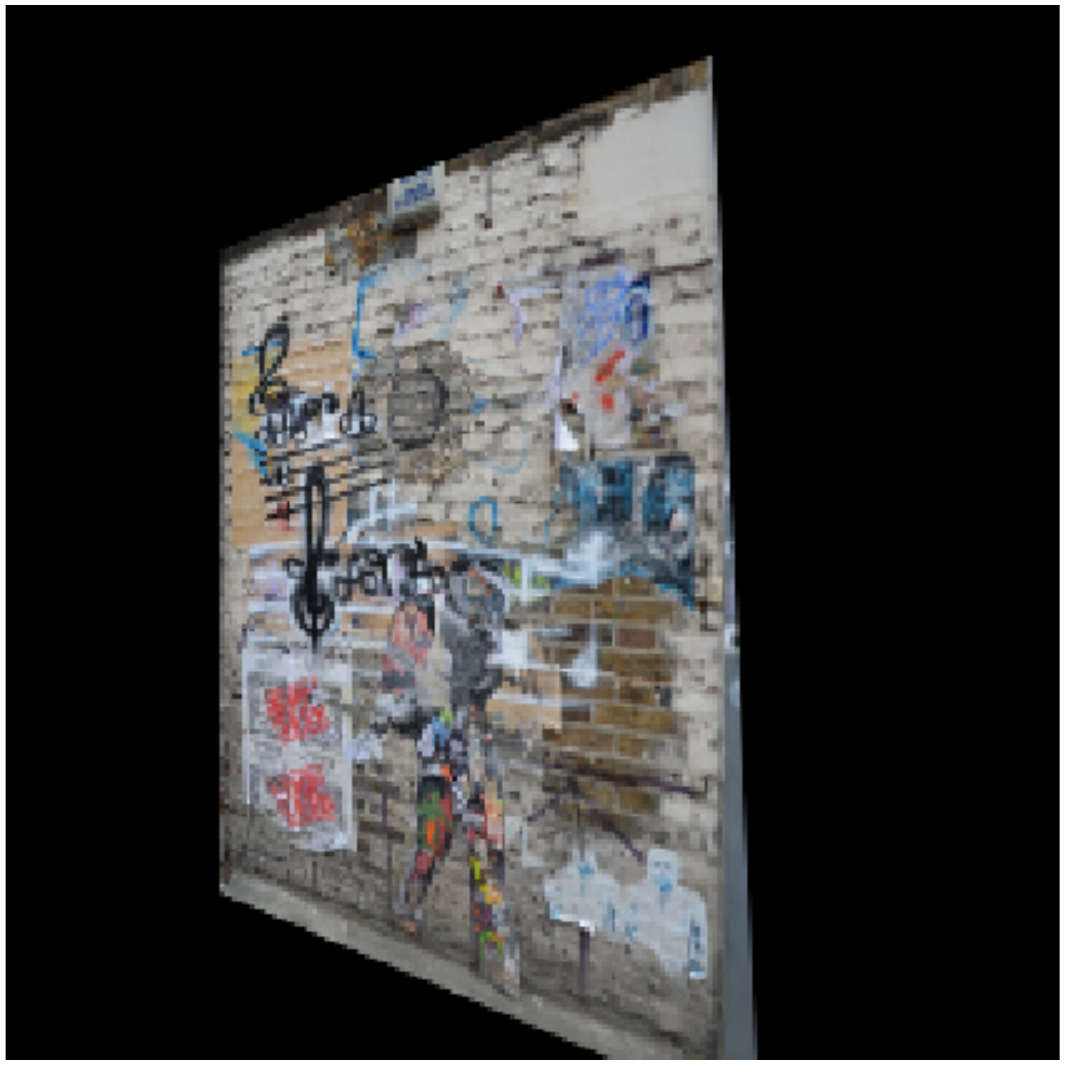}
    \includegraphics[width=0.135\textwidth,height=0.12\textwidth]{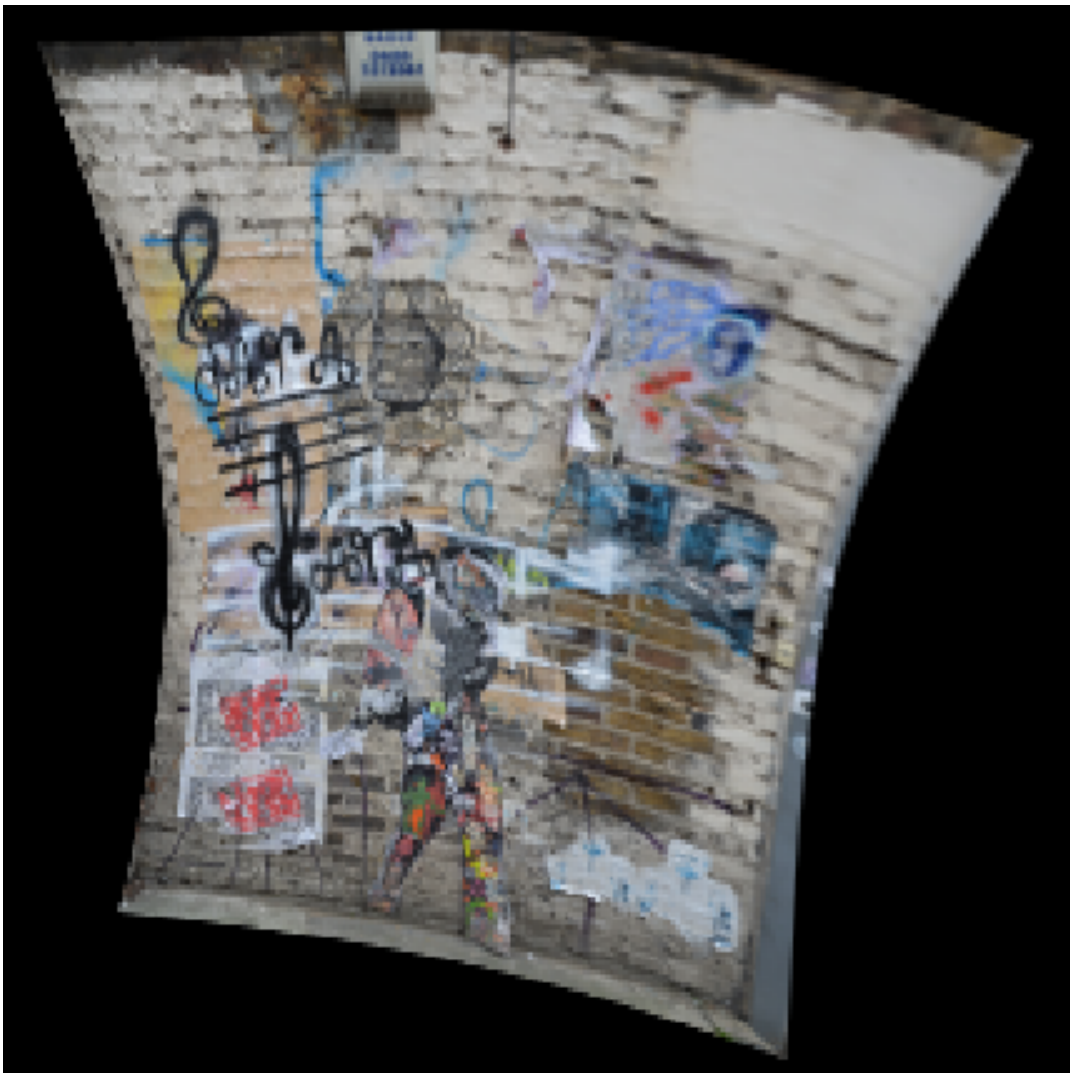}
    \includegraphics[width=0.135\textwidth,height=0.12\textwidth]{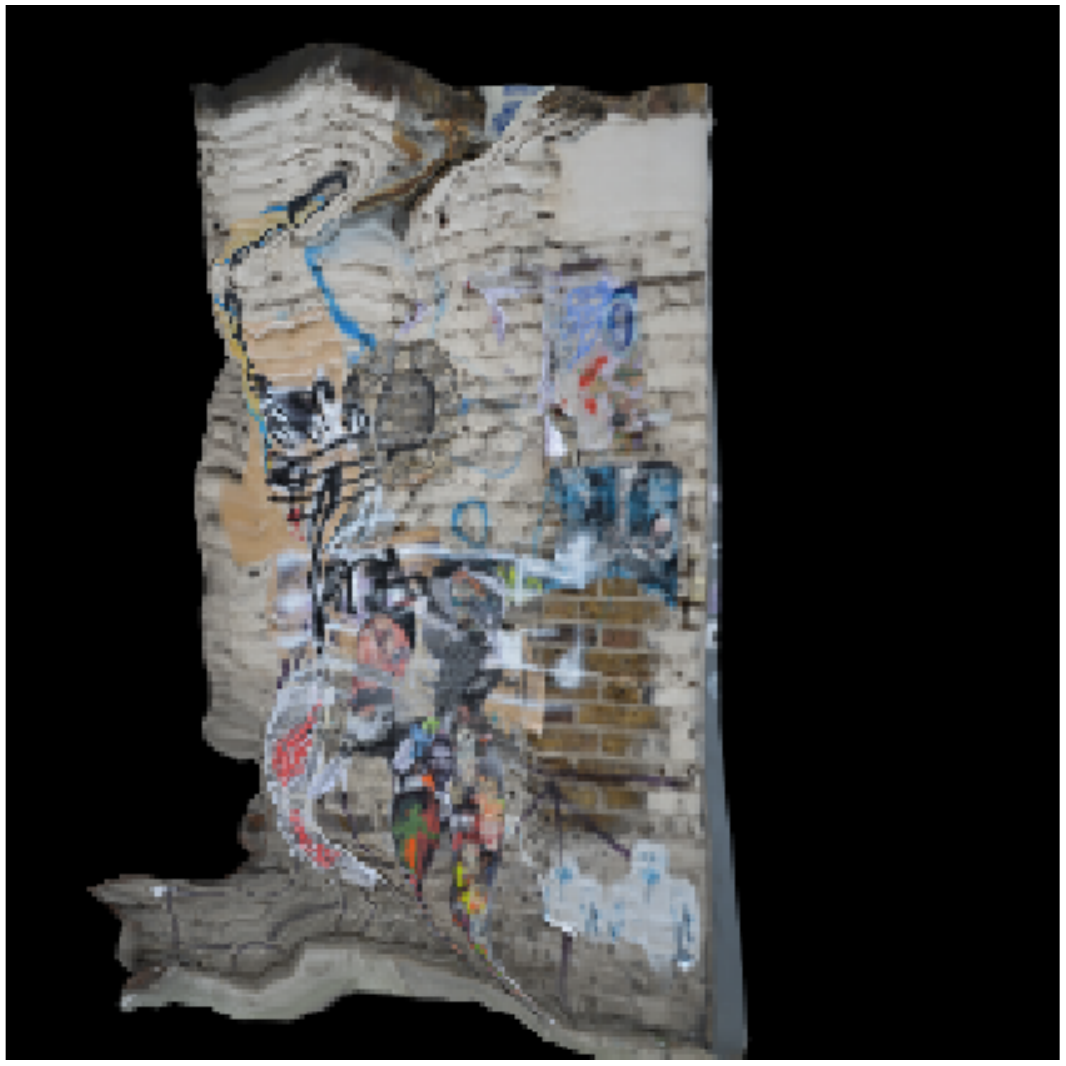}
    \includegraphics[width=0.135\textwidth,height=0.12\textwidth]{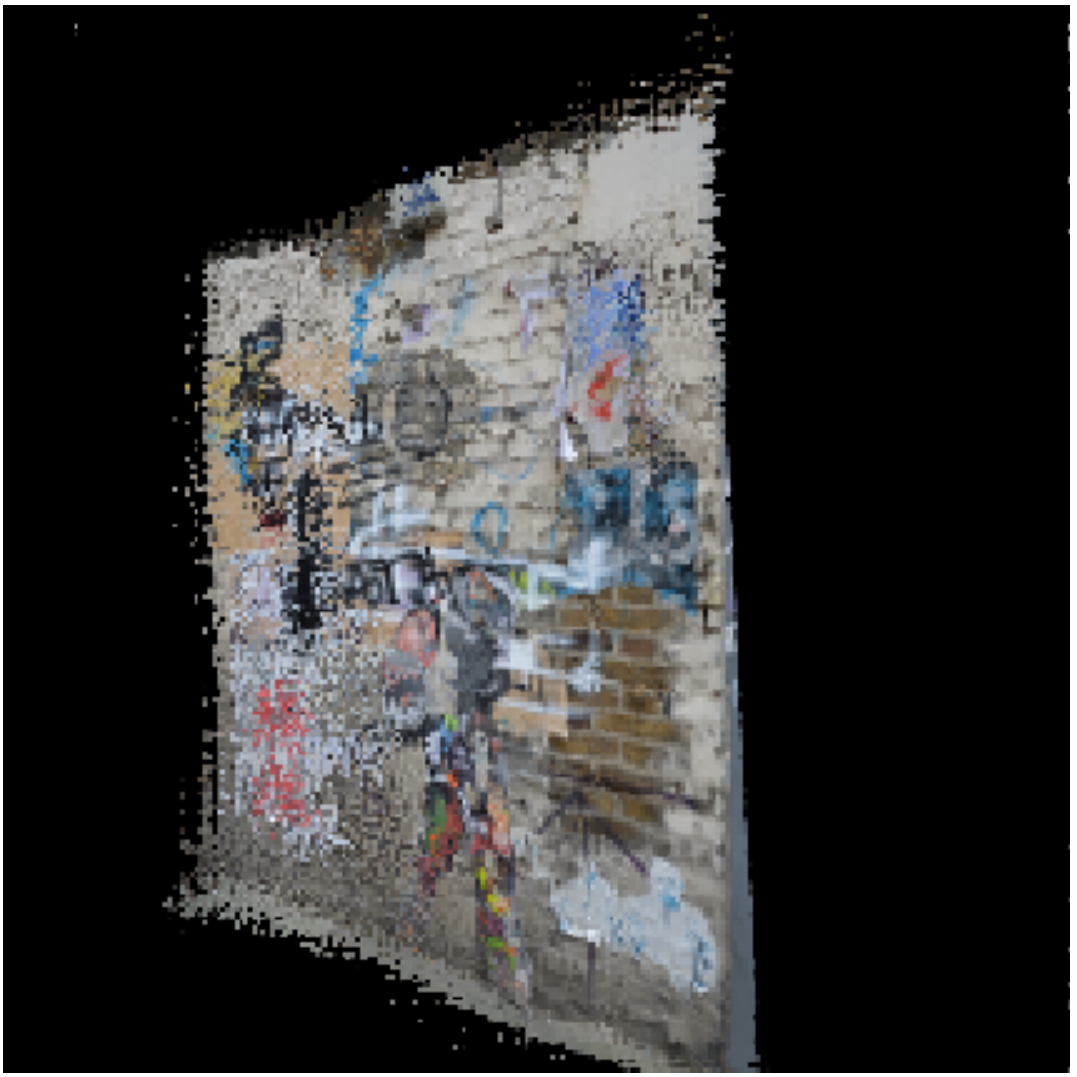}
    \includegraphics[width=0.135\textwidth,height=0.12\textwidth]{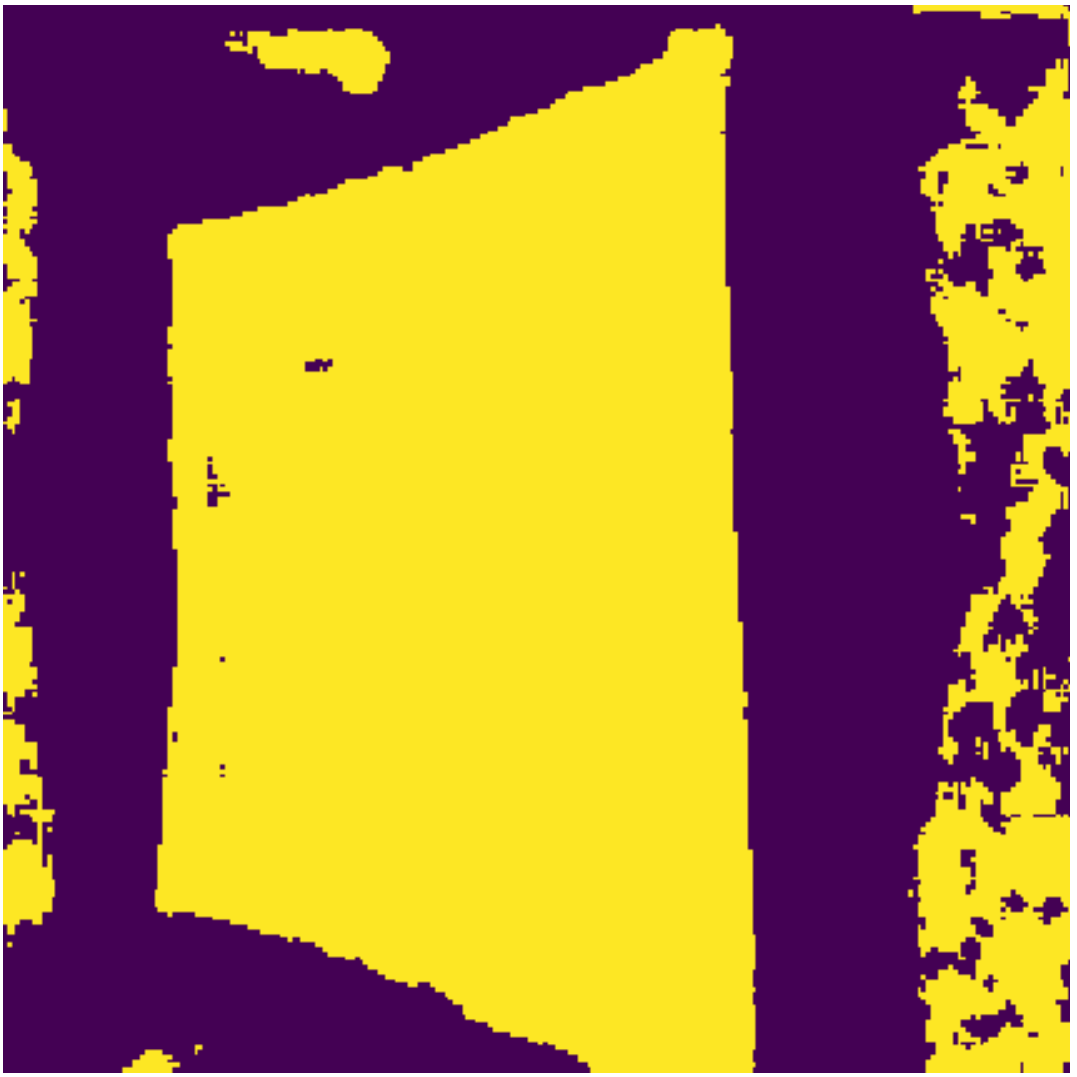}
    \vspace{0.5mm}
\small{
\hfill
\mpage{0.100}{Reference} \hfill
\mpage{0.100}{Target} \hfill
\mpage{0.100}{Ground Truth} \hfill
\mpage{0.100}{Rocco~\etal~\cite{Rocco17}}\hfill
\mpage{0.100}{PWC-Net~\cite{PWC-Net}} \hfill
\mpage{0.100}{DGC+M-Net} \hfill
\mpage{0.100}{Predicted mask} \hfill
}
\vspace{0pt}
\caption{Qualitative comparisons between different algorithms on the HPatches dataset. Our model produces more accurate correspondence map leading to better image alignment.}
	\label{fig:comparison_hpatches_qual}
\end{figure*}

\begin{figure*}[t!]
	\centering
    \includegraphics[width=0.155\textwidth,height=0.12\textwidth]{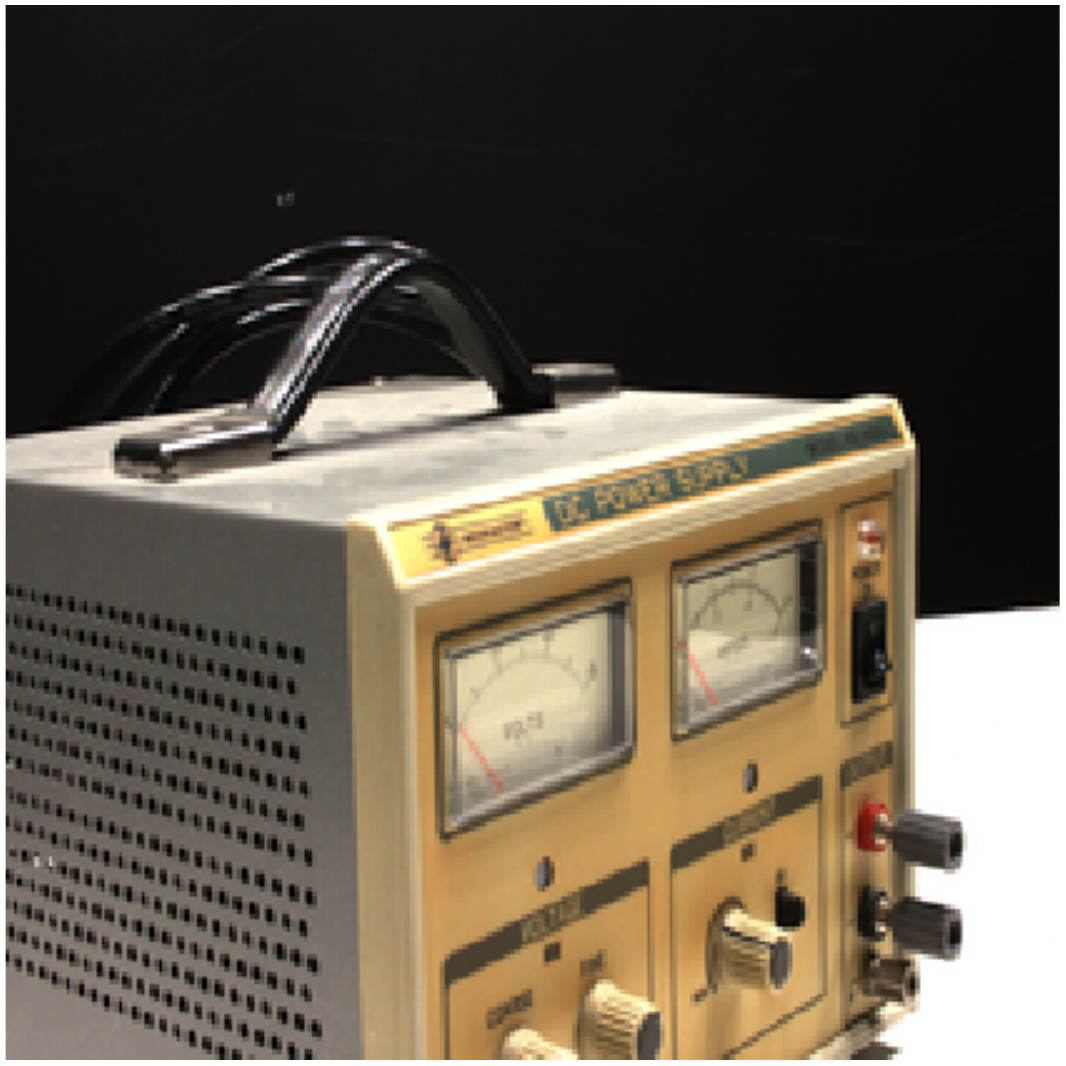}
    \includegraphics[width=0.155\textwidth,height=0.12\textwidth]{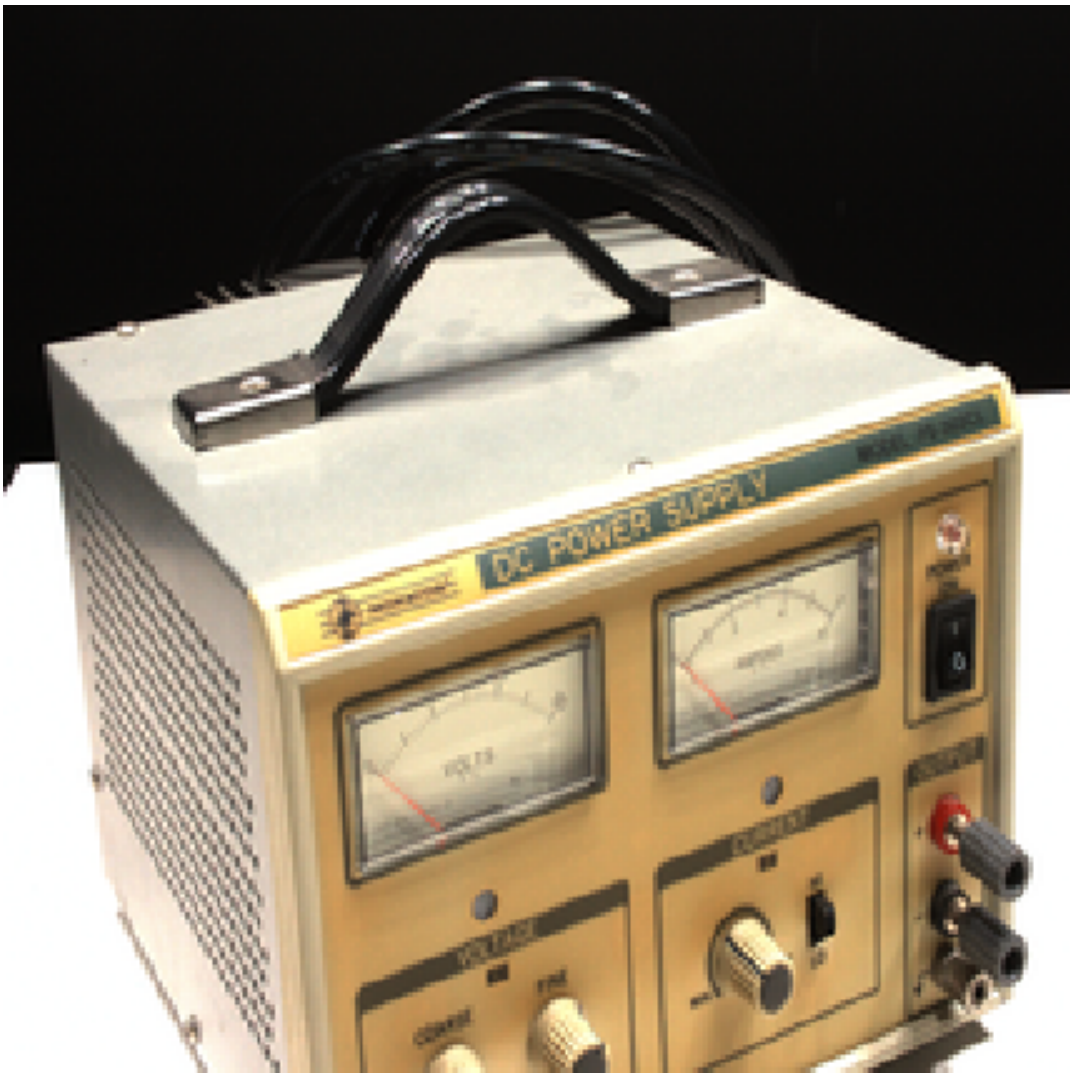}
    \includegraphics[width=0.155\textwidth,height=0.12\textwidth]{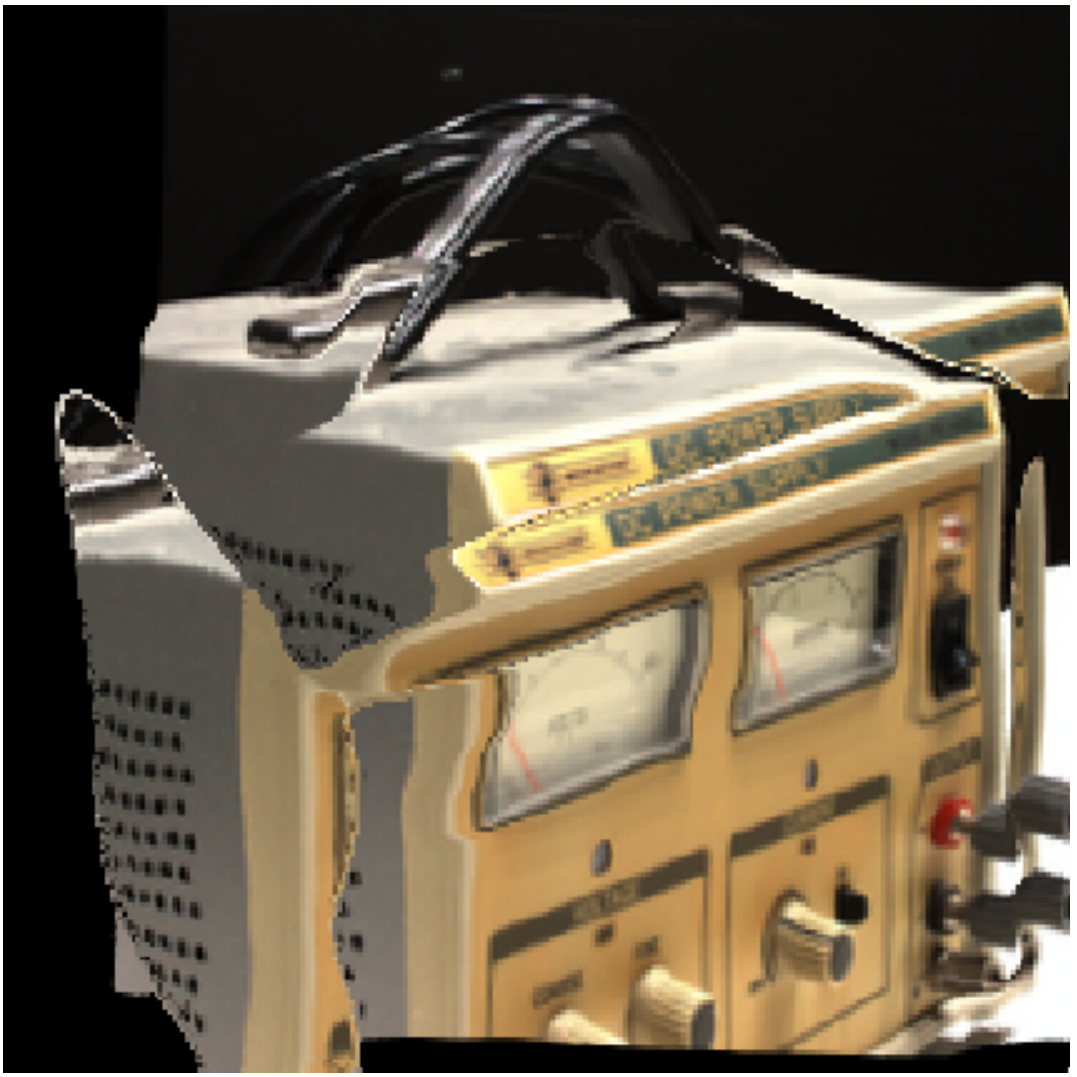}
    \includegraphics[width=0.155\textwidth,height=0.12\textwidth]{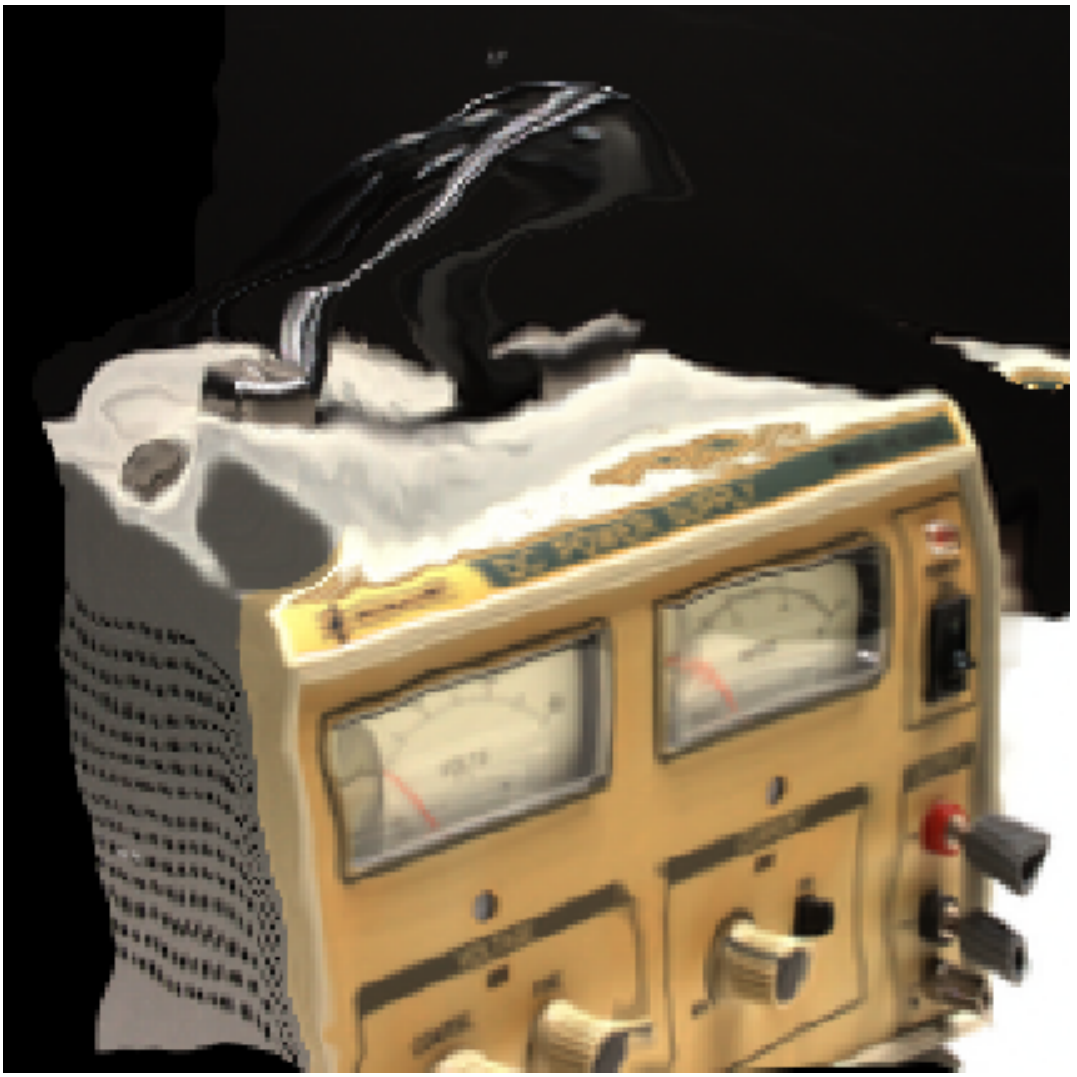}
    \includegraphics[width=0.155\textwidth,height=0.12\textwidth]{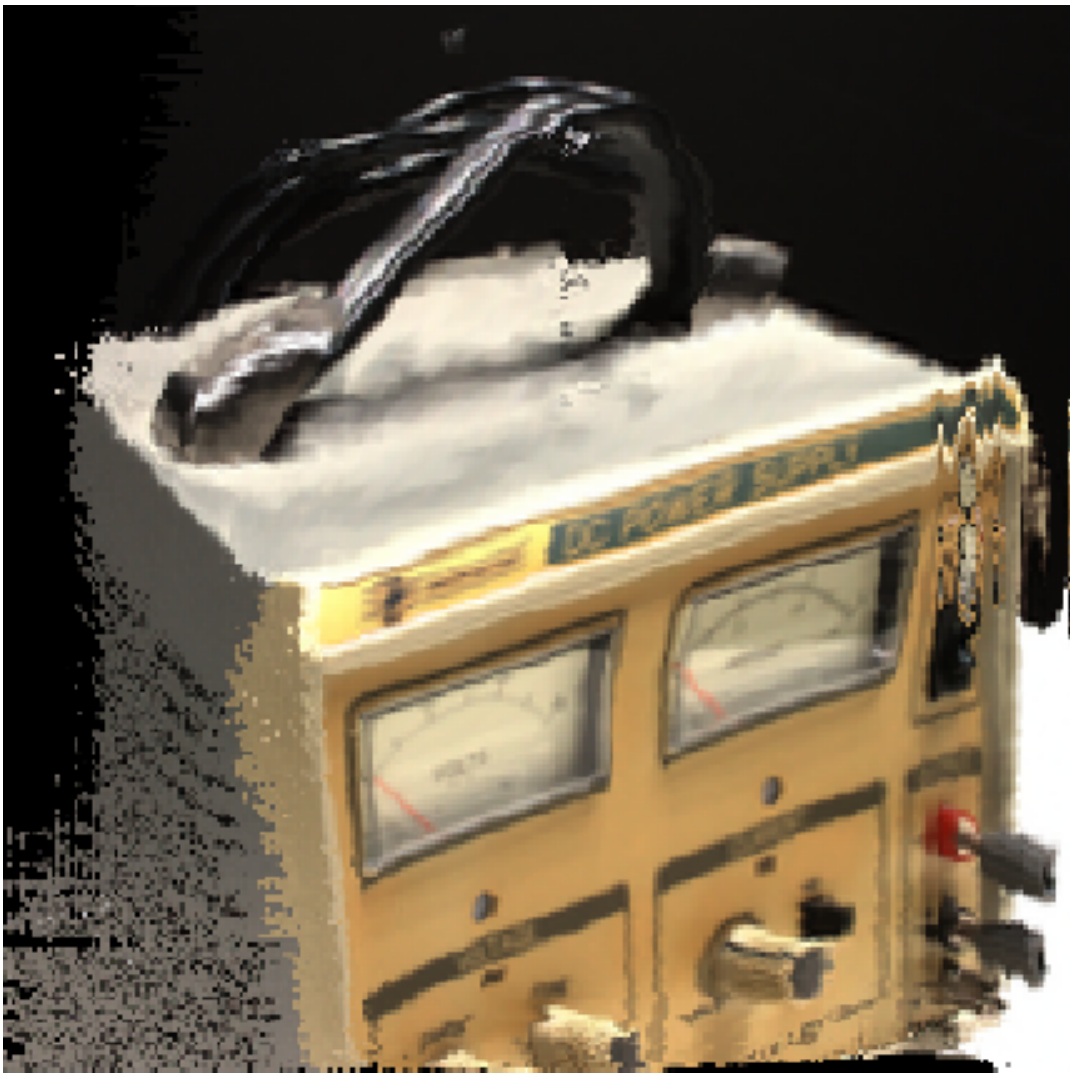}
    \includegraphics[width=0.155\textwidth,height=0.12\textwidth]{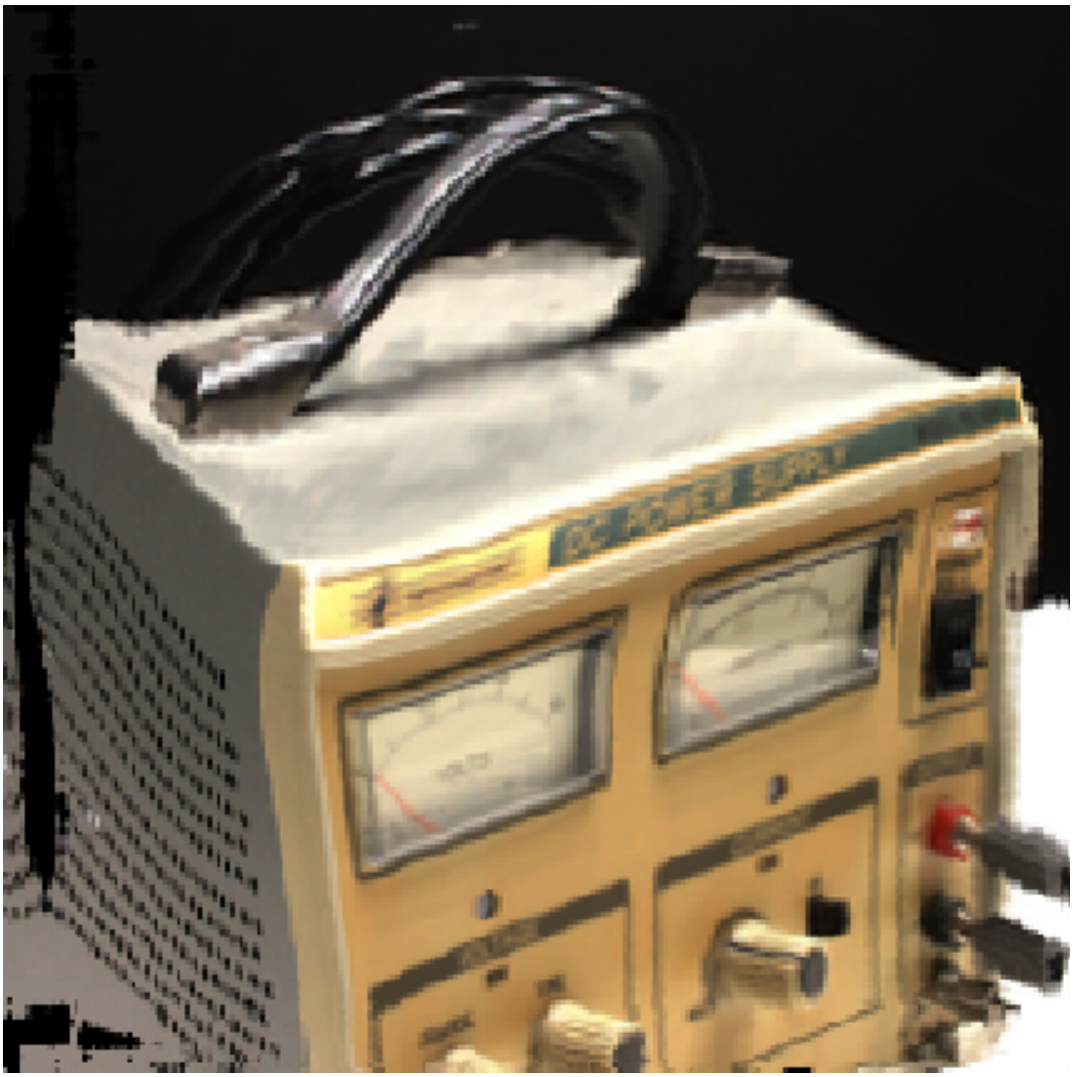}
    \vspace{0.3mm}
    \includegraphics[width=0.155\textwidth,height=0.12\textwidth]{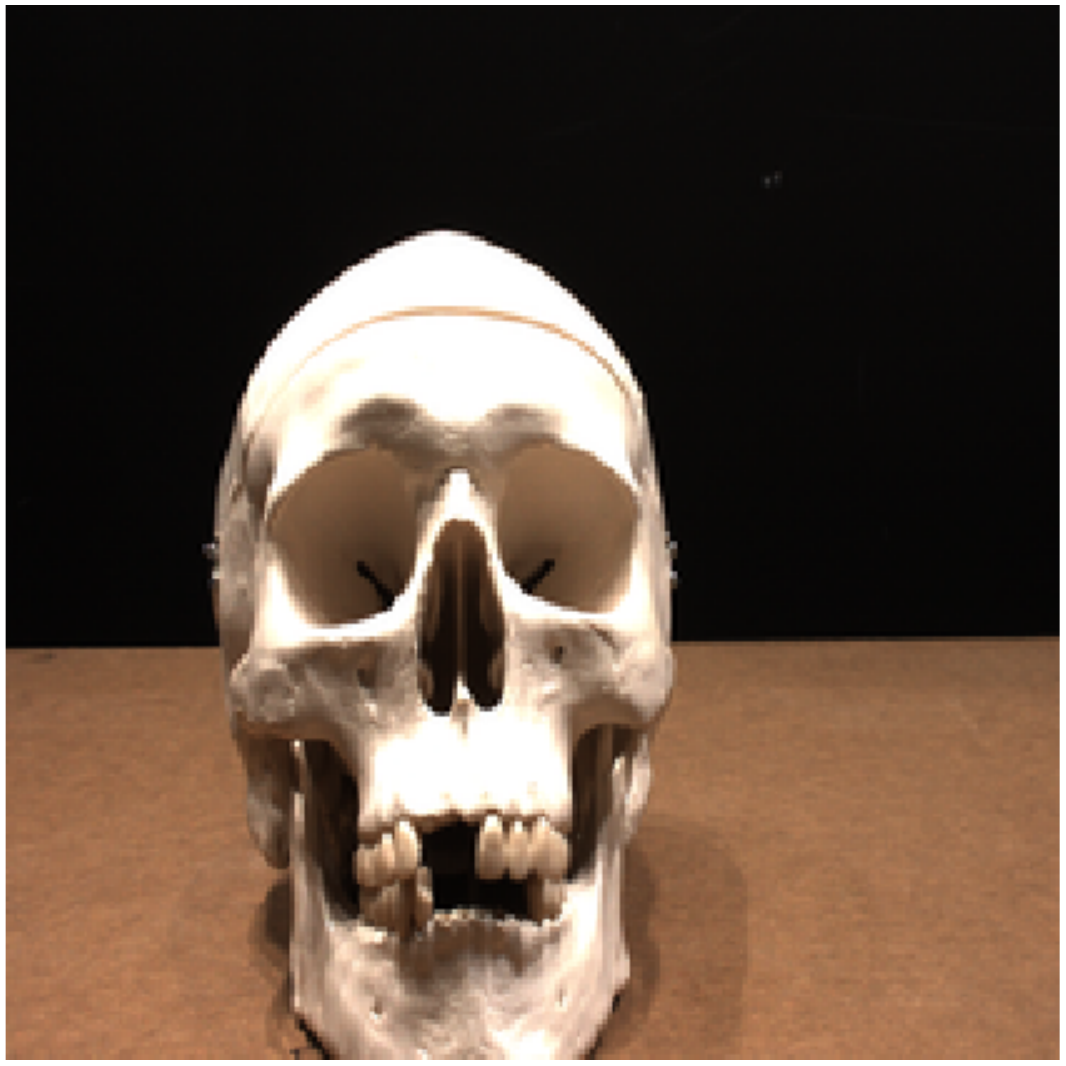}
    \includegraphics[width=0.155\textwidth,height=0.12\textwidth]{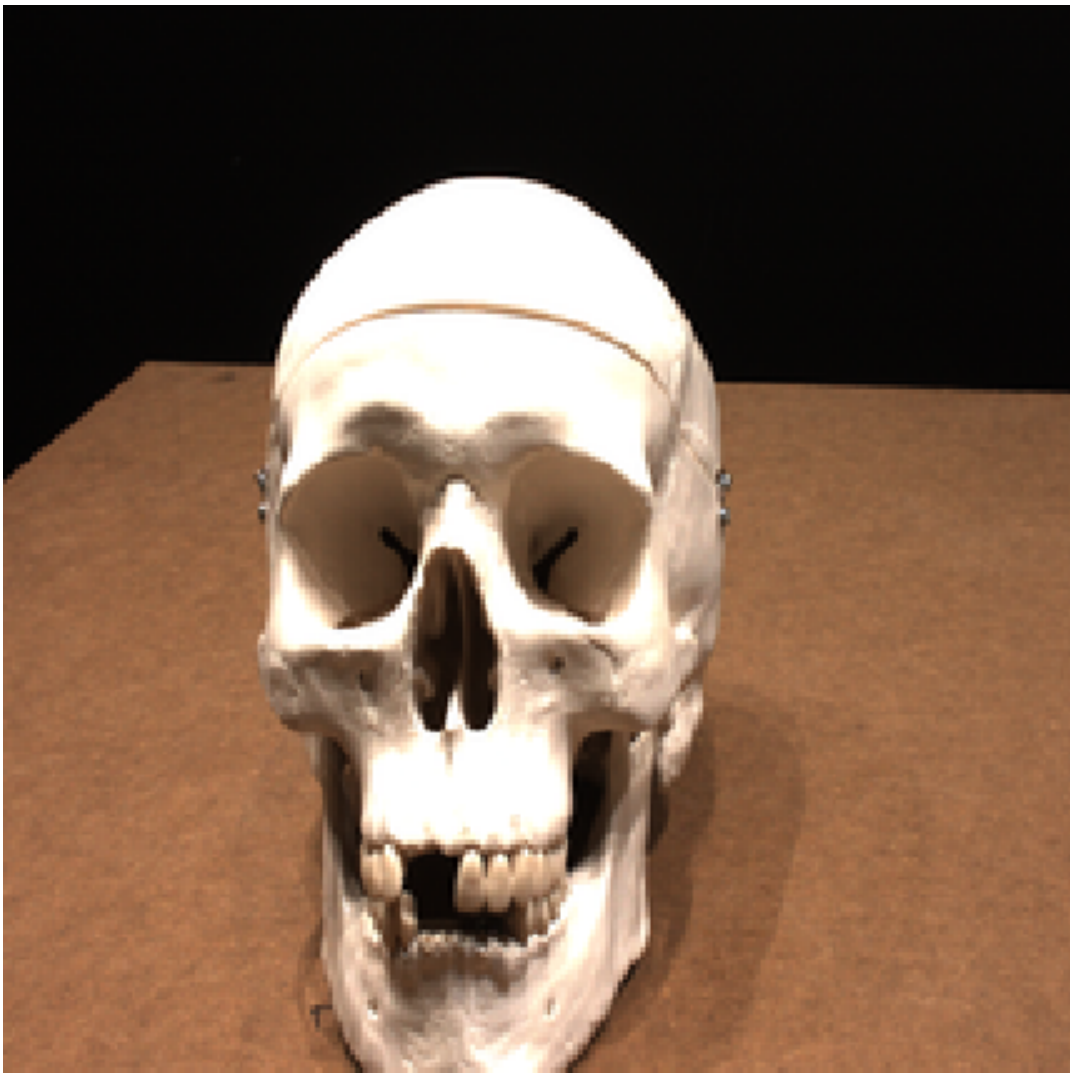}
    \includegraphics[width=0.155\textwidth,height=0.12\textwidth]{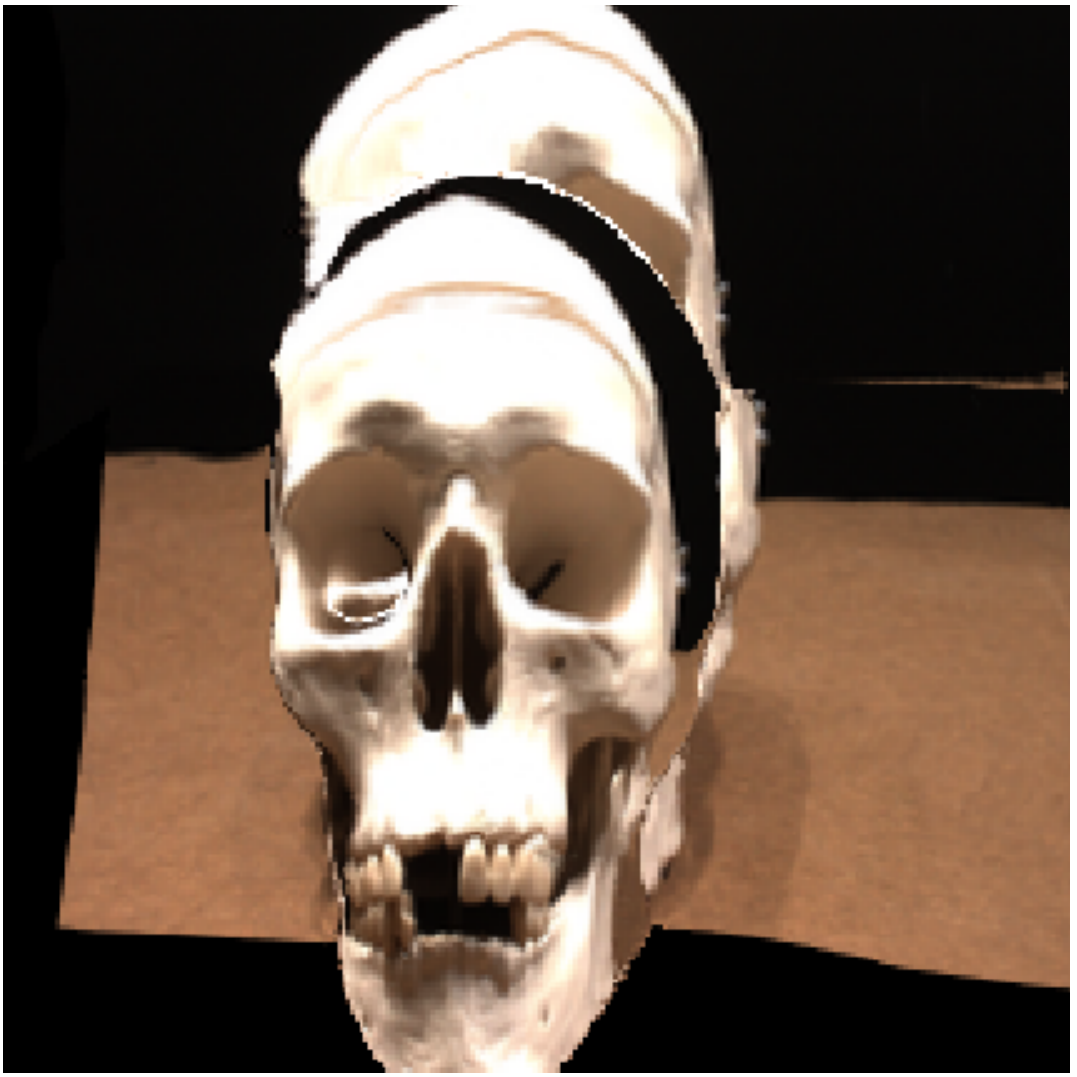}
    \includegraphics[width=0.155\textwidth,height=0.12\textwidth]{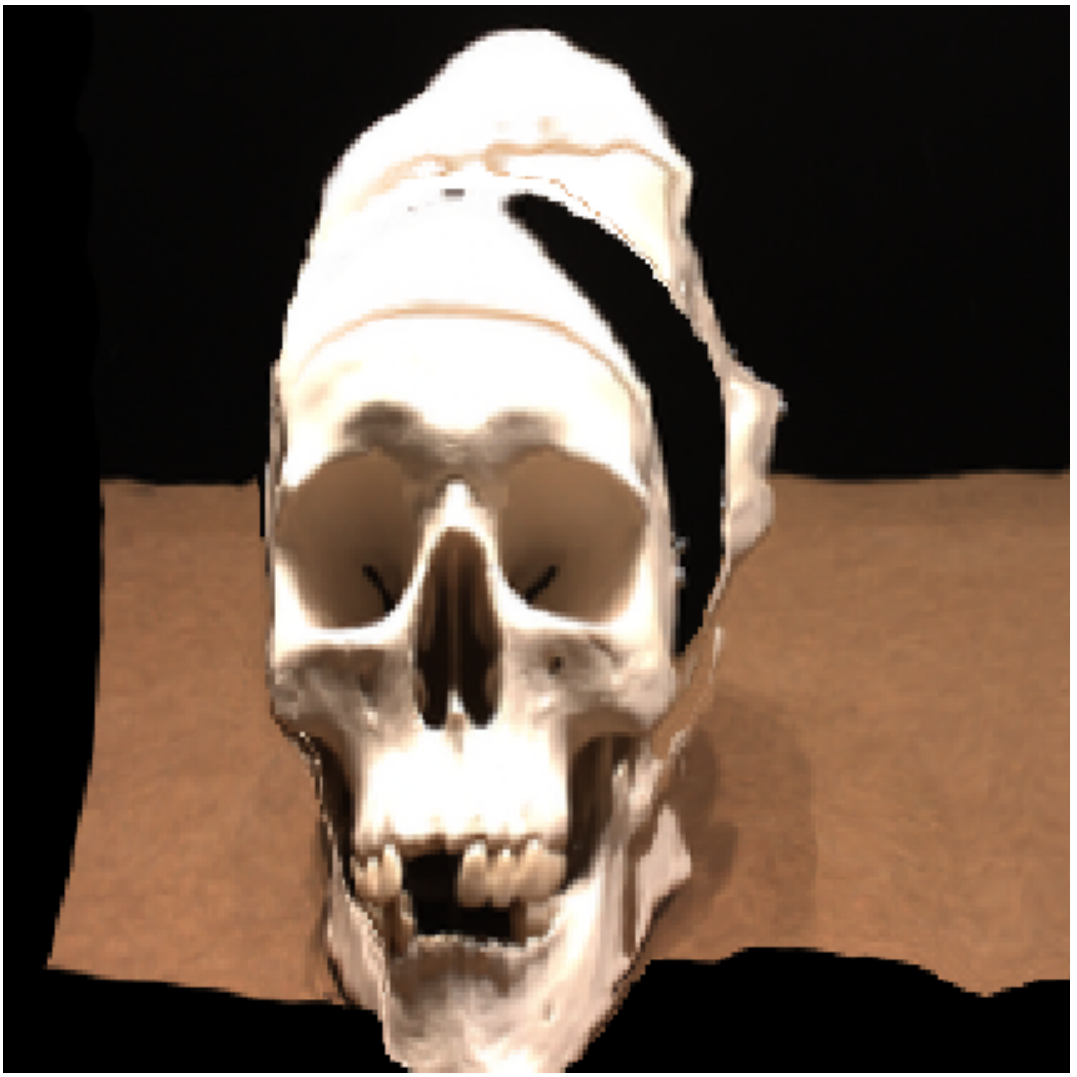}
    \includegraphics[width=0.155\textwidth,height=0.12\textwidth]{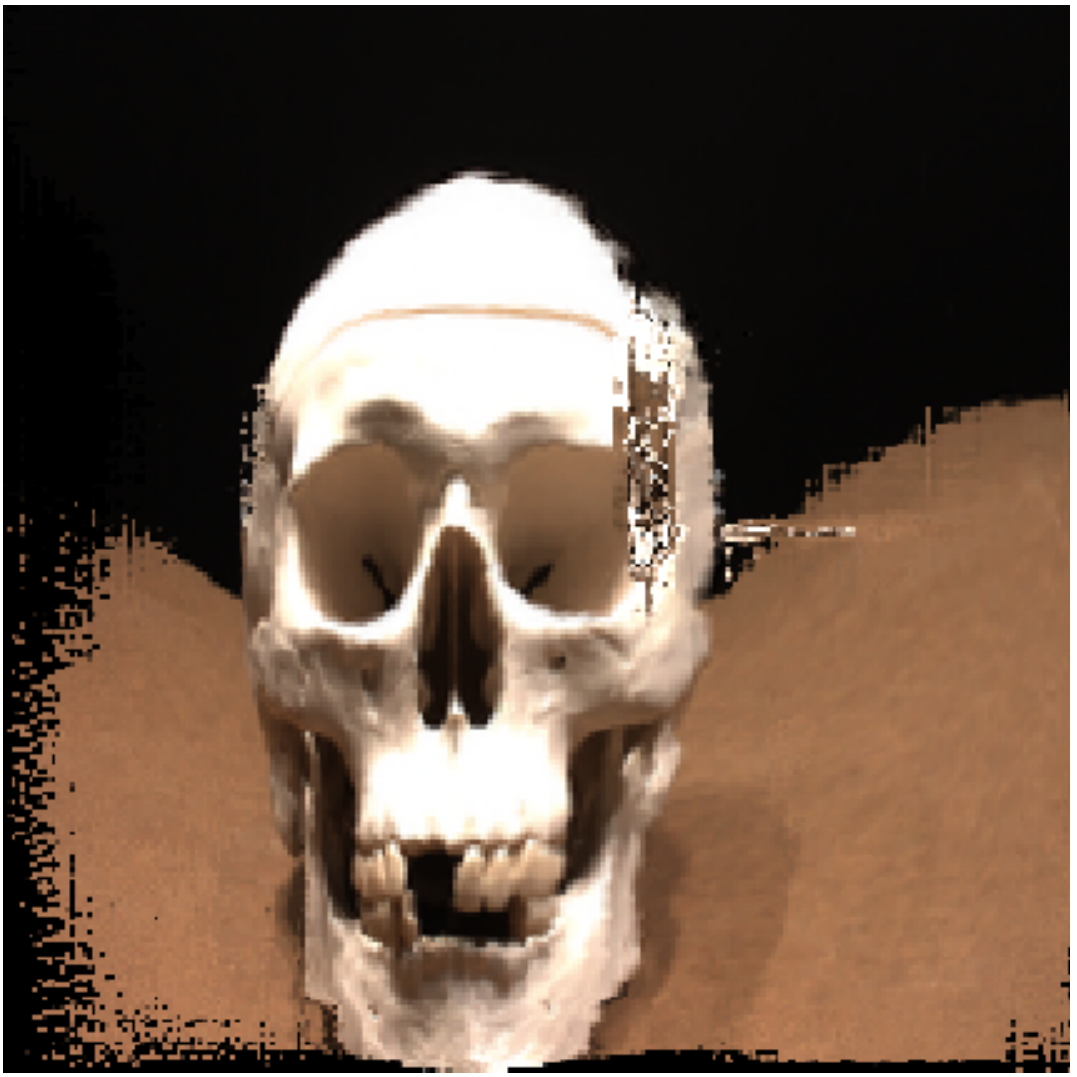}
    \includegraphics[width=0.155\textwidth,height=0.12\textwidth]{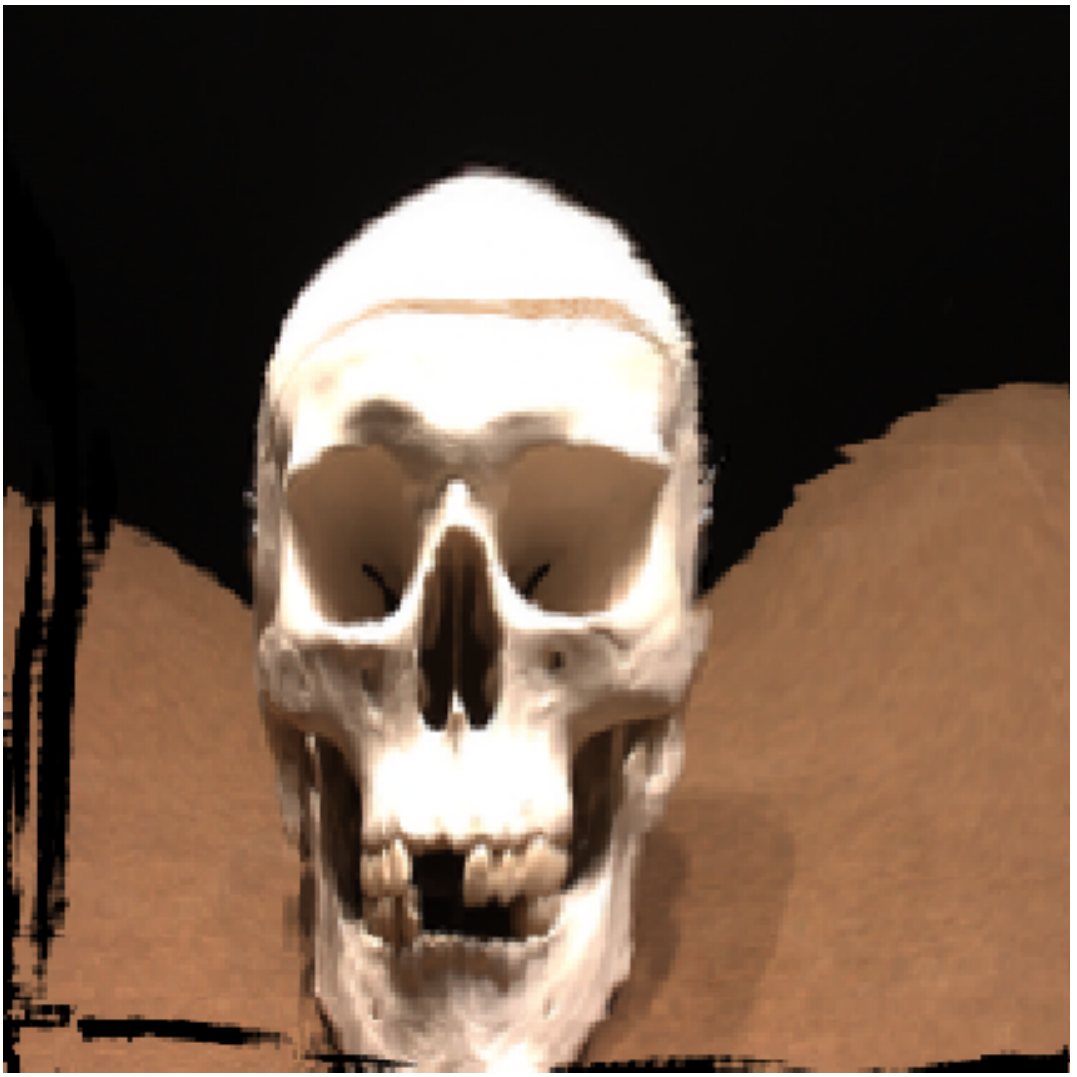}
    \vspace{0.3mm}
    \includegraphics[width=0.155\textwidth,height=0.12\textwidth]{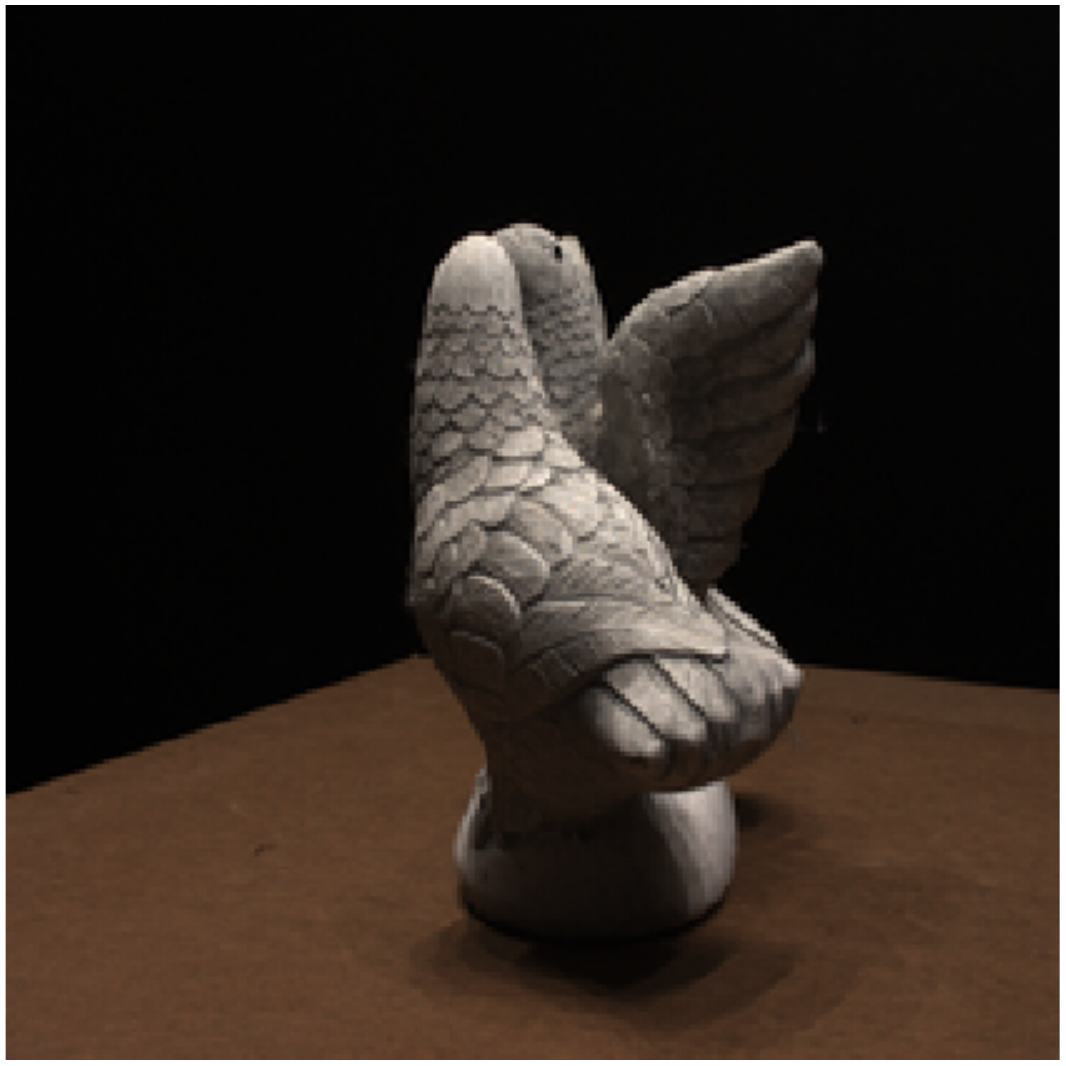}
    \includegraphics[width=0.155\textwidth,height=0.12\textwidth]{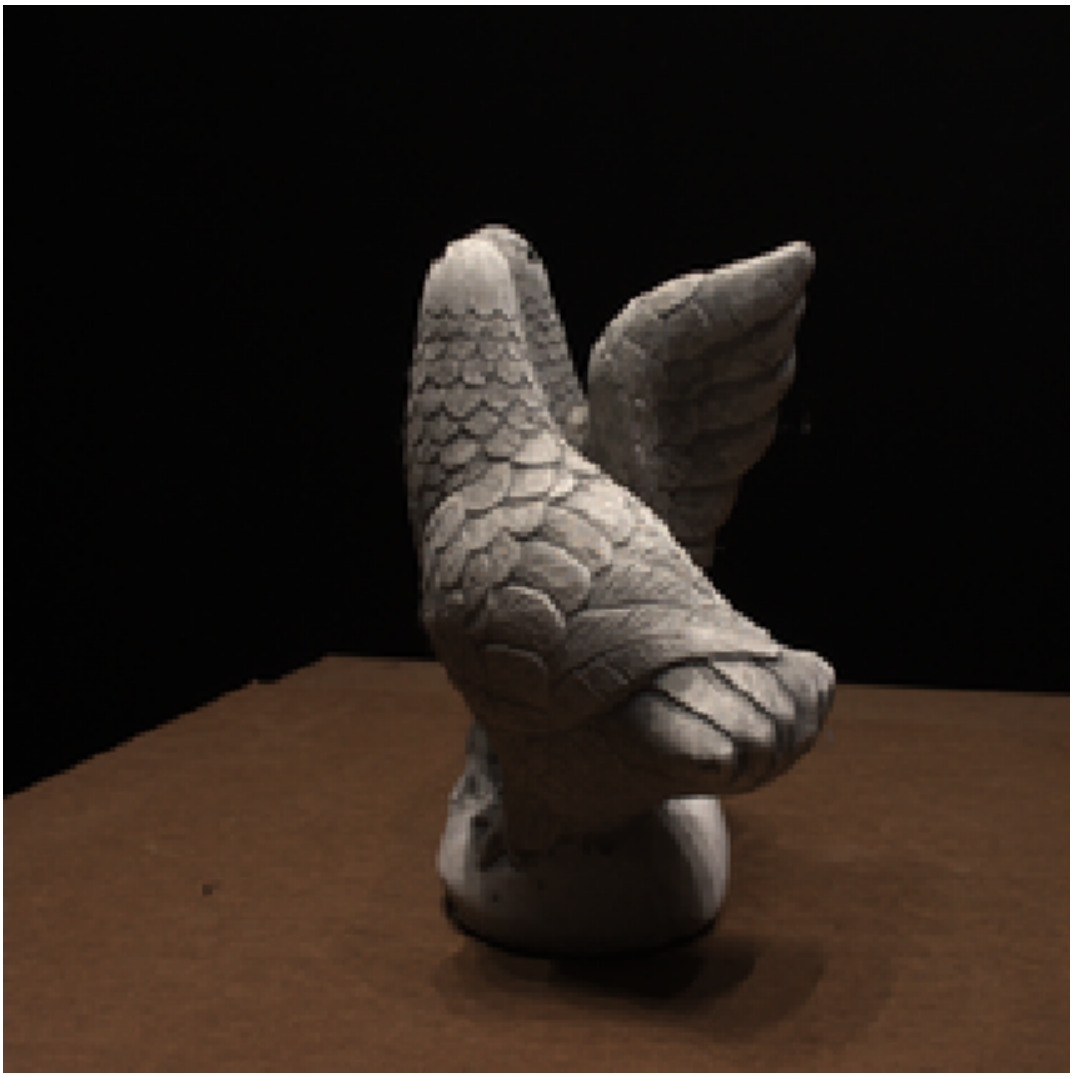}
    \includegraphics[width=0.155\textwidth,height=0.12\textwidth]{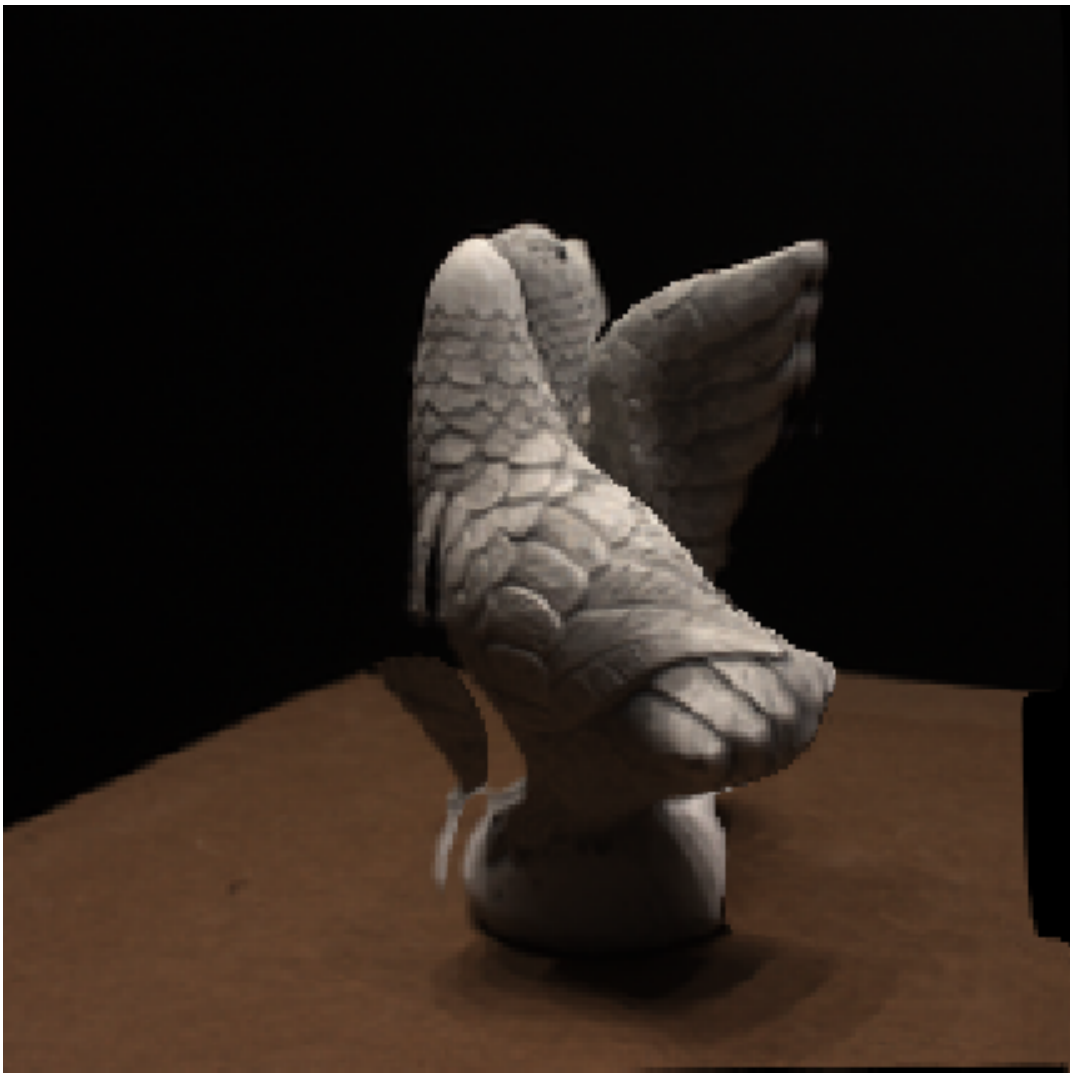}
    \includegraphics[width=0.155\textwidth,height=0.12\textwidth]{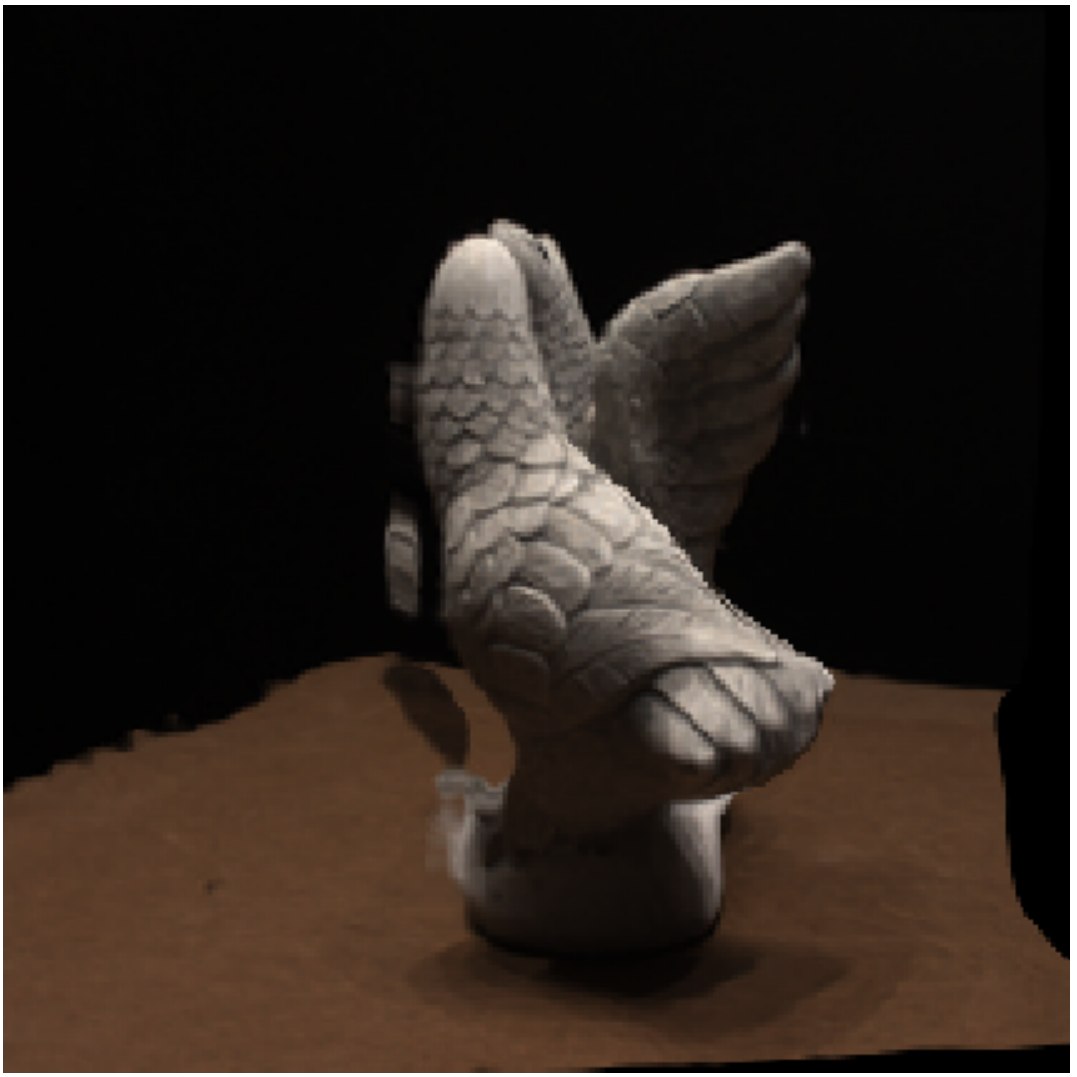}
    \includegraphics[width=0.155\textwidth,height=0.12\textwidth]{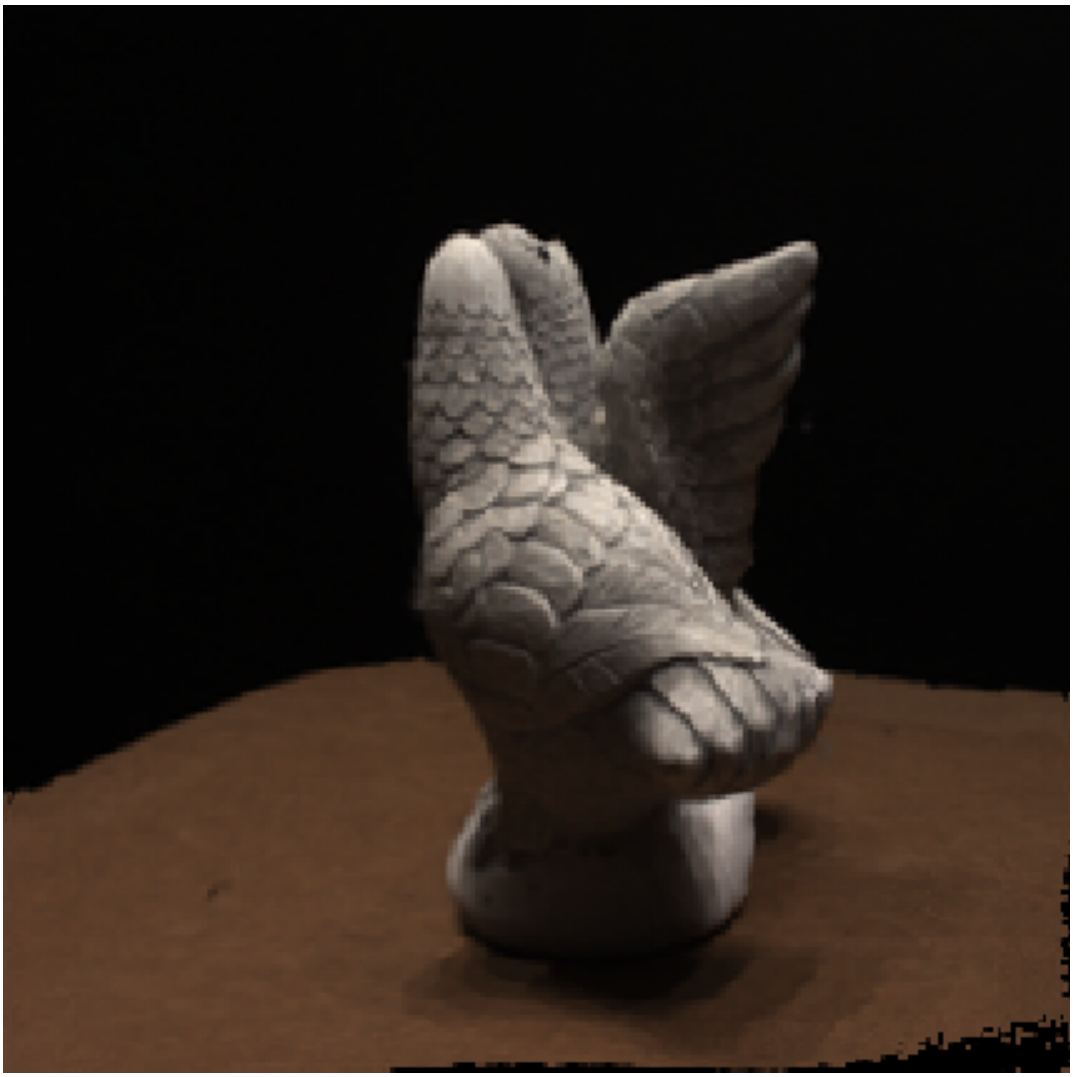}
    \includegraphics[width=0.155\textwidth,height=0.12\textwidth]{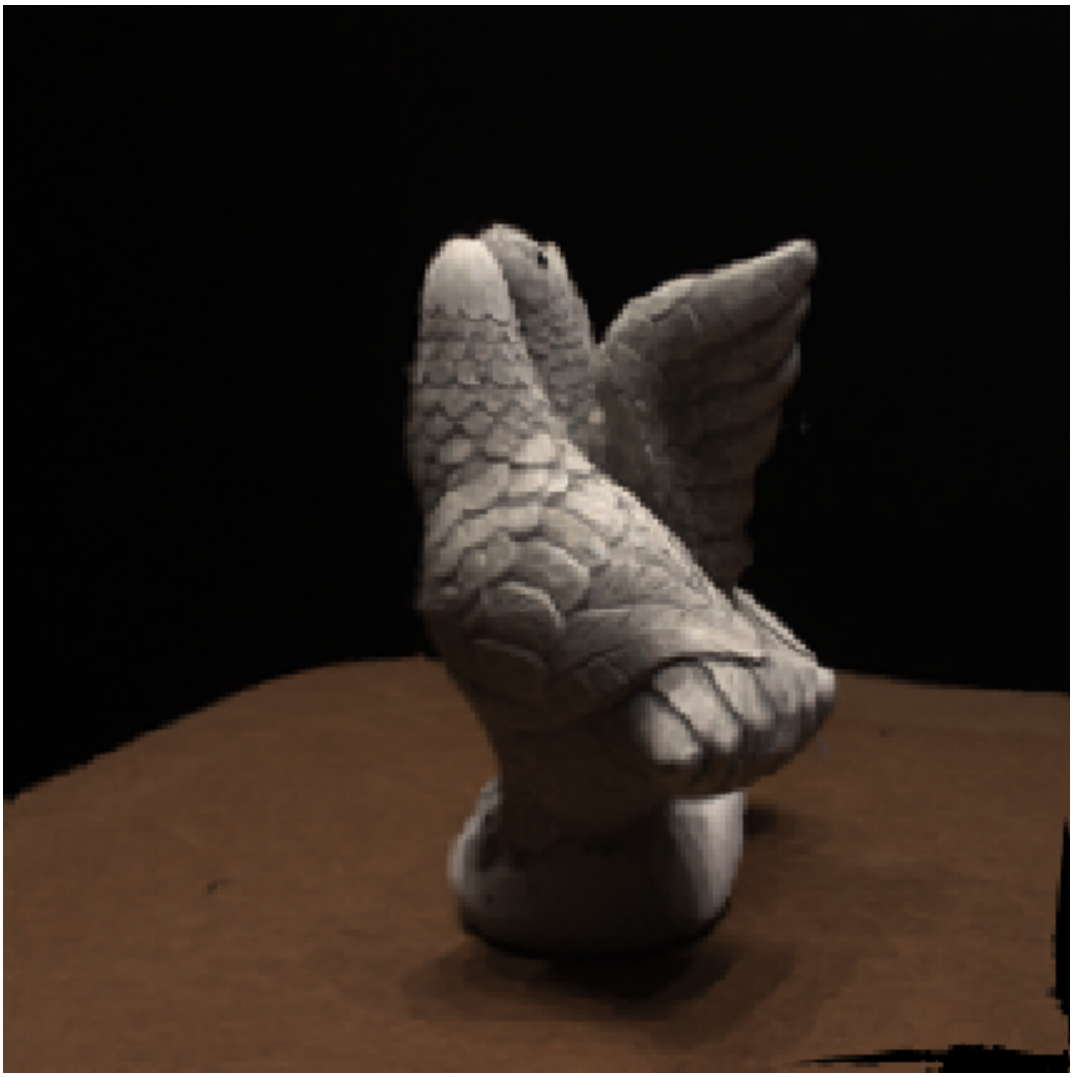}
    \vspace{0.3mm}
    \includegraphics[width=0.155\textwidth,height=0.12\textwidth]{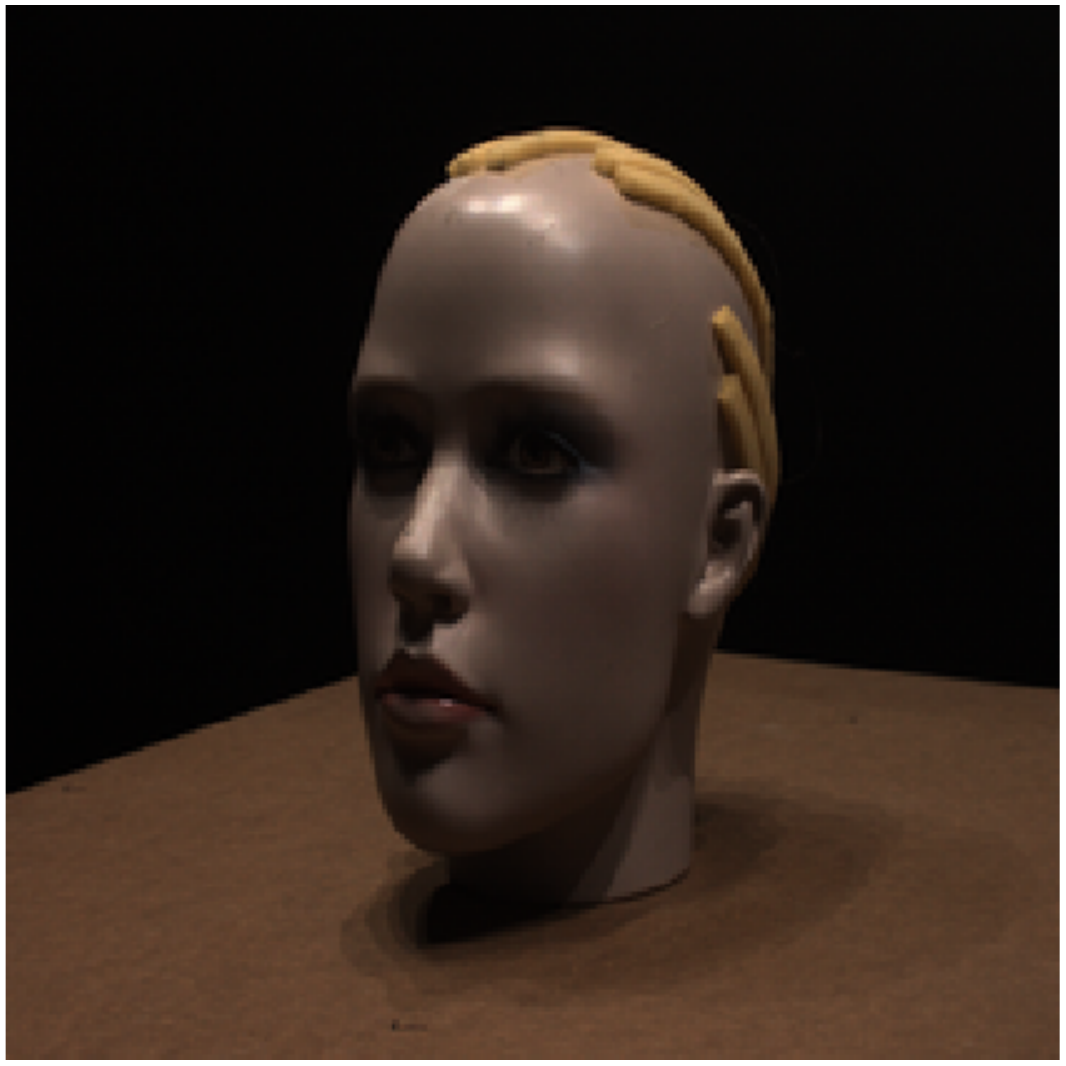}
    \includegraphics[width=0.155\textwidth,height=0.12\textwidth]{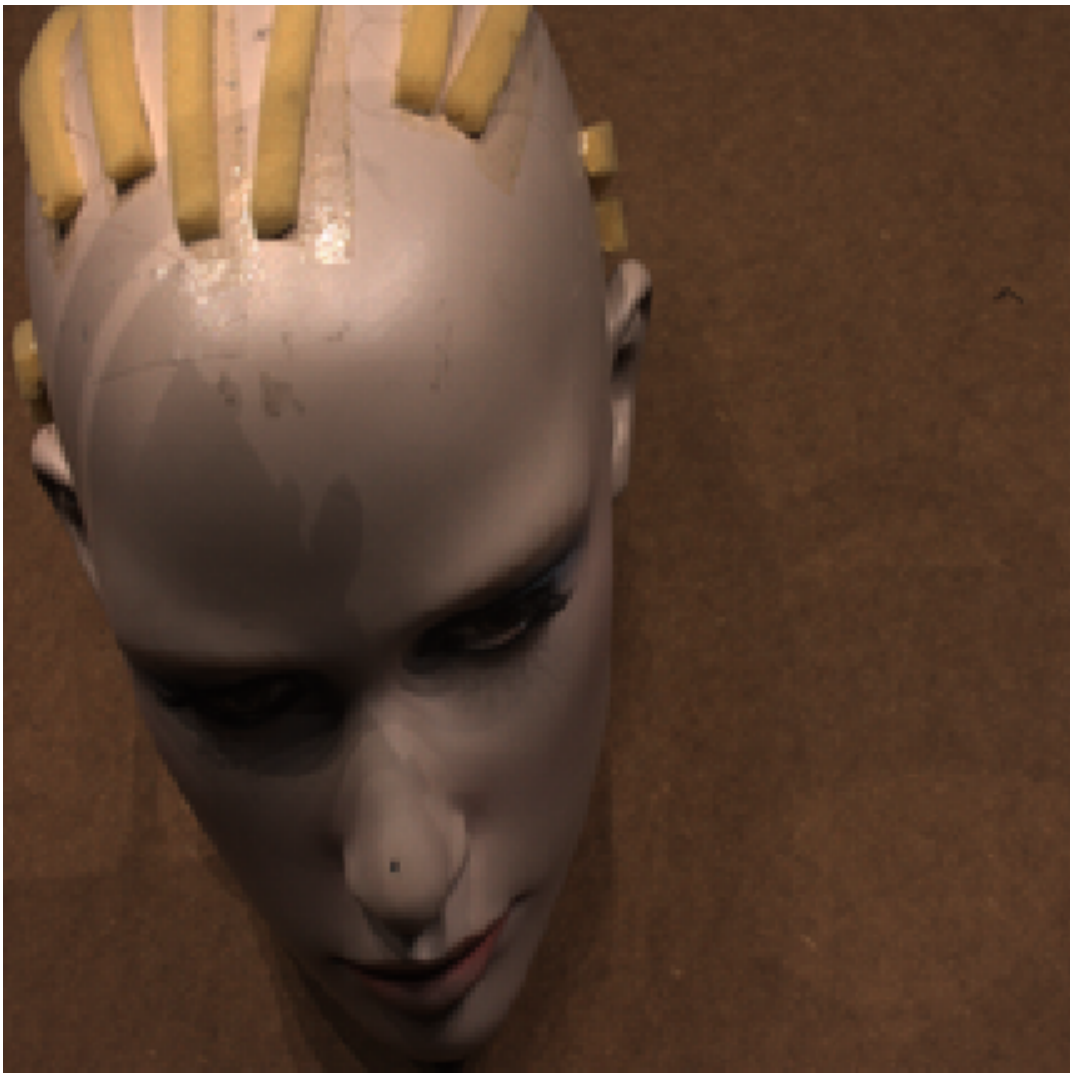}
    \includegraphics[width=0.155\textwidth,height=0.12\textwidth]{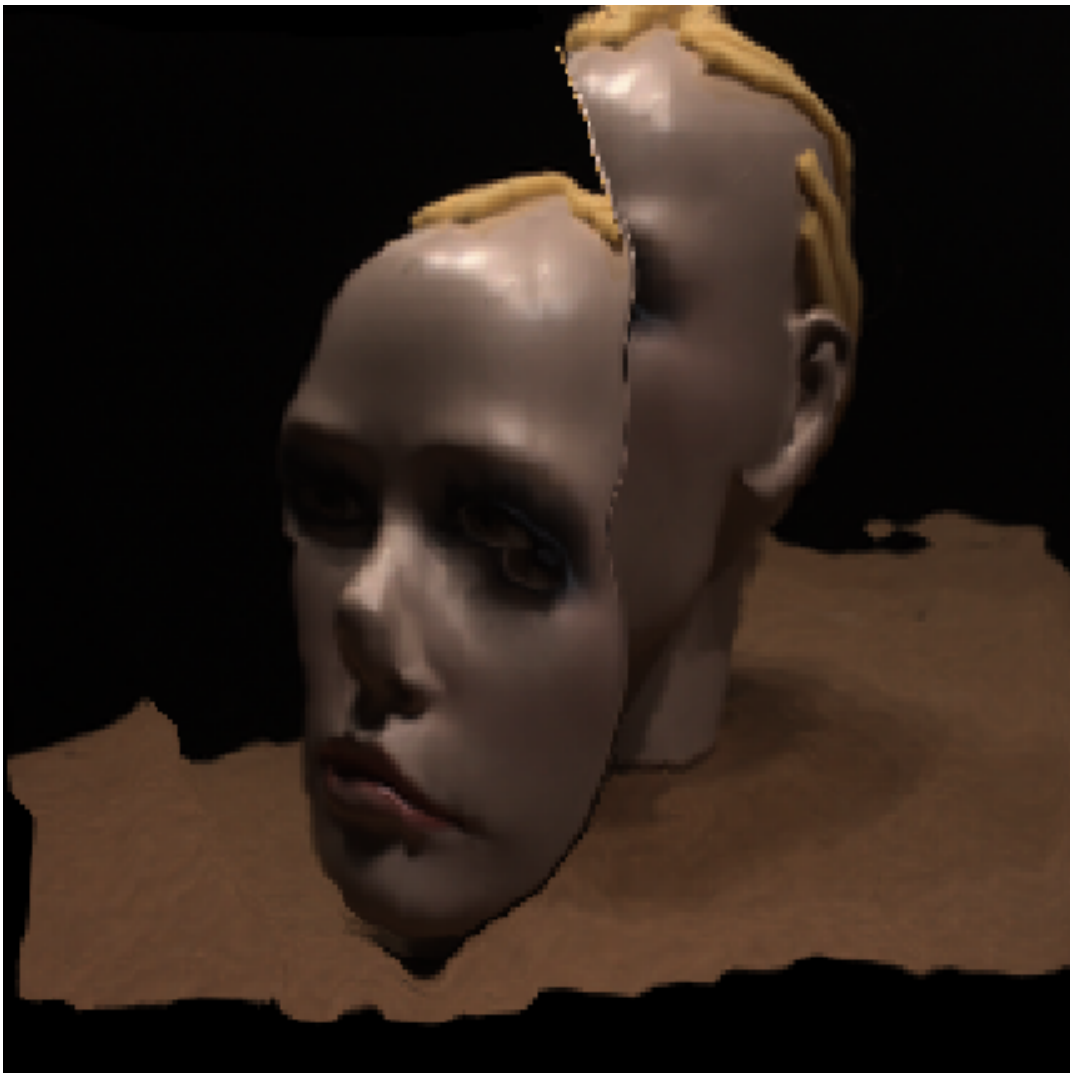}
    \includegraphics[width=0.155\textwidth,height=0.12\textwidth]{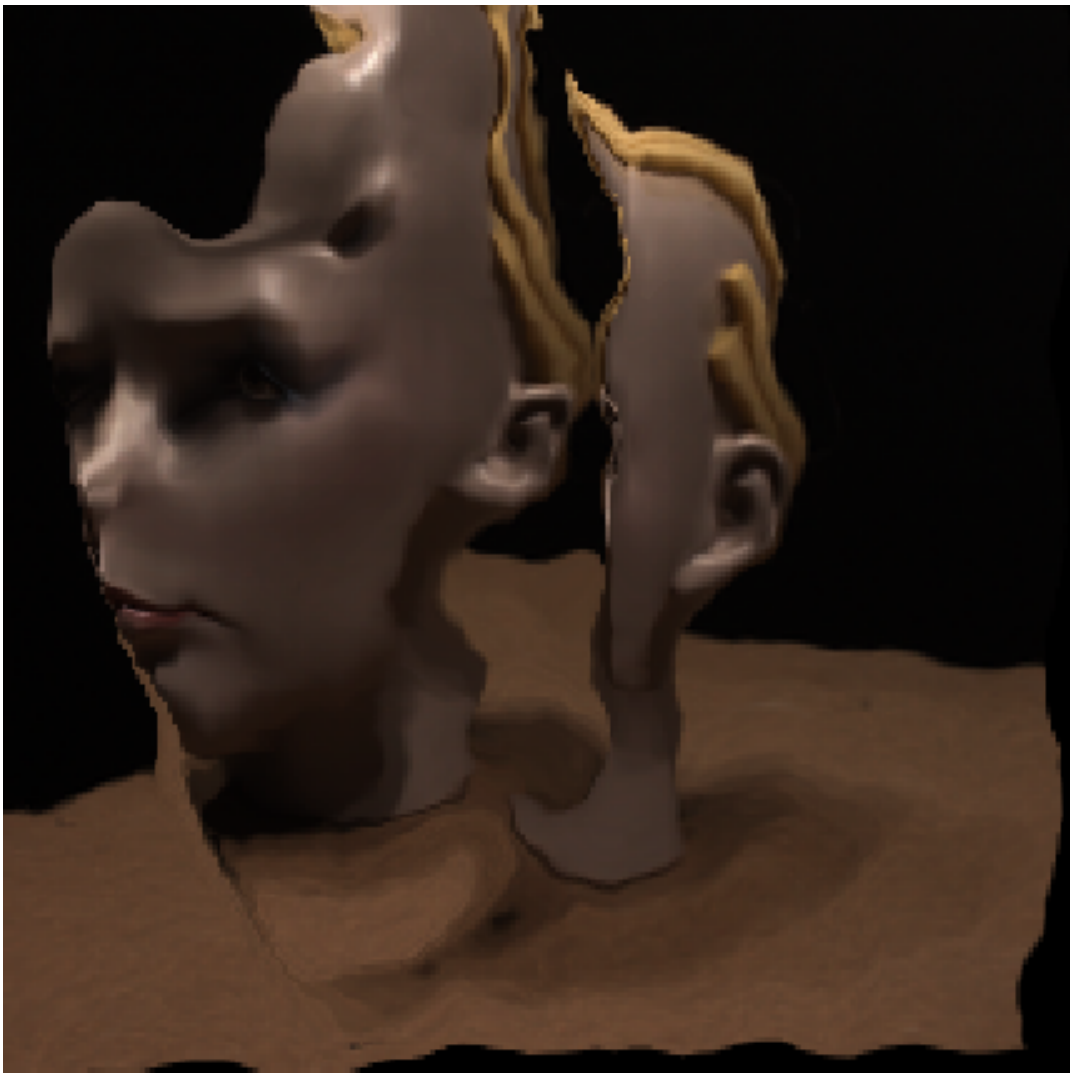}
    \includegraphics[width=0.155\textwidth,height=0.12\textwidth]{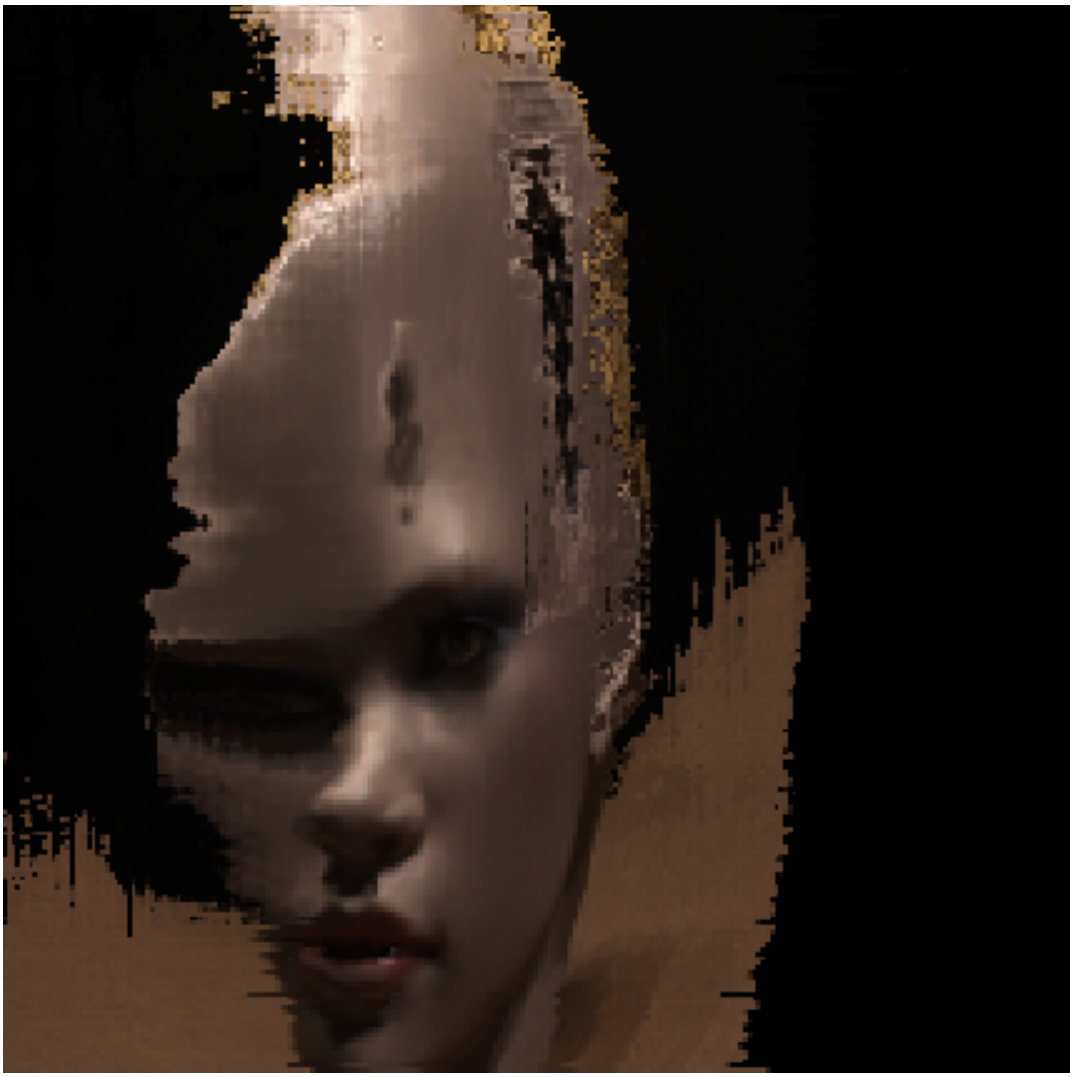}
    \includegraphics[width=0.155\textwidth,height=0.12\textwidth]{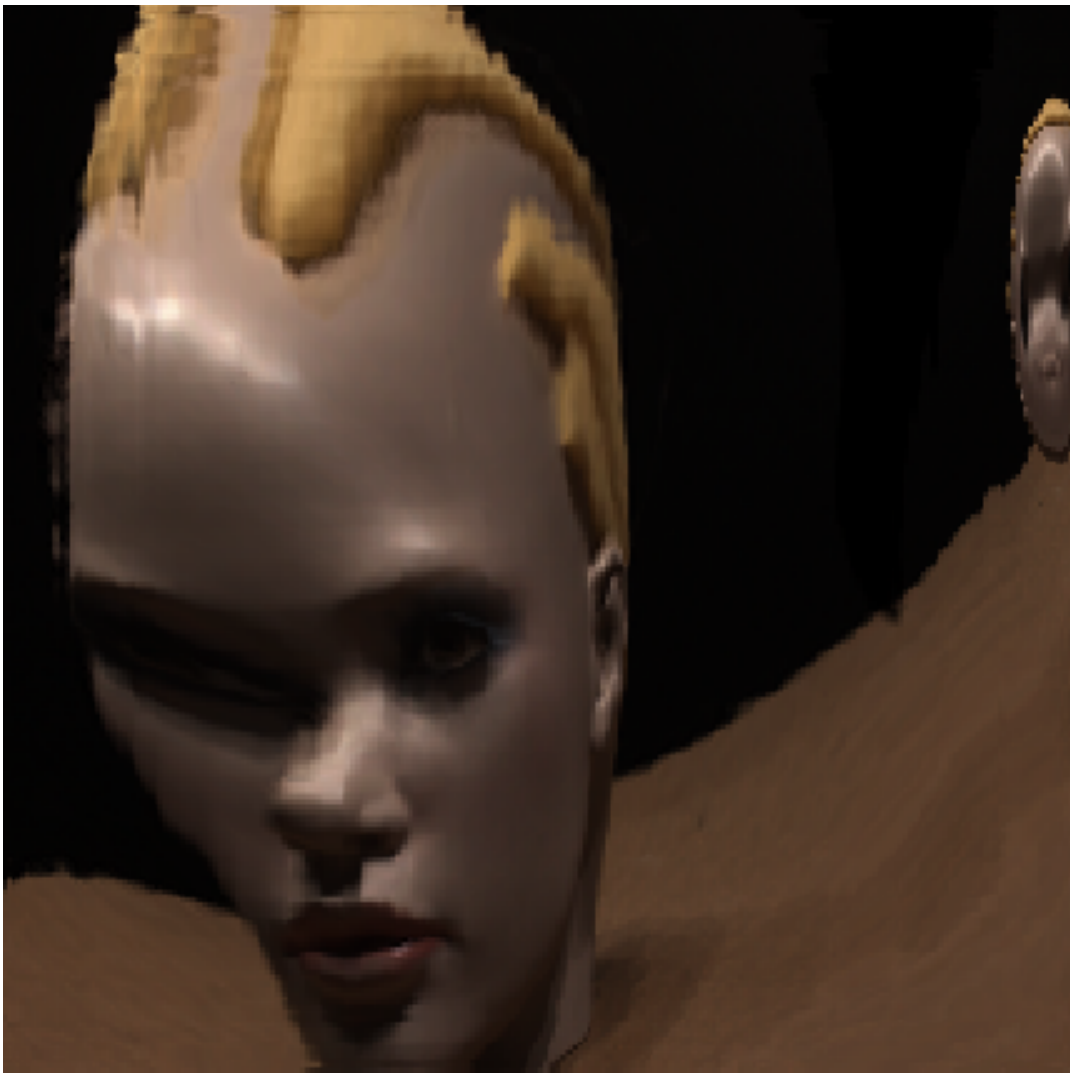}
    \vspace{0.3mm}
\small{
\hfill
\mpage{0.100}{Reference} \hfill
\mpage{0.100}{Target} \hfill
\mpage{0.100}{FlowNet2~\cite{FlowNet2}} \hfill
\mpage{0.100}{PWC-Net~\cite{PWC-Net}} \hfill
\mpage{0.100}{DGC+M-Net} \hfill
\mpage{0.100}{DGC-Net} \hfill
}
\vspace{0pt}
\caption{Qualitative results produced by different baseline methods on DTU. All optical flow approaches produce artifacts caused by warping the reference image. In contrast, DGC-Net and DGC+M-Net give the best results on this dataset without any object duplicates in the warped image. More examples presented in the supplementary. }\label{fig:comparison_dtu_qual}
\end{figure*}

{\small
\bibliographystyle{ieee}
\bibliography{egbib}
}

\IfFileExists{./supp.pdf}{%
\clearpage


\section*{Supplementary material}

# DGC-Net: Dense Geometric Correspondence Network
– Supplementary Material –

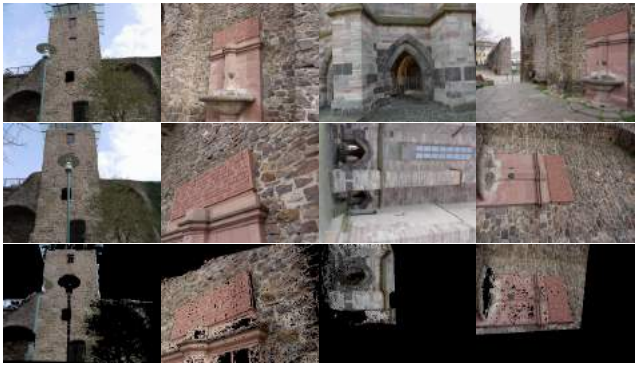

Figure 1: Citywall dataset [2] samples used for fine-tuning the model: the top row is a set of reference images; the middle row is corresponding target images; the bottom row is the warped reference images based on ground truth transformation data.

## 1. Implementation details

We train our network end-to-end using Adam [4] solver with $\beta_1 = 0.9$ and $\beta_2 = 0.999$. As a preprocessing step, the training images are resized to $240 \times 240$ and further mean-centered and normalized using mean and standard deviation of ImageNet dataset [1]. We use a batch size of 32, an initial learning rate of $10^{-2}$ which is gradually decreased during training. For fine-tuning on the Citywall dataset (Fig. 1), the learning rate is set to $10^{-4}$. The weight decay is initialized to $10^{-5}$ in all experiments and no dropout was used in our experiments. Our method is implemented using PyTorch framework [5] and trained on two NVIDIA Titan X GPUs.

## 2. Ablation study

**Dilated convolutions:** The quantitative evaluation of the proposed method without any dilation factors used in the correspondence map decoders is presented in Tab. 1. The dilated convolutions consistently improve the results on both, synthetic and realistic, datasets.

**Feature pyramid:** Additionally, we report AEPE metric for the estimates obtained at different levels of the feature pyramid of DGC-Net model in Tab. 2. We apply bilinear

| Method | Train:<br>Test: | Tokyo Time Machine (T) | |
|---|---|---|---|
| | | (T)-*aff* | (P)-*aff* |
| Rocco[6] | | 4.10 | 4.45 |
| DGC-Net *no dilation* | | 1.03 | 1.17 |
| DGC-Net | | **0.90** | **1.03** |

(a) Synthetic data.

| Method | Viewpoint ID | | | | |
|---|---|---|---|---|---|
| | I | II | III | IV | V |
| Rocco [6] | 15.02 | 28.23 | 29.27 | 36.57 | 43.68 |
| DGC-Net *no dilation* | 5.36 | **12.60** | 16.27 | 20.67 | 27.61 |
| DGC-Net | **5.12** | 13.01 | **15.08** | **20.14** | **26.47** |

(b) Realistic data (HPatches).

Table 1: The effect of using dilated convolutions. We report AEPE on synthetic 1a and realistic 1b datasets for the models solely trained on synthetic *affine* transformations provided by [6].

| Pyramid layer | Viewpoint ID | | | | |
|---|---|---|---|---|---|
| | I | II | III | IV | V |
| layer 0 $[15 \times 15]$ | 5.55 | 10.00 | 12.79 | 15.69 | 20.11 |
| layer 1 $[30 \times 30]$ | 3.34 | 7.65 | 10.80 | 13.37 | 18.30 |
| layer 2 $[60 \times 60]$ | 2.59 | 6.62 | 9.95 | 12.51 | 17.56 |
| layer 3 $[120 \times 120]$ | 1.80 | 5.83 | 9.25 | 11.90 | 16.96 |
| layer 4 $[240 \times 240]$ | **1.55** | **5.53** | **8.98** | **11.66** | **16.70** |

Table 2: AEPE metric for different viewpoint IDs of the HPatches dataset (lower is better) for the estimates obtained at different levels of the feature pyramid.

interpolation with according scale factors, *i.e.* $\{16, 8, 4, 2\}$ to compare the estimated and ground-truth correspondence maps pixel-wise. Clearly, Tab. 2 supports the idea of a hierarchical structure, as the error steadily decreasing from the top (layer 0) to the bottom (layer 4) layer of the pyramid.

## 3. Qualitative results

We show more qualitative results of pixel-wise dense correspondence estimation on the HPatches and DTU datasets in Fig. 4 and Fig. 5 respectively.

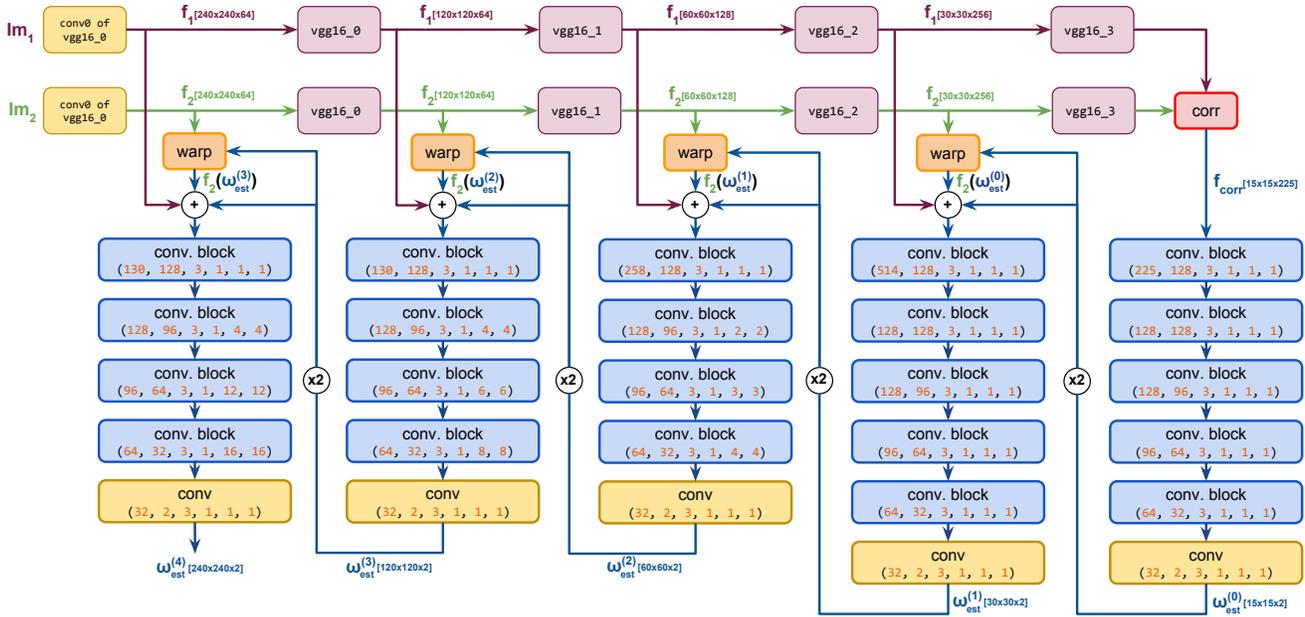

Figure 2: Overview of our proposed hierarchical fully-convolutional architecture `DGC-Net` consisting of 5 correspondence map decoders. Each decoder incorporates a chain of convolutional blocks. The structure of the convolutional block is illustrated in Fig. 3a.

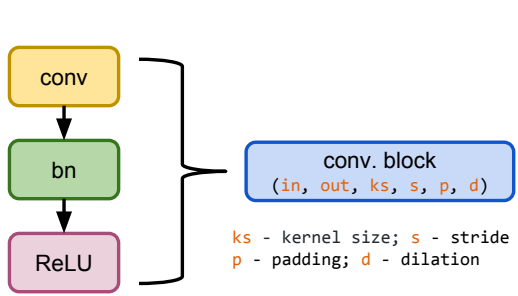

(a) Convolutional block

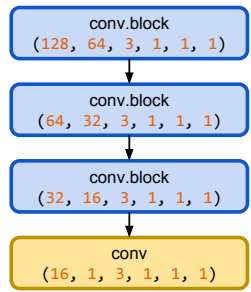

(b) Matchability decoder

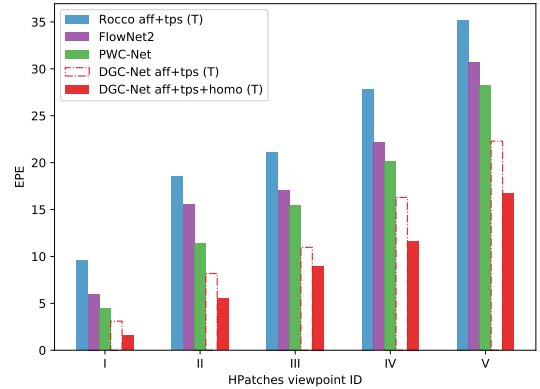

(c) AEPE metric on HPatches (lower is better)

Figure 3: The structure of the convolutional block 3a and the matchability decoder 3b used by DGC+M-Net model to predict an explainability mask. The proposed model clearly outperforms all strong baselines on realistic data 3c.

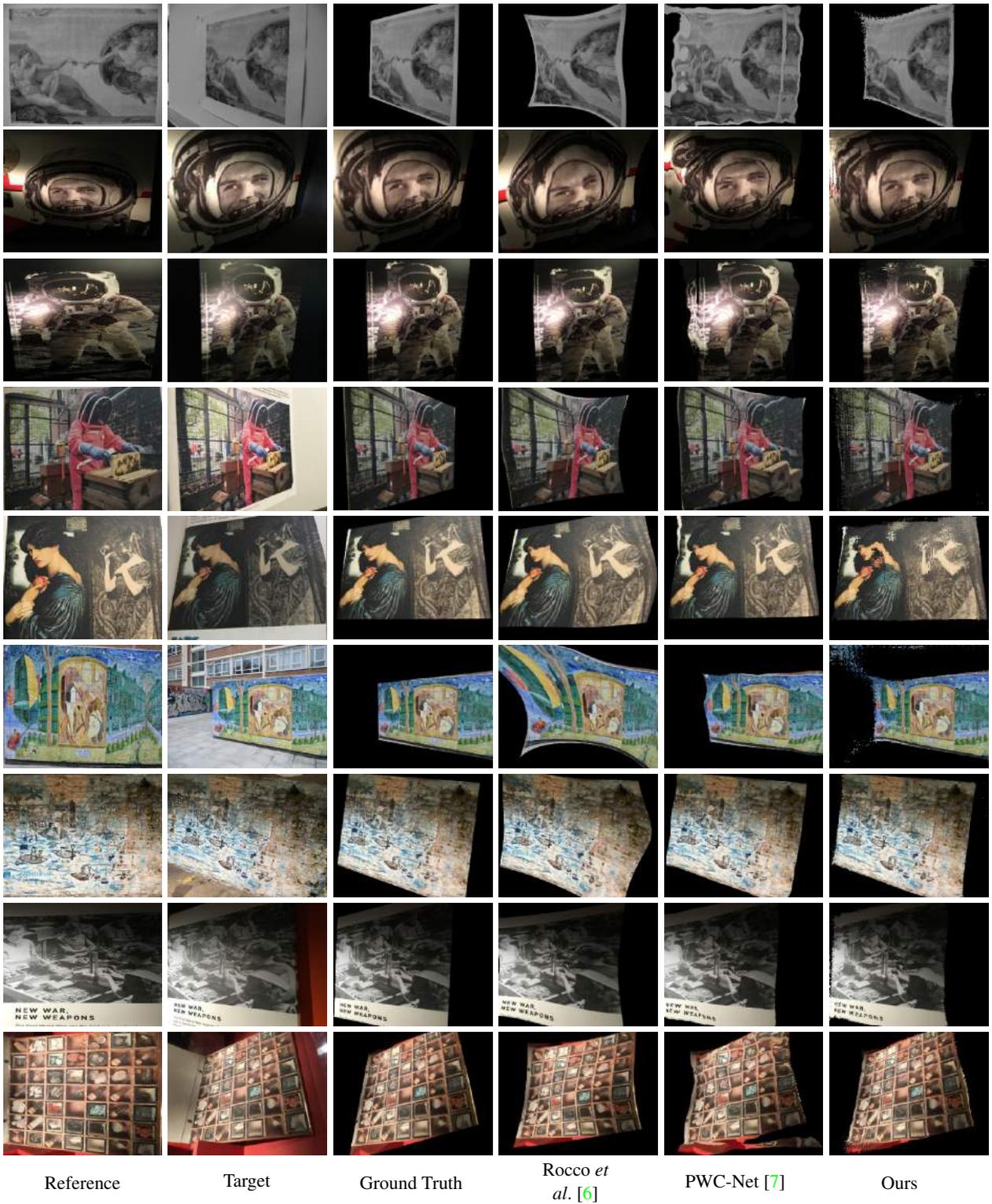

Figure 4: Qualitative comparisons between different algorithms on HPatches.

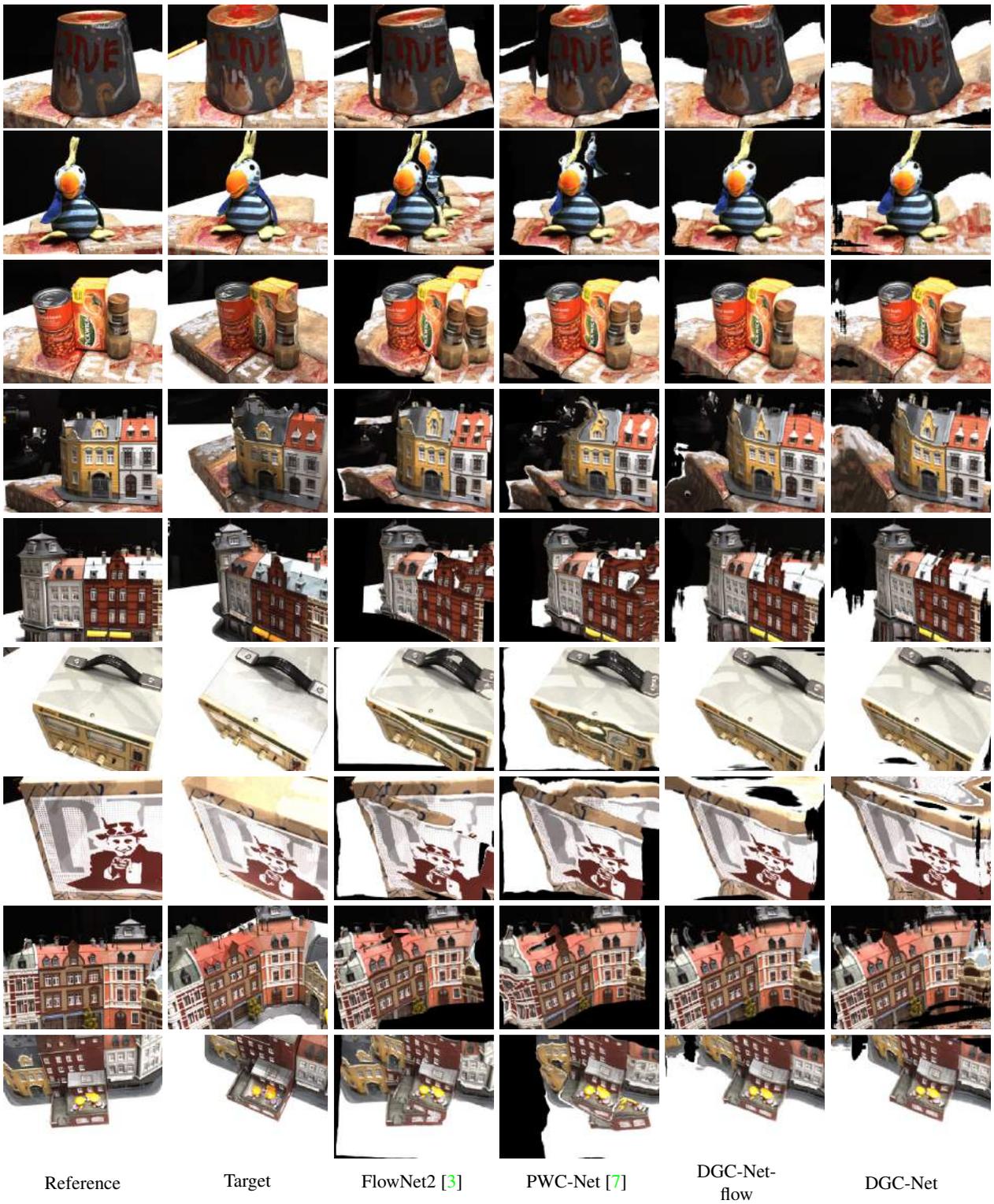

Figure 5: Qualitative comparisons between different algorithms on DTU. DGC-Net can handle relatively large transformations producing accurate results.


}

\end{document}